\pdfoutput=1

\documentclass[11pt]{article}

\usepackage[preprint]{acl}

\usepackage{xspace}
\usepackage{algorithm}
\usepackage{algpseudocode}
\usepackage{tikz}
\usepackage{fancybox}
\usepackage{booktabs}
\usepackage{multirow} 
\usepackage{threeparttable}
\usepackage{tabularx}
\usepackage{ragged2e}
\usepackage{enumitem}
\setlist{nolistsep}
\usepackage{booktabs} 
\usepackage{subcaption}
\usepackage{adjustbox}
\usepackage{caption}
\usepackage{graphicx} 
\newcommand{\SuperFewBench}[0]{\textsc{FewMMBench}: A Benchmark for Multimodal Few-Shot Learning
\xspace}
\newcommand{\SuperFewBenchShort}[0]{\textsc{FewMMBench}\xspace}

\newcommand{\CoTLong}[0]{Chain-of-Thought\xspace}

\usepackage{times}
\usepackage{latexsym}

\usepackage[T1]{fontenc}

\usepackage[utf8]{inputenc}

\usepackage{microtype}

\usepackage{inconsolata}

\usepackage{graphicx}

\title{\SuperFewBench}

\author{Mustafa Dogan\textsuperscript{1}
\quad Ilker Kesen\textsuperscript{2}
\quad \textbf{Iacer Calixto\textsuperscript{3,4}}
\quad \textbf{Aykut Erdem\textsuperscript{5,6}}
\quad \textbf{Erkut Erdem\textsuperscript{6,7}}
\vspace{0.2cm}\\
\textsuperscript{1} Aselsan Research  \textsuperscript{2} Department of Computer Science, University of Copenhagen \\
\textsuperscript{3} Department of Medical Informatics, Amsterdam UMC, University of Amsterdam \\ \textsuperscript{4} Amsterdam Public Health, Methodology, Amsterdam, The Netherlands\\
\textsuperscript{5} Ko\c{c} University, Department of Computer Engineering 
\textsuperscript{6} Ko\c{c} University, KUIS AI Center \\
\textsuperscript{7} Hacettepe University, Department of Computer Engineering \\ \texttt{\href{mustafadogan@aselsan.com}{mustafadogan@aselsan.com}}
}

\newcommand{\benchmark}{\textsc{FewMMBench}\xspace}

\begin{document}
\maketitle
\begin{abstract}

As multimodal large language models (MLLMs) advance in handling interleaved image-text data, assessing their few-shot learning capabilities remains an open challenge. In this paper, we introduce \benchmark, a comprehensive benchmark designed to evaluate MLLMs under few-shot conditions, with a focus on In-Context Learning (ICL) and Chain-of-Thought (CoT) prompting. Covering a diverse suite of multimodal understanding tasks, from attribute recognition to temporal reasoning, \benchmark enables systematic analysis across task types, model families, and prompting strategies. We evaluate 26 open-weight MLLMs from six model families across zero-shot, few-shot, and CoT-augmented few-shot settings. Our findings reveal that instruction-tuned models exhibit strong zero-shot performance but benefit minimally, or even regress, with additional demonstrations or CoT reasoning. Retrieval-based demonstrations and increased context size also yield limited gains. These results highlight \benchmark as a rigorous testbed for diagnosing and advancing few-shot capabilities in multimodal LLMs. The data is available at: \href{https://huggingface.co/datasets/mustafaa/FewMMBench}{this URL.}
\end{abstract}

\section{Introduction}
Multimodal large language models (MLLMs), models that jointly process images and text, have advanced significantly over the past two years. Beginning with GPT-4V \citep{openai2024gpt4technicalreport}, these models have evolved along two major axes. First, they have become more general-purpose, exhibiting stronger multimodal reasoning and instruction-following abilities; and second, they now support interleaved image–text inputs within a single prompt, enabling sequential and contextualized few-shot learning.

\begin{figure}[ht!]
    \centering
    \includegraphics[width=\columnwidth]{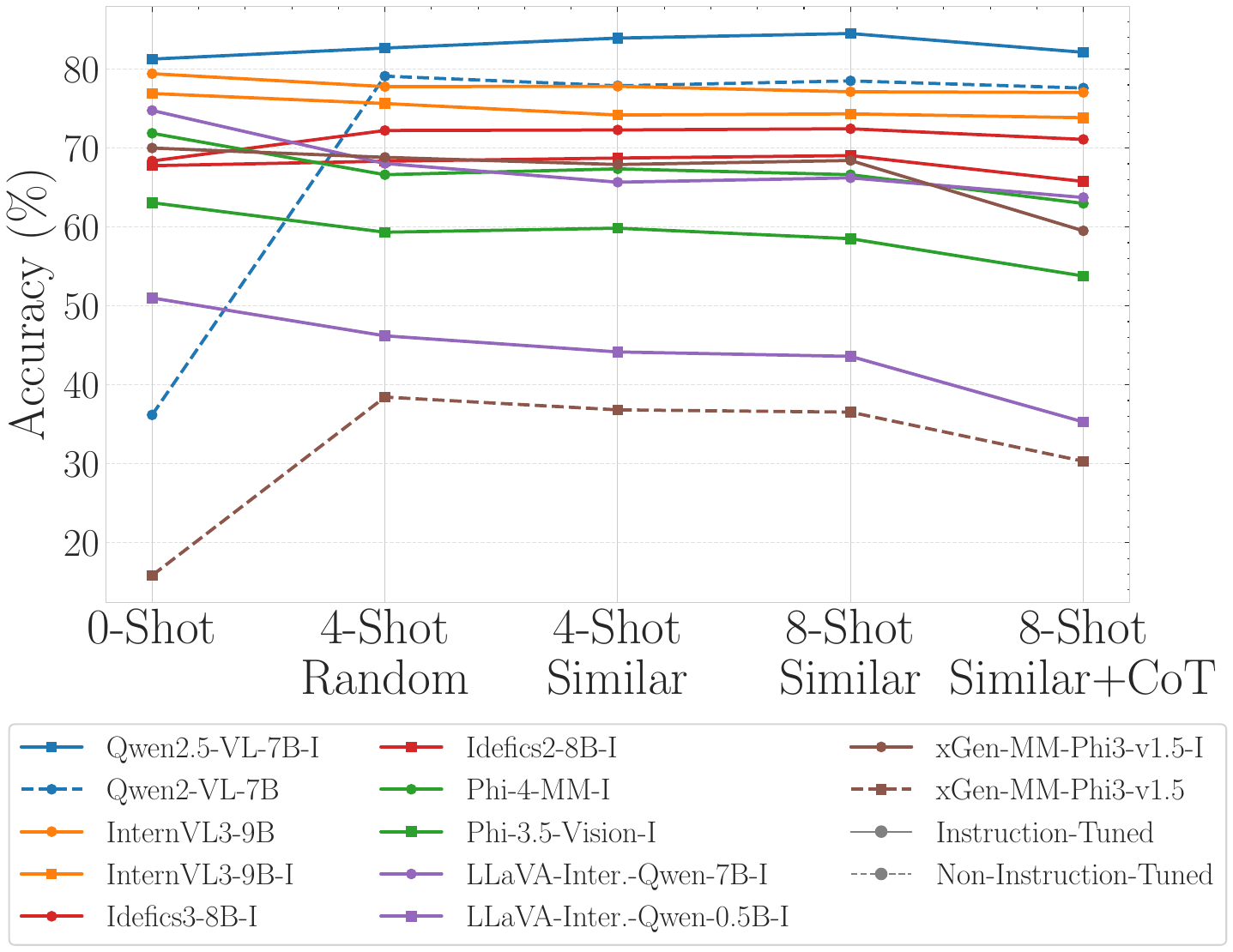}
    \caption{\textbf{Performance of selected MLLMs on \benchmark across different evaluation settings.} We compare instruction-tuned and non-instruction-tuned models under zero-shot, few-shot (random and similarity-based), and CoT-augmented few-shot configurations. Results show that few-shot prompting does not consistently improve performance for instruction-tuned models, even when demonstrations are semantically similar to the query or when the number of examples increases. Notably, CoT prompting often leads to a performance drop, suggesting modality-specific limitations in current CoT strategies.}
    \label{fig:fewmmbench_findings}
\end{figure}

Early benchmarks such as MMBench \citep{liu2024mmbench} and SeedBench \citep{li2024seed} offered fine-grained evaluations across a wide spectrum of vision–language skills, ranging from basic object recognition to complex social reasoning. Similarly, MME \citep{fu2024mmecomprehensiveevaluationbenchmark} assessed both low-level perception (e.g., object existence, color identification) and high-level cognition (e.g., commonsense inference) using binary question formats. More recent efforts, such as VisualCoT \citep{shao2024visual} and VL-ICL Bench \citep{zong2025vlicl}, have begun to probe specific reasoning capabilities, such as multi-step visual inference and in-context learning with a few visual exemplars. However, these benchmarks primarily focus on zero-shot evaluation or isolated abilities, and do not systematically test MLLMs in a \emph{few-shot}, \emph{in-context} setting with \emph{interleaved} image–text streams and \emph{chain-of-thought} prompting.

To address this gap, we introduce \benchmark, a comprehensive benchmark for evaluating multimodal few-shot learning. \benchmark is designed to assess two emerging capabilities: \textbf{In-Context Learning}, where models are guided by a small set of image–text demonstration examples, and \textbf{Chain-of-Thought prompting}, where models are encouraged to generate intermediate reasoning steps. The benchmark spans nine diverse tasks, ranging from low-level perception (e.g., attribute recognition, object counting) to higher-level reasoning (e.g., spatial relations, grounded coreference, commonsense, and temporal inference). For each task, we construct fixed support sets of 4-8 examples, selected either randomly or via a retrieval-based strategy that identifies semantically similar exemplars. 

We evaluate 26 open-weight MLLMs from six model families, Qwen~\citep{Qwen2.5-VL, Qwen2-VL}, InternVL~\citep{wang2024enhancingreasoningabilitymultimodal, chen2024expanding, chen2024far, zhu2025internvl3exploringadvancedtraining}, Idefics~\citep{laurencon2024building, NEURIPS2024_a0303731}, Phi~\citep{microsoft2025phi4minitechnicalreportcompact, abdin2024phi3technicalreporthighly}, LLaVA~\citep{li2025llavanextinterleave}, and xGen-MM~\citep{xue2024xgenmmblip3familyopen}, ranging from 1B to 9B parameters, under four evaluation settings: \textbf{zero-shot}, \textbf{random few-shot}, \textbf{retrieved few-shot}, and \textbf{few-shot with CoT prompting}. Figure~\ref{fig:fewmmbench_findings} summarizes the overall empirical trends observed across models, including the varying effects of demonstrations, retrieval strategies, and CoT prompting.

Beyond these empirical observations, the primary contributions of \benchmark lie in its benchmark design and evaluation methodology. Specifically, \benchmark (i) introduces a controlled framework that jointly supports random and semantically similar demonstration examples, where similar exemplars are selected using a graph-cut-based algorithm to balance relevance and diversity; (ii) provides detailed chain-of-thought rationales within demonstrations, enabling systematic analysis of reasoning-guided multimodal inference; and (iii) supports an exhaustive evaluation protocol that extends beyond response accuracy to include perplexity-based pairwise comparisons and uncertainty quantification.

\section{Related Work}
\subsection{Few-Shot Learning in Multimodal Models}
Early multimodal large language models (MLLMs) were generally limited in their ability to process multiple image–text pairs within a single prompt, with only a few constrained exceptions~\citep{tsimpoukelli2021multimodal,koh2023grounding,alayrac2022flamingo}. The release of GPT-4V marked a turning point, demonstrating robust reasoning over interleaved image–text sequences~\citep{openai2024gpt4technicalreport} and motivating the creation of large-scale, interleaved multimodal corpora~\citep{zhu2023multimodal,laurenccon2023obelics,li2023mimicitmultimodalincontextinstruction,zhao2024mmicl}. As a result, newer models natively support mixed-modality inputs~\citep{laurencon2024building,xue2024xgenmmblip3familyopen,Qwen2.5-VL,wang2024enhancingreasoningabilitymultimodal}.

Despite this progress, the mechanisms and limitations of few-shot multimodal reasoning remain underexplored, particularly in light of recent findings on CoT and ICL. Prior studies have revealed that CoT exemplars may offer limited benefit relative to zero-shot CoT~\citep{cheng-etal-2025-revisiting-chain}, that multimodal models struggle with ICL despite mixed-modality pretraining~\citep{doveh2024multimodalincontextlearningvision}, that demonstrations often help by providing structure rather than ground-truth labels~\citep{min-etal-2022-rethinking}, and that fine-tuning can degrade or alter CoT reasoning behaviors~\citep{lobo-etal-2025-impact}. These works highlight shortcomings in exemplar dependence, instruction dominance, ICL fidelity, and the stability of reasoning traces.

Our benchmark directly targets these open questions by providing controlled few-shot settings with structured demonstrations and explicit CoT rationales, enabling systematic analysis of how multimodal models acquire and apply reasoning patterns in context.

\begin{figure*}[t!]
    \centering
    \includegraphics[width=\linewidth,trim={0 3.3cm 0 0},clip]{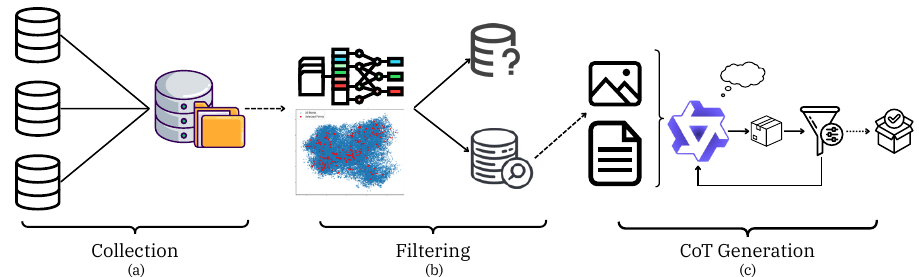}
    \caption{\textbf{Dataset Curation Pipeline for \benchmark.} (a) Task instances are collected and organized based on linguistically meaningful phenomena.
(b) We extract visual and textual features for each instance and construct query sets using a Graph Cut-based submodular selection strategy, ensuring both diversity and representativeness.
(c) CoT rationales are generated using the Qwen2.5-VL-7B-Instruct model. If the initial prediction is incorrect, the correct answer is injected and a new rationale is generated. An automated filter retains only high-quality examples.}
    \label{fig:fewmmbench_creation}
\end{figure*}

\subsection{Existing Vision-Language Benchmarks}
Recent advances in MLLM evaluation have introduced numerous benchmarks targeting different aspects of vision–language understanding. Broad frameworks such as VHELM~\citep{lee2024vhelm} offer standardized evaluation across perception, reasoning, fairness, multilinguality, and safety, while more focused efforts like MM-Vet v2~\citep{yu2024mmvetv2} and CURE~\citep{chen-etal-2024-measuring} assess specific abilities such as compositional reasoning and chain-of-thought consistency.

Few-shot prompting has also been explored, but limitations remain. VL-ICL Bench~\citep{zong2025vlicl} lacks CoT exemplars and relies on zero-shot prompting. M\textsuperscript{3}CoT~\citep{chen-etal-2024-m3cot} offers CoT rationales but pairs them with randomly selected demonstrations, reducing contextual alignment. VisualCoT~\citep{shao2024visual} instead provides explicit procedural instructions rather than reasoning trajectories. These issues hinder systematic assessment of multimodal reasoning, whereas \benchmark uses semantically aligned demonstrations with explicit CoT rationales.

Other work emphasizes dynamic or context-rich evaluation. Mementos~\citep{wang-etal-2024-mementos} studies sequential image reasoning, VisDiaHalBench~\citep{cao-etal-2024-visdiahalbench} examines hallucination in multi-turn visual dialogue, and multi-image benchmarks like MMIU~\citep{meng2025mmiu} and MIBench~\citep{liu-etal-2024-mibench} probe spatial reasoning and multi-image coordination. These benchmarks, however, remain orthogonal to few-shot reasoning, as they do not provide structured demonstrations or analyze how models learn reasoning patterns.

In contrast, \benchmark offers a unified, linguistically grounded few-shot framework. Beyond curated demonstrations and CoT rationales, it introduces two methodological contributions absent from prior work: a graph-cut–based construction of representative, non-redundant support sets, and perplexity-based accuracy with uncertainty quantification to mitigate order-sensitivity in ICL. This yields a more stable and diagnostically robust evaluation of multimodal reasoning.

\section{Curating \SuperFewBenchShort}
\benchmark is a probing benchmark specifically designed to evaluate the linguistic and reasoning capabilities of MLLMs in few-shot settings. It spans a broad spectrum of tasks, including recognition, numerical reasoning, spatial and temporal inference, commonsense understanding, and linguistic grounding. As shown in Figure~\ref{fig:fewmmbench_creation}, we adopt a systematic dataset construction pipeline to ensure diversity, representativeness, and quality across all evaluation scenarios. This pipeline involves two core steps: (i) selection of representative and challenging samples guided by visual-textual similarity, and (ii) generation of explicit chain-of-thought explanations to guide model inference.

We begin by outlining the benchmark tasks in Section~\ref{tasks}, with illustrative examples and dataset statistics shown in Figure~\ref{fig:fewmmbench_overview}. Further statistics are presented in Appendix~\ref{appd_data_statistics}. We then describe two key factors critical to our few-shot evaluation setup: example selection (Section~\ref{example_s}) using submodular optimization, and CoT rationales generation (Section~\ref{cot_d_g}) to support step-by-step reasoning.

\begin{figure*}[t!]
    \centering
    \includegraphics[width=\linewidth,trim={0 0.6cm 0.3cm 0},clip]{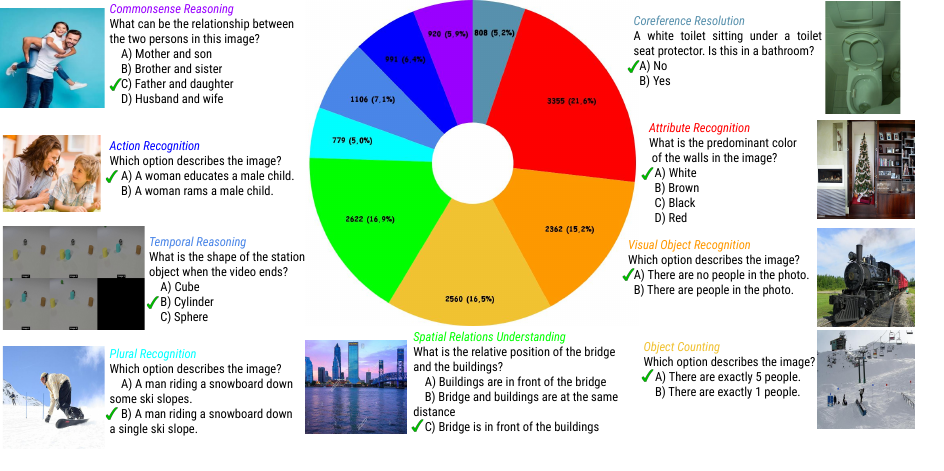}
    \caption{\textbf{Task distributions and representative examples in \benchmark.} The pie chart illustrates the distribution of samples across the nine tasks in the benchmark. Each surrounding example depicts a sample question corresponding to a specific task.%
    These examples highlight the diversity of visual-linguistic reasoning skills covered by \SuperFewBenchShort, spanning low-level perception, numerical reasoning, and high-level cognitive inference.}
    \label{fig:fewmmbench_overview}
\end{figure*}

\subsection{Tasks}
\label{tasks}
\textbf{Attribute Recognition} evaluates a model’s ability to associate visual properties with objects in images. We employ two formats: MCQA and caption/foil classification. The MCQA task is derived from the \textit{Attribute Recognition} dimension of MMBench \citep{liu2024mmbench} and the \textit{Instance Attributes} dimension of SeedBench \citep{li2024seed}, assessing recognition of attributes such as color, shape, and material. Foiling task, adapted from the \textit{Visual Genome Attribution} subset of ARO \citep{yuksekgonul2023when}, tests the ability to distinguish between correct and incorrect attribute associations in descriptive captions. 
The sample pool for MCQA consists of $\sim$5K instances, and the caption/foil classification pool contains $\sim$30K instances.

\textbf{Object Recognition} evaluates a model’s ability to detect and identify objects within images, including specific instances, general categories, and object presence. We assess this through both MCQA and caption/foil classification. For MCQA, we adopt the \textit{Instance Identity} dimension from SeedBench \citep{li2024seed}, which examines a model's ability to recognize specific objects, their categories, or their presence in a given scene. Additionally, we include the \textit{Multi-Class Identification} task from GVTBench \citep{wang2023makes}, where models must determine the presence of specified objects in diverse visual contexts drawn from the MS-COCO \citep{lin2014microsoft} and Visual Commonsense Reasoning \citep{zellers2019vcr} datasets. For the caption/foil classification task, we utilize the \textit{Existence} dimension from VALSE \citep{parcalabescu-etal-2022-valse}, assessing whether models accurately interpret existential quantifiers by determining if textual descriptions correctly represent the visual content. Our evaluation set includes 20K MCQA and 505 foil detection examples.

\textbf{Plurality Recognition} assesses a model's ability to distinguish singular from plural entities within visual scenes, requiring an understanding of numerical differences and alignment between textual descriptions and images.
We evaluate this capability by adopting the \textit{Plurality} foiling task from VALSE \citep{parcalabescu-etal-2022-valse}.
This task specifically tests whether models can correctly identify images depicting single versus multiple object instances, effectively evaluating their comprehension of singular and plural semantics in visual contexts.
The evaluation set consists of 851 samples.

\textbf{Object Counting} assesses a model’s ability to accurately determine the number of specific objects present within an image, integrating object recognition with numerical reasoning.
We evaluate this skill again using two formats: MCQA and caption/foil classification.
The MCQA evaluation adopts the \textit{Instance Counting} dimension from SeedBench \citep{li2024seed}, testing whether models can accurately identify and enumerate particular objects within visual scenes.
For caption/foil classification, we employ the \textit{Counting} foiling task from VALSE, requiring models to distinguish captions that correctly represent the number of entities from those that do not. This evaluates the models' abilities to interpret and reason about visual quantities. The evaluation dataset includes approximately 2,500 samples for each evaluation format, covering diverse counting scenarios and complexities.

\textbf{Spatial Relations Understanding} assesses a model's capability to interpret the relative positions of objects within visual scenes, which involves accurately grounding objects and identifying spatial relationships such as ``above", ``below", or ``next to”. We evaluate this skill using both MCQA and caption/foil classification tasks. For MCQA, we incorporate the \textit{Spatial Relationship} dimension from MMBench \citep{liu2024mmbench}, which tests models' proficiency in identifying relative object positions, and the \textit{Spatial Relations} dimension from SeedBench \citep{li2024seed}, which further challenges models to locate and compare two referenced objects to determine their spatial relationship accurately. For caption/foil classification, we aggregate data from multiple sources: the \textit{Spatial Relations} task from VALSE, evaluating the interpretation of spatial descriptions; VSR \citep{liu-etal-2023-visual}, comprising image-caption pairs labeled for spatial accuracy; the Visual Genome Relation category from ARO \citep{yuksekgonul2023when}, testing the grounding of spatial terms; and What’sUp \citep{kamath-etal-2023-whats}, featuring controlled image-caption pairs specifically designed to isolate spatial reasoning skills. The evaluation set includes 1K  MCQA and 29K caption/foil classification samples.

\textbf{Action Recognition} evaluates a model’s ability to identify and interpret actions depicted within images. Unlike object recognition, this task requires understanding dynamic interactions, contextual cues, and the specific roles of participants in a scene. We assess this ability using caption/foil classification, focusing on the \textit{Action} task from VALSE, which addresses two primary challenges: action replacement and actant swap. In the action replacement scenario, models must determine whether the described action accurately reflects the visual content (e.g., distinguishing ``a teacher looks/stands”). The actant swap task tests whether the model correctly identifies the participants involved in the action and their respective roles (e.g., discerning whether the teacher or the student is performing the action). These tasks require fine-grained understanding of actions and scene context. The evaluation set includes nearly 1600 samples.

\textbf{Commonsense Reasoning} measures a model’s ability to integrate visual perception with real-world knowledge, requiring logical inference that goes beyond basic object recognition. We assess this using MCQA samples sourced from four different datasets, each addressing a unique aspect of commonsense understanding. SeedBench \citep{li2024seed} contributes two key dimensions: \textit{Visual Reasoning}, which evaluates whether models can draw logical conclusions from visual input (e.g., determining if a bicycle with a flat tire is still usable), and \textit{Visual Referring Expression}, which tests the ability to locate objects based on descriptive cues (e.g., identifying which person in a group is holding a phone). MMBench \citep{liu2024mmbench} provides tasks on \textit{Social Relations} (e.g., recognizing familial connections through resemblance), \textit{Physical Relations} (e.g., predicting which object is likely to roll down a slope), and \textit{Natural Relations} (e.g., identifying animals likely to migrate). The \textit{Commonsense Reasoning} task from MME \citep{fu2024mmecomprehensiveevaluationbenchmark} focuses on everyday knowledge, such as whether ``sunglasses are suitable for indoor use'' or ``if an umbrella is needed on a sunny day''. Finally, CV-Bench \citep{tong2024cambrian} introduces a \textit{Relative Distance} task, challenging models to identify spatial proximity (e.g., determining which object is closest to the edge of a table). The evaluation set contains over 1500 samples.

\textbf{Temporal Reasoning} assesses a model’s ability to understand event sequences, causality, and changes that occur over time, requiring interpretation beyond static images. We evaluate this using MCQA samples from MileBench \citep{dingjie2024milebench}, where multiple images are concatenated into a single composite to preserve temporal coherence. MileBench focuses on three key aspects: (i) \textit{Counterfactual Reasoning and State Change}, which tests whether models can infer outcomes under hypothetical scenarios and track visual state transitions (e.g., ``If the door had remained closed, would the cat still be outside?”); (ii) \textit{Object and Scene Understanding}, which assesses the detection of objects, movements, and interactions across frames (e.g., ``Which object disappeared from the table between the second and third frame?”); and (iii) \textit{Action Understanding and Prediction}, which evaluates the ability to recognize and anticipate actions in a sequence (e.g., ``After the person picks up the umbrella, what is the most likely next action?”). The evaluation set consists of approximately 2500 samples.

\textbf{Coreference Resolution} evaluates a model’s ability to accurately link pronouns to their correct referents within a visual context, ensuring coherent interpretation of captions. This skill is crucial for understanding references to specific objects, individuals, or entire scenes in images. We assess this ability using caption/foil classification based on the \textit{Coreference} task from VALSE, which addresses two primary cases: (i) pronouns referring to nouns or noun phrases grounded in the image (e.g., ``A girl is reading a book. Is she sitting on a chair?”), and (ii) pronouns referring to the broader scene or a specific region (e.g., ``Is this an indoor setting?”). 
We include over $800$ samples in the evaluation set.

\subsection{Example Selection}
\label{example_s}
Each task in \SuperFewBenchShort includes a query set and a demonstration set. While the query set contains image–text pairs the model must solve, the demonstration set provides few-shot exemplars intended to support ICL. Effective example selection is non-trivial: it requires maximizing task coverage while minimizing redundancy. Traditional clustering approaches often fail to achieve both.

To address this, we adopt a submodular selection strategy based on a Graph Cut function:
\[
f(X) = \sum_{i \in V \setminus X,\;j \in X} s_{ij} - \lambda \sum_{i, j \in X} s_{ij}
\]
where $X \subseteq V$ is the selected subset, $s_{ij}$ denotes similarity between samples $i$ and $j$, and $\lambda$ controls the diversity–redundancy trade-off. The first term favors diverse, well-connected examples, while the second penalizes internal redundancy.

We extract visual and textual features using a CLIP encoder and concatenate them to form joint embeddings. These are fed into the Graph Cut function to select 250 representative query samples per task. We set $\lambda = 50$ for most tasks, and $\lambda = 200$ for larger task pools (e.g., Attribute Recognition, Spatial Relations) to ensure adequate coverage.

For each query, we then retrieve its 8 most similar samples (via cosine similarity) from the remaining pool to form the demonstration set. This ensures contextual relevance and supports localized generalization during few-shot evaluation.

\subsection{\CoTLong Description Generation}
\label{cot_d_g}
To study the effect of step-by-step reasoning, we augment few-shot demonstrations with CoT rationales. Since our source datasets lack such annotations, we generate them using \textit{Qwen2.5-VL-7B-Instruct}~\citep{Qwen2.5-VL}. The model is prompted to output both an answer and a textual explanation.

If the predicted answer is incorrect or the rationale is incomplete, we inject the correct label and regenerate the CoT trace. This iterative correction process ensures that the final CoT exemplars are faithful to the ground truth and maintain high factual quality. A subset of these rationales was further validated by humans (see Appendix~\ref{appd_data_statistics}).

\section{Experimental Setup}
\subsection{Models}

We evaluate \SuperFewBenchShort using 26 open-weight MLLMs spanning six major model families: \textbf{Qwen}, \textbf{InternVL}, \textbf{Idefics}, \textbf{Phi}, \textbf{LLaVA}, and \textbf{xGen-MM}. These models vary in size from 0.5B to 9B parameters and include both instruction-tuned and non-instruction-tuned variants.

Specifically, our benchmark includes: Qwen2.5-VL~\citep{Qwen2.5-VL}, Qwen2-VL~\citep{Qwen2-VL}, InternVL3~\citep{zhu2025internvl3exploringadvancedtraining}, InternVL2.5-MPO~\citep{wang2024enhancingreasoningabilitymultimodal}, InternVL2.5~\citep{chen2024expanding}, Idefics3~\citep{laurencon2024building}, Idefics2~\citep{NEURIPS2024_a0303731}, Phi-4~\citep{microsoft2025phi4minitechnicalreportcompact}, Phi-3~\citep{abdin2024phi3technicalreporthighly}, LLaVA-Interleave~\citep{li2025llavanextinterleave}, and xGen-MM~\citep{xue2024xgenmmblip3familyopen}. See Appendix~\ref{appd_models} for a detailed overview of models.

\begin{figure*}[t!]
    \centering
    \includegraphics[width=\linewidth]{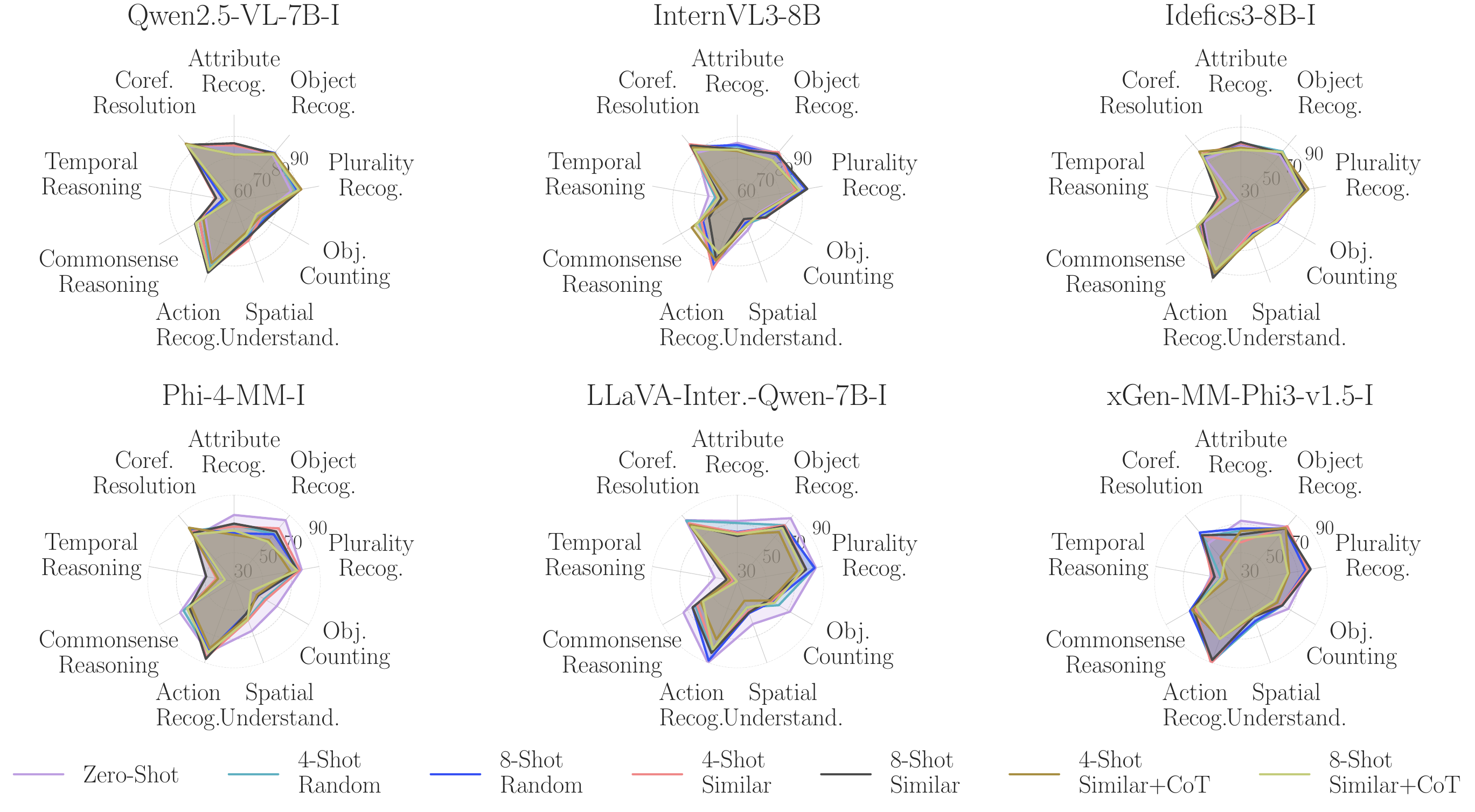}
    \caption{Accuracy performance of top-performing MLLMs from each model family on \textbf{\benchmark}, evaluated with 0-4-8 shots across three settings: Random, Similar, and Similar with CoT settings.}
    \label{fig:main_results}
\end{figure*}

\begin{figure*}[htbp]
    \centering
    \includegraphics[width=\linewidth]{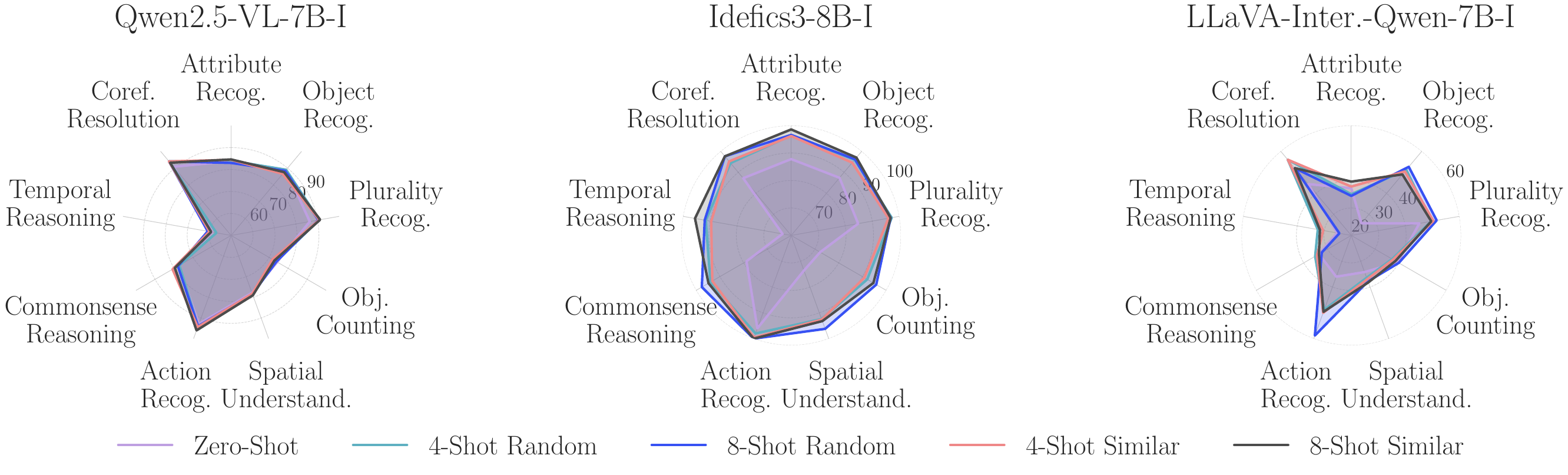}
    \caption{Pairwise accuracy performance of the top-performing model from three model families on \textbf{\benchmark}, evaluated with 0, 4, and 8 shots across two settings: Random and Similar.}
    \label{fig:perp_results}
\end{figure*}

\subsection{Evaluation Metrics}
We adopt a multiple-choice evaluation format across all tasks, with each question offering 2-4 candidate answers. The primary metric is \textbf{accuracy}, computed as the proportion of correctly answered questions based on ground-truth labels.

To ensure consistency in answer extraction, especially across models with varying output styles, we apply regex-based post-processing to normalize model predictions. While efficient and scalable, this approach can occasionally misinterpret open-ended or unexpected responses. To address this limitation, we additionally implement a \textbf{perplexity-based pairwise ranking} method: for each image–text query, we compute perplexity scores over all candidate options and rank them to determine the most likely choice. This alternative approach reduces dependency on fixed output formats and mitigates biases stemming from answer order.

We employ three prompt templates, tailored to task formats:
\begin{itemize}[leftmargin=*]
    \item For \textbf{MCQA tasks}, each prompt is designed to directly test the target skill in a structured question-answering format.
    \item For \textbf{caption/foil classification tasks}, the model is shown an image and two textual descriptions, one correct and one incorrect, and asked to select the caption that best matches the image content.
    \item For samples from the \textbf{VSR} dataset~\citep{liu-etal-2023-visual}, which contain a single caption-image pair labeled as true or false, we adapt the prompt to a binary verification format: “Does this sentence accurately describe the image?”
\end{itemize}

\section{Results and Findings}
\begin{figure*}[htbp]
    \centering
    \includegraphics[width=\linewidth]{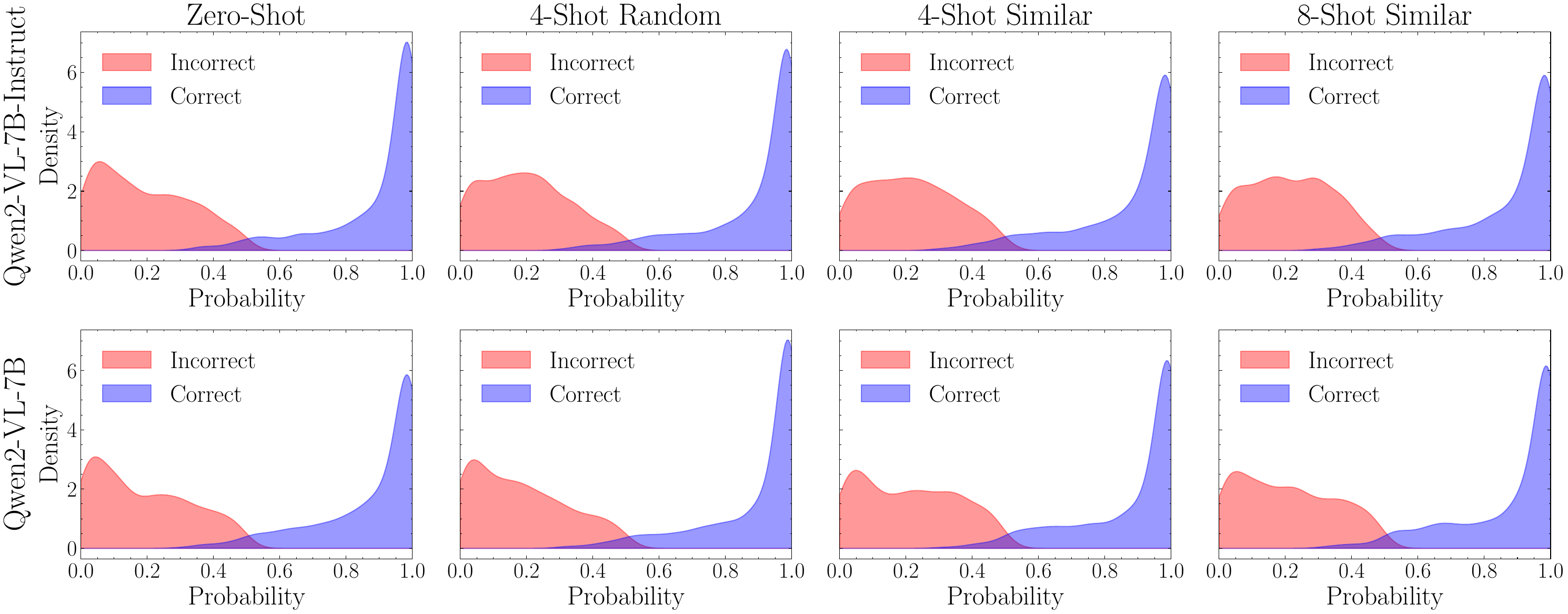}
    \caption{Uncertainty quantification on \textbf{\benchmark} for Qwen model family, showing probability distributions over correct (blue) and incorrect (red) answers under 0, 4, and 8-shot prompting with Random and Similar settings.}
    \label{fig:uncertainity_quantification}
\end{figure*}

We present key insights\footnote{Additional analyses and qualitative examples are provided in Appendix~\ref{appd_additional_analysis}, and Appendix~\ref{appd_qualitative_examples}.} from our evaluation of six state-of-the-art MLLM families on \benchmark. Results are summarized in Figures~\ref{fig:main_results} and~\ref{fig:perp_results}, covering zero-shot, 4-shot, and 8-shot settings. We analyze model behavior under both randomly sampled and retrieval-based demonstrations, as well as under CoT prompting. We conducted an uncertainty analysis to assess the model’s answer probabilities; results are shown in Figure~\ref{fig:uncertainity_quantification}.

\vspace{0.5em} \noindent \ovalbox{\begin{minipage}{0.95\linewidth}
\emph{Finding 1.} Few-shot learning yields limited gains for instruction-tuned MLLMs.
\end{minipage}}\vspace{0.25em}
\noindent Instruction-tuned models such as Qwen2.5-VL-7B (I), Phi-4-Multimodal (I), and InternVL2.5-8B-MPO achieve strong performance in the zero-shot setting across a broad range of tasks, confirming their robust generalization capabilities. However, few-shot demonstrations, whether randomly sampled or retrieved, produce only marginal improvements, often within experimental variance or even leading to slight performance drops. This suggests that instruction tuning already equips these models with effective strategies for interpreting and solving tasks, leaving little room for further benefit from in-context examples. In contrast, base models without instruction tuning exhibit significantly larger gains under few-shot prompting, with some showing double-digit improvements. This highlights that the models which benefit most from few-shot examples are those lacking prior task-specific adaptation.\vspace{0.3em}

\vspace{0.3em} \noindent \ovalbox{\begin{minipage}{0.95\linewidth}
\emph{Finding 2.} Retrieval-based demonstrations show no consistent advantage over random selection.
\end{minipage}}\vspace{0.25em}

\noindent A core question in in-context learning is whether task-relevant demonstrations can enhance performance relative to randomly selected ones. Our results show that retrieval-based prompting provides no clear advantage: for nearly all models and tasks, model performance with retrieved examples is nearly identical to that with random ones. This suggests that current MLLMs may either fail to leverage fine-grained distributional cues in retrieved examples, or that their generation is primarily driven by the query rather than contextual demonstrations. These findings challenge common practices in few-shot evaluation and motivate the development of more model-sensitive retrieval and demonstration curation techniques.\vspace{0.3em}

\vspace{0.3em} \noindent \ovalbox{\begin{minipage}{0.95\linewidth}
\emph{Finding 3.} More demonstrations do not guarantee better performance.
\end{minipage}}\vspace{0.25em}

\noindent We compare the effect of using 4-shot versus 8-shot support sets and find no consistent gains from increasing the number of demonstrations. In several cases, accuracy even declines with more examples. This may be due to \emph{context dilution}, where additional demonstrations introduce noise or distract from task-relevant information. It may also reflect the saturation of useful signal for already well-performing models. These findings suggest that optimal context length in multimodal prompting is task- and model-dependent, and that more demonstrations are not always better.\vspace{0.3em}

\vspace{0.3em} \noindent \ovalbox{\begin{minipage}{0.95\linewidth}
\emph{Finding 4.} Chain-of-Thought prompting degrades performance for all tested models. 
\end{minipage}}\vspace{0.25em}

\noindent Surprisingly, CoT prompting, while effective in many text-only LLM benchmarks, consistently leads to worse performance across all tested MLLMs and task types in \benchmark. Both instruction-tuned and base models underperform when prompted to reason step-by-step. We hypothesize that, for visually grounded tasks, CoT rationales may encourage unnecessary elaboration, introduce irrelevant or hallucinated content, or distract from the core visual grounding. This result exposes a mismatch between current CoT techniques and the needs of multimodal reasoning, pointing to the need for modality-aware CoT prompting. We analyze these failure modes further in Appendix~\ref{appd_why_cot_fails}.\vspace{0.5em}

\section{Conclusion}
We presented \benchmark, a benchmark for evaluating few-shot learning in MLLMs. Through systematic experiments across six model families, we identified key limitations of current MLLMs in leveraging in-context examples, including the limited gains from few-shot prompting, the ineffectiveness of retrieval-based selection, and the consistent performance drop with CoT reasoning. \benchmark provides a focused testbed for diagnosing and improving multimodal generalization under minimal supervision. The limitations of \benchmark are discussed in Appendix \ref{sec:limitations}.

\section*{Acknowledgments}
We sincerely thank Büşra Elif Kıvrak Doğan, Gülşah Yıldız, and Sümeyye Narin for their valuable efforts in validating \benchmark. Their support and careful feedback significantly contributed to the quality of this work. We also gratefully acknowledge the KUIS AI Center for providing the computational support that made this study possible. IC is funded by the project CaRe-NLP with file number NGF.1607.22.014 of the research programme AiNed Fellowship Grants which is (partly) financed by the Dutch Research Council (NWO).

\bibliography{custom}
\clearpage
\appendix
\section*{Appendix}
This appendix provides additional details and analyses to complement the main text. \textbf{Appendix~\ref{appd_models}} lists the MLLMs evaluated in our study. \textbf{Appendix~\ref{appd_data_statistics}} presents statistical insights into the composition and distribution of tasks in the FewMMBench benchmark. \textbf{Appendix  \ref{appd_additional_analysis}} offers an in-depth analysis of each task category, highlighting specific challenges and model behaviors. In \textbf{Appendix \ref{appd_why_cot_fails}}, we explore the underlying reasons why CoT prompting fails in certain scenarios within our benchmark. Finally, \textbf{Appendix \ref{appd_qualitative_examples}} showcases qualitative examples, including representative prompts and model predictions, to illustrate key findings and model limitations observed in our evaluation.

\section{Models}
\label{appd_models}
Here, we describe the models used in our benchmark along with their implementation details. We also present a comparative table (Table \ref{tab:model_overview}) that summarizes model architectures, configurations, and corresponding performance metrics.

\subsection{Evaluated MLLMs}
\begin{table*}[ht]
\centering
\caption{\textbf{Overview of MLLMs evaluated on \benchmark}. Models are summarised by architecture, fusion mechanism, and training regimen, with performance reported under zero‑shot and three few‑shot prompting conditions.}
\label{tab:model_overview}
\begin{adjustbox}{width=\textwidth}
\begin{threeparttable}
\begin{tabular}{@{}lcccccccccccc@{}}
\toprule
\multirow{2}{*}{\textbf{Model Name}} &
\multicolumn{4}{c}{\textbf{Model}} &
\multicolumn{2}{c}{\textbf{Training \& Tuning}} &
\multicolumn{5}{c}{\textbf{Evaluation}} \\
\cmidrule{2-12}
& \textbf{Params} & \textbf{Vision Enc.} & \textbf{Fusion} & \textbf{Text Dec.}
& \textbf{Pretrain Data} & \textbf{Instr. Tuned}
& \textbf{0‑Shot} & \textbf{4‑Rand} & \textbf{4‑Sim} & \textbf{4‑Sim+CoT} & \textbf{8‑Sim+CoT} \\
\midrule
Qwen2.5-VL & 3B & Custom ViT & Projection-Based & Qwen2.5 & Custom ($1.2$ T Token)& Yes & 75.27 & 78.98 & 79.07 & 78.42 & 73.98 \\
Qwen2.5-VL & 7B & Custom ViT & Projection-Based & Qwen2.5 & Custom ($1.2$ T Token)& Yes & 81.27 & 82.67 & 83.93 & 84.51 & 82.13 \\
Qwen2-VL & 2B & Custom ViT & Projection-Based & Qwen LM & Custom ($0.6$ T Token)& No & 18.4 & 62.18 & 63.53 & 61.2 & 54.16 \\
Qwen2-VL & 2B & Custom ViT & Projection-Based & Qwen LM & Custom ($0.6$ T Token)& Yes & 65.53 & 68.33 & 64.82 & 64.33 & 65.22 \\
Qwen2-VL & 7B & Custom ViT & Projection-Based & Qwen LM & Custom ($0.6$ T Token)& No & 36.18 & 79.11 & 77.89 & 78.51 & 77.6 \\
Qwen2-VL & 2B & Custom ViT & Projection-Based & Qwen LM & Custom ($0.6$ T Token) & Yes & 75.98 & 80.22 & 80.27 & 79.98 & 78.2 \\
InternVL3-1B & 1B & Intern ViT & Projection-Based & Qwen2.5 & Custom ($0.2$ T Token) & Yes & 40.58 & 44.6 & 44.82 & 46.29 & 40.56 \\
InternVL3-1B & 1B & Intern ViT & Projection-Based & Qwen2.5 & Custom ($0.2$ T Token & Yes & 39.29 & 45.82 & 43.4 & 43.42 & 35.22 \\
InternVL3-2B & 2B & Intern ViT & Projection-Based & Qwen2.5 & Custom ($0.2$ T Token & Yes & 68.56 & 68.56 & 69.84 & 68.87 & 64.67 \\
InternVL3-2B & 2B & Intern ViT & Projection-Based & Qwen2.5 & Custom ($0.2$ T Token & Yes & 62.96 & 67.93 & 66.67 & 66.6 & 57.6 \\
InternVL3-8B & 8B & Intern ViT & Projection-Based & Qwen2.5 & Custom ($0.2$ T Token & Yes & 82.62 & 82.51 & 82.38 & 81.6 & 81.09 \\
InternVL3-8B & 8B & Intern ViT & Projection-Based & Qwen2.5 & Custom ($0.2$ T Token & Yes & 72.58 & 79.91 & 79.89 & 78.98 & 75.96 \\
InternVL3-9B & 9B & Intern ViT & Projection-Based & Qwen2.5 & Custom ($0.2$ T Token & Yes & 79.42 & 77.78 & 77.82 & 77.13 & 77.04 \\
InternVL3-9B & 9B & Intern ViT & Projection-Based & Qwen2.5 & Custom ($0.2$ T Token & Yes & 76.91 & 75.64 & 74.18 & 74.33 & 73.84 \\
InternVL2.5-MPO & 4B & Intern ViT & Projection-Based & Qwen2.5-3B Instruct & MMDU \& OmniCorpus \& InstVLM-SA-1B-Caption & Yes & 77.4 & 74.02 & 72.91 & 71.0 & 52.38 \\
InternVL2.5-MPO & 8B & Intern ViT & Projection-Based & Qwen2.5-3B Instruct & MMDU \& OmniCorpus \& InstVLM-SA-1B-Caption & Yes & 81.62 & 79.8 & 78.96 & 77.73 & 76.69 \\
InternVL2.5 & 4B & Intern ViT & Projection-Based & Qwen2.5-3B Instruct & MMDU \& OmniCorpus \& InstVLM-SA-1B-Caption & Yes & 73.47 & 73.24 & 72.22 & 70.09 & 45.96 \\
InternVL2.5 & 4B & Intern ViT & Projection-Based & Qwen2.5-3B Instruct & MMDU \& OmniCorpus \& InstVLM-SA-1B-Caption & Yes & 76.91 & 77.27 & 74.89 & 75.71 & 69.53 \\
Idefics3 & 8B & SigLIP-SO400M & Perceiver Resampler & Llama 3.1 Instruct & OBELICS & Yes & 68.36 & 72.22 & 72.29 & 72.44 & 71.09 \\
Idefics2 & 8B & SigLIP-SO400M & Perceiver Resampler & Mistral-7B-v0.1 & OBELICS  & Yes & 67.76 & 68.33 & 68.73 & 69.04 & 65.76 \\
Phi-4-Multimodal & 5.57B & Finetuned SigLIP-400M & LoRA-Based Modular Fusion & Phi-4-Mini & Custom ($0.5$ T Token) & Yes & 71.87 & 66.62 & 67.36 & 66.62 & 62.98 \\
Phi-3.5-Vision-Instruct & 4.15B & CLIP ViT-L/14 & Token Concatenation & Phi-3.5-Mini & OBELICS  & Yes & 63.07 & 59.33 & 59.84 & 58.51 & 53.78 \\
LLaVA-NeXT-Interleave & 8.14B & SigLIP-400M & Projection Based & Qwen1.5-7B-Chat & M4-Instruct  & Yes & 74.76 & 68.04 & 65.67 & 66.22 & 63.73 \\
LLaVA-NeXT-Interleave & 0.864B & SigLIP-400M & Projection Based & Qwen1.5-7B-Chat & M4-Instruct  & Yes & 51.0 & 46.2 & 44.16 & 43.6 & 35.29 \\
xGen-MM-Mini-Instruct-v1.5 & 4.36B & SigLIP & Perceiver Resampler & Phi-3 & MINT-1T  \& OBELICS  & Yes & 70.0 & 68.82 & 67.91 & 68.42 & 59.51 \\
xGen-MM-Mini-Base-v1.5 & 4.36B & SigLIP & Perceiver Resampler & Phi-3 & MINT-1T \& OBELICS & No & 15.84 & 38.44 & 36.82 & 36.53 & 30.31 \\
\bottomrule
\end{tabular}
\begin{tablenotes}
\item \textbf{Vision Encoders:} SigLIP-SO400M~\citep{zhai2023sigmoid}, SigLIP-400M~\citep{zhai2023sigmoid}, CLIP ViT-L/14~\citep{radford2021learning}
\item \textbf{Text Decoders:} Qwen~\citep{bai2023qwentechnicalreport}, Qwen1.5~\citep{bai2023qwentechnicalreport}, Qwen2.5~\citep{qwen2025qwen25technicalreport}, Llama 3.1~\citep{dubey2024llama}, Phi-3~\citep{abdin2024phi3technicalreporthighly}
\item \textbf{Pretraining datasets:} OBELICS~\citep{laurenccon2023obelics}, M4-Instruct~\citep{li2025llavanextinterleave}, MINT-1T~\citep{driess2023palm}, MMDU~\citep{liu2024mmdu} OmniCorpus~\citep{li2025omnicorpus} \& InstVLM-SA-1B-Caption~\citep{chen2023internvl}
\end{tablenotes}
\end{threeparttable}
\end{adjustbox}
\end{table*}

\paragraph{\textbf{Qwen Family.}} 
We evaluate two models from the Qwen family: Qwen2-VL \citep{Qwen2-VL} and Qwen2.5-VL \citep{Qwen2.5-VL}, ranging from 2B to 7B parameters. Qwen2-VL introduced Naive Dynamic Resolution and Multimodal Rotary Position Embedding to improve visual tokenization and cross-modal alignment, showing strong performance across vision-language tasks. Qwen2.5-VL builds on these advances with a native dynamic-resolution ViT, Window Attention, and absolute time encoding, enabling precise object localization, robust document parsing, and long-video understanding. It offers better efficiency and broader capabilities, making it more effective than Qwen2-VL in both visual and linguistic tasks.
\paragraph{\textbf{InternVL Family.}}
We include InternVL2.5 \citep{chen2024expanding}, InternVL2.5-MPO \citep{wang2024enhancingreasoningabilitymultimodal}, and InternVL3 \citep{zhu2025internvl3exploringadvancedtraining} in our evaluation. InternVL2.5 builds upon InternVL2.0 with improved training strategies and higher-quality data, achieving strong results across diverse multimodal tasks such as document understanding, video comprehension, and multilingual reasoning. It demonstrates robust Chain-of-Thought (CoT) performance and competitive accuracy against leading commercial models.

InternVL2.5-MPO enhances this further through Mixed Preference Optimization, a method that mitigates distribution shifts in CoT reasoning by optimizing on a large-scale preference dataset. This leads to substantial improvements in multimodal reasoning, outperforming its base counterpart and rivaling much larger models. In our experiments, InternVL2.5-MPO consistently outperforms InternVL2.5, making it the stronger variant within the InternVL2.5 family.

InternVL3 introduces a unified multimodal pre-training paradigm, jointly learning from multimodal and text-only corpora rather than relying on post-hoc adaptation. It incorporates Variable Visual Position Encoding (V2PE), advanced fine-tuning strategies including MPO and SFT, and test-time scaling to enhance reasoning and generalization. In our experiments on \benchmark, InternVL3 outperforms both InternVL2.5 and InternVL2.5-MPO, establishing itself as the most capable model in the InternVL family.

\paragraph{\textbf{Idefics Family.}}
We evaluate Idefics2 \citep{NEURIPS2024_a0303731} and Idefics3 \citep{laurencon2024building}, both 8B-parameter vision-language models. Idefics2 was developed through extensive empirical analysis of architecture, data, and training strategies, resulting in a highly efficient VLM that achieves competitive performance within its size class, often matching much larger models.

Idefics3 improves upon its predecessor with a more streamlined and principled training pipeline, leveraging only open datasets and introducing Docmatix, a large-scale dataset tailored for document understanding. Thanks to these enhancements, Idefics3 achieves notably stronger results than Idefics2 across a range of multimodal tasks, particularly in structured document reasoning, making it the superior model in this family.

\paragraph{\textbf{Phi Family.}}
Our evaluation includes Phi-3.5-Vision-Instruct \citep{abdin2024phi3technicalreporthighly} and Phi-4-Multimodal \citep{microsoft2025phi4minitechnicalreportcompact}, both based on compact yet capable language models. Phi-3.5-Vision-Instruct, derived from Phi-3.5-Mini, is a 4.2B parameter model optimized for vision-language tasks, capable of handling both single and multi-image inputs. It demonstrates strong reasoning and language capabilities within a small footprint, benefiting from high-quality filtered web and synthetic data.

Phi-4-Multimodal builds upon this with notable improvements in both reasoning and multimodal flexibility. It introduces a modality extension mechanism using LoRA adapters and modality-specific routers, allowing it to handle combinations of text, vision, and audio without interference. Despite its compact size (3.8B), it outperforms larger models across vision-language and speech-language benchmarks. In our benchmark, Phi-4-Multimodal consistently surpasses Phi-3.5-Vision-Instruct, particularly in reasoning-intensive and multi-modal integration tasks.
\paragraph{\textbf{LLaVA Family.}}
We evaluate both the 0.5B and 7B versions of LLaVA-NeXT-Interleave \citep{li2025llavanextinterleave}, a model series designed to unify visual instruction tuning across diverse multi-image scenarios. LLaVA-NeXT-Interleave introduces a general-purpose interleaved data format and the large-scale M4-Instruct dataset, enabling the model to handle tasks involving multi-image, video (multi-frame), 3D (multi-view), and multi-patch inputs within a unified framework.

The 7B version consistently outperforms the 0.5B variant across all task types, demonstrating stronger cross-modal reasoning and better generalization across settings. Notably, it maintains solid performance on single-image tasks while excelling in complex multi-image and temporal scenarios, making it the more capable model in our benchmark evaluation.
\paragraph{\textbf{xGen Family.}}
We evaluate two 4.36B parameter models from the xGen-MM \citep{xue2024xgenmmblip3familyopen} (also known as BLIP-3) family: xGen-MM-Mini-Base-v1.5 and xGen-MM-Mini-Instruct-v1.5. Built under the BLIP-3 framework, these models leverage a carefully curated training recipe and large-scale datasets to support a wide range of vision-language tasks, including single and multi-image understanding with interleaved image-text inputs.

The Instruct variant, fine-tuned for visual instruction following, demonstrates improved reasoning, alignment, and task generalization. In our evaluation, it consistently outperforms the Base model, making it the stronger choice for multimodal comprehension within the xGen-MM family.
\subsection{Implementation Details}
For our experiments in \benchmark, we utilized the official model implementations available in the HuggingFace repository. We adopted half-precision (fp16) inference for the Qwen models, InternVL, Idefics2, Idefics3, and all LLaVA variants to optimize memory and speed. In addition to half-precision, we applied 8-bit quantization for InternVL models and 4-bit quantization for Idefics2 and LLaVA variants. Idefics3 was evaluated solely with half-precision. The Phi and xGen-MM models were tested without half-precision due to compatibility constraints. All experiments were conducted using either Tesla T4 or A40 GPUs, depending on the model’s resource requirements.

\section{Data Statistics}
\label{appd_data_statistics}
In this section, we provide supporting statistics for \benchmark. Section~\ref{appd_data_summary} summarizes dataset statistics. Section~\ref{appd_demo_exp_sim} analyzes the similarity between our selected demonstrations and randomly sampled examples. Finally, Section~\ref{appd_reasoning_validation} describes our procedure for validating the quality of CoT demonstrations.
\subsection{Data Summary}
\label{appd_data_summary}
\begin{table*}[ht]
\centering
\caption{A table demonstrating number of data samples from source datasets in \SuperFewBenchShort including query and demonstration sets.}
\label{tab:task_benchmark_map}
\begin{adjustbox}{width=\textwidth}
\begin{tabular}{@{}l|cccccccccccccccc@{}}
\toprule
\multirow{2}{*}{\textbf{Tasks}} & 
\multicolumn{8}{c}{\textbf{Query Set}} & 
\multicolumn{8}{c}{\textbf{Demonstration Set}}  \\
\cmidrule(lr){2-17} %
& \multicolumn{4}{c}{\textbf{Multiple Choice QA}} & \multicolumn{4}{c}{\textbf{Caption/Foil}} 
& \multicolumn{4}{c}{\textbf{Multiple Choice QA}} & \multicolumn{4}{c}{\textbf{Caption/Foil}} \\
\midrule
\multirow{2}{*}{Attribute Recognition} 
    & \multicolumn{2}{c}{MMBench}  & \multicolumn{2}{c}{SeedBench} & \multicolumn{4}{c}{ARO} 
    & \multicolumn{2}{c}{MMBench}  & \multicolumn{2}{c}{SeedBench} & \multicolumn{4}{c}{ARO} \\
    & \multicolumn{2}{c}{24}  & \multicolumn{2}{c}{226} & \multicolumn{4}{c}{250} 
    & \multicolumn{2}{c}{68}  & \multicolumn{2}{c}{1317} & \multicolumn{4}{c}{1470} \\

\midrule

\multirow{2}{*}{Visual Object Recognition} 
    & \multicolumn{2}{c}{SeedBench}  & \multicolumn{2}{c}{GVTBench} & \multicolumn{4}{c}{VALSE} 
    & \multicolumn{2}{c}{SeedBench}  & \multicolumn{2}{c}{GVTBench} & \multicolumn{4}{c}{VALSE} \\
    & \multicolumn{2}{c}{171}  & \multicolumn{2}{c}{79} & \multicolumn{4}{c}{250} 
    & \multicolumn{2}{c}{557}  & \multicolumn{2}{c}{1054} & \multicolumn{4}{c}{251} \\

\midrule

\multirow{2}{*}{Plurality Recognition} 
    & \multicolumn{4}{c}{-}  &  \multicolumn{4}{c}{VALSE} & \multicolumn{4}{c}{-}  &  \multicolumn{4}{c}{VALSE} \\
    & \multicolumn{4}{c}{-}  & \multicolumn{4}{c}{250} & \multicolumn{4}{c}{-} 
    & \multicolumn{4}{c}{529}  \\

\midrule

\multirow{2}{*}{Object Counting} 
    & \multicolumn{4}{c}{SeedBench}  &  \multicolumn{4}{c}{VALSE} & \multicolumn{4}{c}{SeedBench}  &  \multicolumn{4}{c}{VALSE} \\
    & \multicolumn{4}{c}{250}  & \multicolumn{4}{c}{250} & \multicolumn{4}{c}{1064} 
    & \multicolumn{4}{c}{996}  \\

\midrule

\multirow{2}{*}{Spatial Relations Understanding} 
    & \multicolumn{2}{c}{MMBench}  & \multicolumn{2}{c}{SeedBench} & VALSE & VSR & ARO & \multicolumn{1}{c}{What'sUp}
    & \multicolumn{2}{c}{MMBench}  & \multicolumn{2}{c}{SeedBench} & VALSE & VSR & ARO & \multicolumn{1}{c}{What'sUp} \\
    & \multicolumn{2}{c}{64}  & \multicolumn{2}{c}{186} & 106 & 14 & 102 & \multicolumn{1}{c}{28} 
    & \multicolumn{2}{c}{82}  & \multicolumn{2}{c}{395} & 114 & 122 & 1122 & \multicolumn{1}{c}{287} \\

\midrule

\multirow{2}{*}{Action Recognition} 
    & \multicolumn{4}{c}{-}  & \multicolumn{4}{c}{VALSE} &   \multicolumn{4}{c}{-}  & \multicolumn{4}{c}{VALSE} \\
    & \multicolumn{4}{c}{-}  & \multicolumn{4}{c}{250} & \multicolumn{4}{c}{-}  & \multicolumn{4}{c}{741}  \\

\midrule

\multirow{2}{*}{Commonsense Reasoning} 
    & MMBench & SeedBench & MME & \multicolumn{1}{c}{CV-Bench} & \multicolumn{4}{c}{-} & MMBench & SeedBench & MME & \multicolumn{1}{c}{CV-Bench} & \multicolumn{4}{c}{-} \\
    & 46 & 122 & 25 & \multicolumn{1}{c}{57} & \multicolumn{4}{c}{-} & 191 & 172 & 73 & \multicolumn{1}{c}{234} & \multicolumn{4}{c}{-}
     \\

\midrule

\multirow{2}{*}{Temporal Reasoning} 
    & \multicolumn{4}{c}{MileBench}  &  \multicolumn{4}{c}{-} & \multicolumn{4}{c}{MileBench}  &  \multicolumn{4}{c}{-} \\
    & \multicolumn{4}{c}{250}  & \multicolumn{4}{c}{-} & \multicolumn{4}{c}{856} 
    & \multicolumn{4}{c}{-}  \\

\midrule

\multirow{2}{*}{Coreference Resolution} 
    & \multicolumn{4}{c}{-}  &  \multicolumn{4}{c}{VALSE} & \multicolumn{4}{c}{-}  &  \multicolumn{4}{c}{VALSE} \\
    & \multicolumn{4}{c}{-}  & \multicolumn{4}{c}{250} & \multicolumn{4}{c}{-} 
    & \multicolumn{4}{c}{558}  \\
\bottomrule
\end{tabular}
\label{tab:data_sample_statistics}
\end{adjustbox}
\end{table*}

\benchmark comprises $3,250$ carefully curated data samples spanning 9 distinct multimodal reasoning tasks. These samples were selected from a large candidate pool to ensure diversity, representativeness, and non-redundancy. For each selected sample, we identified the most semantically similar examples from the remaining pool to construct the demonstration set, resulting in over $12,000$ samples. Each of these is accompanied by manually crafted reasoning chains to facilitate in-context learning analysis. In addition, we augment the demonstration set with randomly sampled instances; however, these do not include reasoning chains. Table \ref{tab:data_sample_statistics} reports the distribution of selected samples across tasks with respect to their source datasets (excluding random samples). Note that some tasks, Caption/Foil Visual Object Recognition and Plurality Recognition, contain fewer samples due to limited availability in the corresponding parent datasets. All data used in \benchmark is derived from publicly available datasets with permissible licenses, including Apache 2.0 (MMBench, SeedBench, GVT-Bench, VSR, MME, CV-Bench), MIT (ARO, VALSE, What’s Up), and Creative Commons Attribution 2.0 (MileBench).

\subsection{Demonstration Example Similarity}
\label{appd_demo_exp_sim}
\begin{figure*}[t!]
    \centering
    \includegraphics[width=\linewidth]{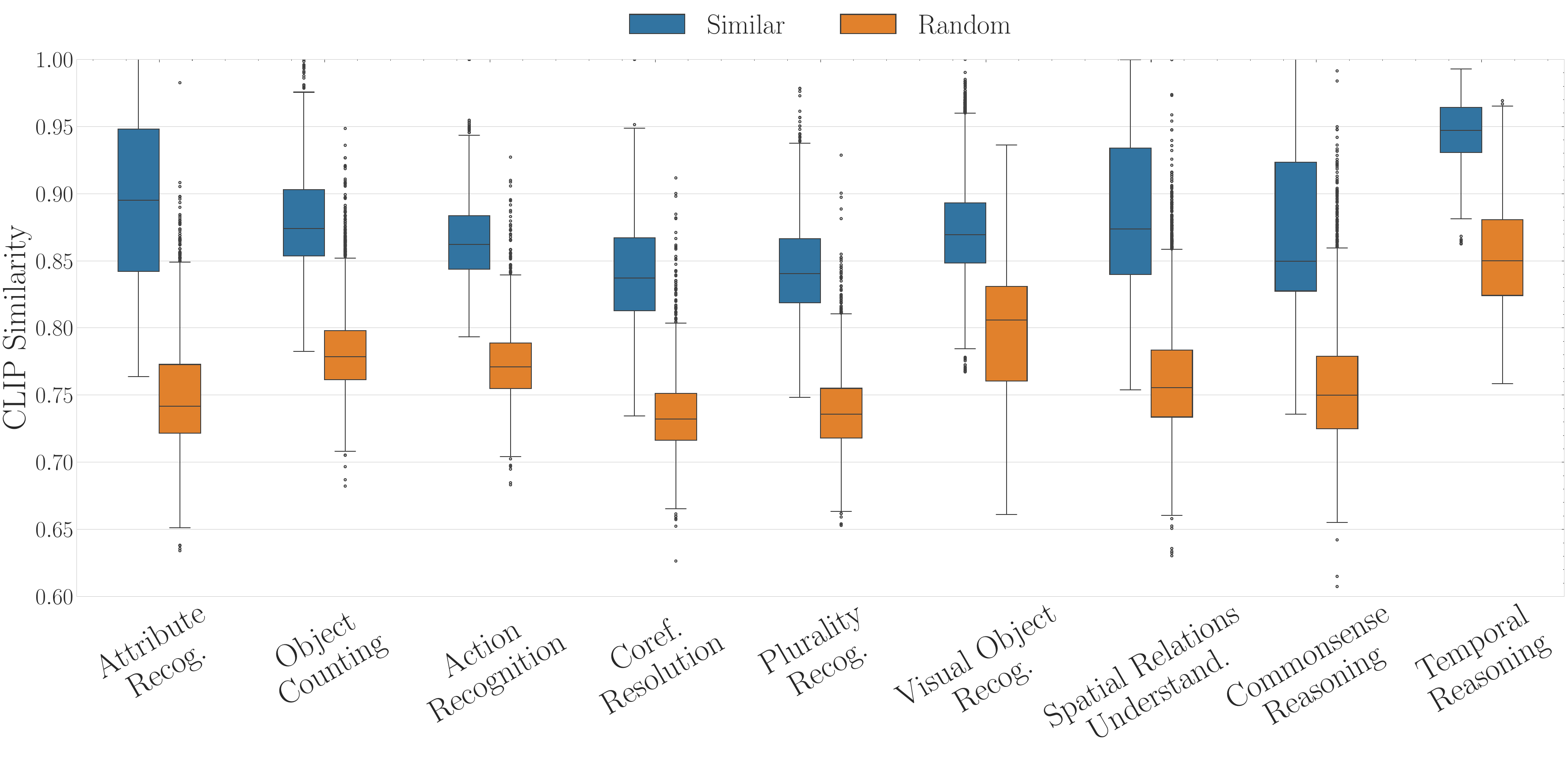}
    \caption{Distribution of semantic similarity scores between query and demonstration examples across tasks in the \benchmark}
    \label{fig:similarity}
\end{figure*}

In Figure \ref{fig:similarity}, we present a box plot of CLIP-based image-text similarity between query and demonstration examples across tasks. Each box compares the Similar-Shot setting (blue) versus Random-Shot setting (orange). Similar-Shot demonstrations are selected based on high semantic similarity to the query; Random-Shot ones are sampled uniformly. Tasks are sorted alphabetically. Across all tasks, Similar-Shot examples consistently exhibit higher similarity scores, indicating effective retrieval. This validates our shot selection method and supports subsequent analyses of in-context learning behavior.

\subsection{Reasoning Chain Validation}
\label{appd_reasoning_validation}
\begin{figure}[ht!]
    \centering
    \includegraphics[width=\columnwidth]{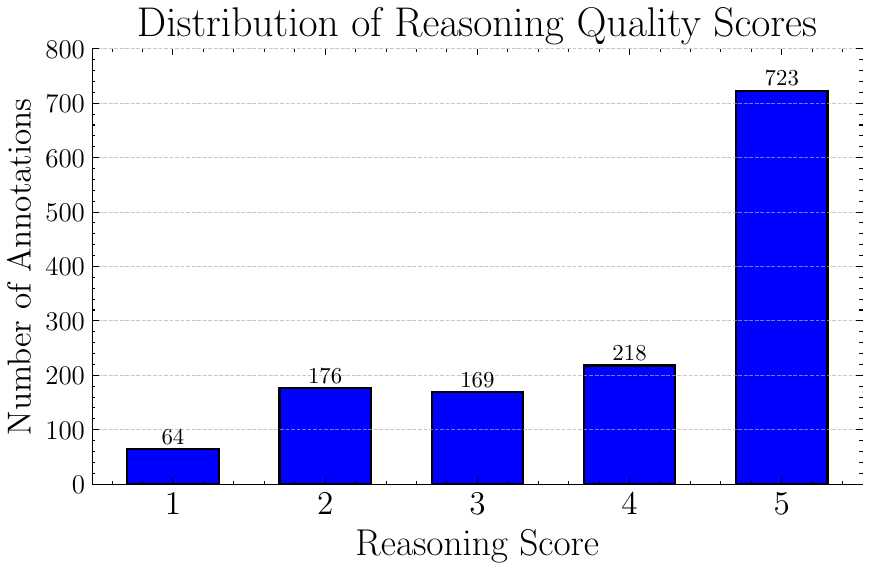}
    \caption{Distribution of reasoning quality scores assigned by human annotators to Chain-of-Thought demonstrations.}
    \label{fig:cot_validation}
\end{figure}

To evaluate the quality of our Chain-of-Thought demonstrations, we randomly sampled $450$ examples ($50$ per task) from a pool of $12,000$. Each example included an image, a question, a model-generated reasoning chain, and a final answer. Three annotators independently assessed each example along two axes: (1) Answer correctness (\textit{True}, \textit{False}, or \textit{Other}), and (2) Reasoning quality, rated on a 5-point Likert scale (1 = incorrect/incoherent, 5 = fully correct and well-structured).

We report majority vote for correctness and mean reasoning score per example. Results show that $82.2\%$ of examples produced correct final answers. The mean reasoning quality was $4.01\pm1.05$, with $66.2\%$ scoring $\ge 4$. Inter-annotator agreement was moderate (Fleiss’ $K = 0.60$ for correctness; Krippendorff’s $\alpha = 0.49$ for reasoning) \citep{krippendorff1989content, fleiss1971measuring}. Figure \ref{fig:cot_validation} presents the full distribution of reasoning scores. These results indicate that demonstration quality is generally high, suggesting that performance limitations under CoT prompting stem from model behavior rather than prompt quality.

All three annotators were volunteers and were not compensated for their participation. They were informed that their annotations would be used to evaluate the quality of the benchmark. All annotators are from Turkey; one currently resides in the U.S., and another recently returned from the U.S. All have English proficiency at or above the C1 level.

\section{Additional Analysis}
\label{appd_additional_analysis}
In this section, we provide additional analysis on \benchmark. We begin by examining model predictions to assess the overall difficulty of the benchmark and identify common failure patterns. We then analyze the uncertainty quantification results to better understand the confidence levels associated with correct and incorrect predictions. Then, we present task-level performance results for each model, followed by an analysis of the selected similar examples used in few-shot settings, focusing on their representativeness and non-redundancy. Table \ref{tab:all_results} presents the accuracy scores of all models across individual tasks, providing an overview of their overall performance. Table \ref{tab:perp_results} reports pairwise accuracy results, highlighting the relative performance differences between model pairs on each task. The abbreviations of MLLMs used in our results table are provided in Table~\ref{tab:abbreviations}.

\begin{table}[!t]
\centering
\resizebox{0.875\linewidth}{!}{
\begin{tabular}{ll}
\toprule
\textbf{Abbreviation} & \textbf{Full Model Name}\\
\midrule
QW2.5VL-3B-I & Qwen2.5-VL-3B-Instruct\\
QW2.5VL-7B-I & Qwen2.5-VL-7B-Instruct\\
QW2VL-2B & Qwen2-VL-2B\\
QW2VL-2B-I & Qwen2-VL-2B-Instruct\\ 
QW2VL-7B & Qwen2-VL-7B\\
QW2VL-7B-I & Qwen2-VL-7B-Instruct\\ 
IVL3-1B & InternVL3-1B \\
IVL3-1B-I & InternVL3-1B-I \\ 
IVL3-2B & InternVL3-2B \\
IVL3-2B-I & InternVL3-2B-I \\ 
IVL3-8B & InternVL3-8B \\
IVL3-8B-I & InternVL3-8B-I \\ 
IVL3-9B & InternVL3-9B \\
IVL3-9B-I & InternVL3-9B-I \\
IVL2.5-4B-M & InternVL2.5-4B-MPO \\
IVL2.5-8B-M & InternVL2.5-8B-MPO \\
IVL2.5-4B & InternVL2.5-4B \\
IVL2.5-8B & InternVL2.5-8B \\
ID3-8B & Idefics3-8B\\
ID2-8B & Idefics2-8B \\
P4M-I & Phi4-Multimodal-Instruct \\
P3.5V-I& Phi3.5-Vision-Instruct \\
LLV-I-QW-7B & LLaVA-NeXT-Interleave-Qwen-7B\\
LLV-I-QW-0.5B & LLaVA-NeXT-Interleave-Qwen-0.5B\\
XG-MM-P3-I& xGen-MM-Phi3-mini-instruct \\
XG-MM-P3-B& xGen-MM-Phi3-mini-base \\
\bottomrule
\end{tabular}
}
\caption{Abbreviations of the evaluated MLLMs.}
\label{tab:abbreviations}
\end{table}

\begin{table*}[!t]
    \centering
    \renewcommand{\arraystretch}{0.97}
    \resizebox{\linewidth}{!}{
    \begin{threeparttable}
    \begin{tabular}{@{}l@{\;}c@{$\;$}c@{$\;$}c@{$\;\;\;$}c@{$\;$}c@{$\;$}c@{$\;\;\;$}c@{$\;$}c@{$\;$}c@{$\;\;\;$}c@{$\;$}c@{$\;$}c@{$\;\;\;$}c@{$\;$}c@{$\;$}c@{$\;\;\;$}c@{$\;$}c@{$\;$}c@{$\;\;\;$}c@{$\;$}c@{$\;$}c@{$\;\;\;$}c@{$\;$}c@{$\;$}c@{$\;\;\;$}c@{$\;$}c@{$\;$}c@{$\;\;\;$}c@{$\;$}c@{$\;$}c@{}}
    \toprule
    \multicolumn{31}{c}{\emph{Zero-Shot Setting}} \\
\midrule
  \textbf{Model} & \multicolumn{3}{@{}c@{}}{\textbf{{Attribute}}} & \multicolumn{3}{@{}c@{}}{\textbf{{Object}}} & \multicolumn{3}{@{}c@{}}{\textbf{{Plurality}}} & \multicolumn{3}{@{}c@{}}{\textbf{{Counting}}} & \multicolumn{3}{@{}c@{}}{\textbf{{Relations}}} & \multicolumn{3}{@{}c@{}}{\textbf{{Action}}} &  \multicolumn{3}{@{}c@{}}{\textbf{{Commonsense}}} & \multicolumn{3}{@{}c@{}}{\textbf{{Temporal}}} & \multicolumn{3}{@{}c@{}}{\textbf{{Coreference}}} & \multicolumn{3}{@{}c@{}}{\textbf{{Avg.}}}\\
\midrule      
QW2.5VL-3B-I & \multicolumn{3}{c}{80.4} & \multicolumn{3}{c}{85.0} & \multicolumn{3}{c}{84.8} & \multicolumn{3}{c}{65.8} & \multicolumn{3}{c}{67.0} & \multicolumn{3}{c}{78.8} & \multicolumn{3}{c}{66.8} & \multicolumn{3}{c}{62.0} & \multicolumn{3}{c}{86.8} & \multicolumn{3}{c}{75.3} \\
QW2.5VL-7B-I & \multicolumn{3}{c}{85.8} & \multicolumn{3}{c}{87.8} & \multicolumn{3}{c}{86.8} & \multicolumn{3}{c}{74.0} & \multicolumn{3}{c}{75.8} & \multicolumn{3}{c}{90.0} & \multicolumn{3}{c}{75.2} & \multicolumn{3}{c}{63.2} & \multicolumn{3}{c}{92.8} & \multicolumn{3}{c}{81.3} \\
QW2VL-2B & \multicolumn{3}{c}{11.6} & \multicolumn{3}{c}{31.8} & \multicolumn{3}{c}{12.0} & \multicolumn{3}{c}{26.0} & \multicolumn{3}{c}{18.6} & \multicolumn{3}{c}{4.0} & \multicolumn{3}{c}{21.2} & \multicolumn{3}{c}{22.8} & \multicolumn{3}{c}{17.6} & \multicolumn{3}{c}{18.4} \\
QW2VL-2B-I & \multicolumn{3}{c}{72.4} & \multicolumn{3}{c}{74.8} & \multicolumn{3}{c}{72.0} & \multicolumn{3}{c}{62.8} & \multicolumn{3}{c}{57.8} & \multicolumn{3}{c}{78.4} & \multicolumn{3}{c}{58.0} & \multicolumn{3}{c}{54.8} & \multicolumn{3}{c}{58.8} & \multicolumn{3}{c}{65.5} \\
QW2VL-7B & \multicolumn{3}{c}{50.0} & \multicolumn{3}{c}{62.6} & \multicolumn{3}{c}{15.6} & \multicolumn{3}{c}{59.2} & \multicolumn{3}{c}{39.4} & \multicolumn{3}{c}{10.4} & \multicolumn{3}{c}{31.6} & \multicolumn{3}{c}{47.6} & \multicolumn{3}{c}{9.2} & \multicolumn{3}{c}{36.2} \\
QW2VL-7B-I & \multicolumn{3}{c}{79.2} & \multicolumn{3}{c}{84.2} & \multicolumn{3}{c}{84.8} & \multicolumn{3}{c}{74.0} & \multicolumn{3}{c}{72.0} & \multicolumn{3}{c}{85.6} & \multicolumn{3}{c}{66.0} & \multicolumn{3}{c}{54.8} & \multicolumn{3}{c}{83.2} & \multicolumn{3}{c}{76.0} \\
IVL3-1B & \multicolumn{3}{c}{38.8} & \multicolumn{3}{c}{51.8} & \multicolumn{3}{c}{50.0} & \multicolumn{3}{c}{34.4} & \multicolumn{3}{c}{35.4} & \multicolumn{3}{c}{58.8} & \multicolumn{3}{c}{30.8} & \multicolumn{3}{c}{28.4} & \multicolumn{3}{c}{36.8} & \multicolumn{3}{c}{40.6} \\
IVL3-1B-I & \multicolumn{3}{c}{36.0} & \multicolumn{3}{c}{49.4} & \multicolumn{3}{c}{44.8} & \multicolumn{3}{c}{31.2} & \multicolumn{3}{c}{35.4} & \multicolumn{3}{c}{52.4} & \multicolumn{3}{c}{33.2} & \multicolumn{3}{c}{30.4} & \multicolumn{3}{c}{40.8} & \multicolumn{3}{c}{39.3} \\
IVL3-2B & \multicolumn{3}{c}{80.6} & \multicolumn{3}{c}{79.4} & \multicolumn{3}{c}{78.8} & \multicolumn{3}{c}{56.2} & \multicolumn{3}{c}{60.4} & \multicolumn{3}{c}{87.6} & \multicolumn{3}{c}{52.8} & \multicolumn{3}{c}{55.2} & \multicolumn{3}{c}{66.0} & \multicolumn{3}{c}{68.6} \\
IVL3-2B-I & \multicolumn{3}{c}{68.8} & \multicolumn{3}{c}{71.4} & \multicolumn{3}{c}{76.4} & \multicolumn{3}{c}{43.4} & \multicolumn{3}{c}{55.8} & \multicolumn{3}{c}{87.2} & \multicolumn{3}{c}{48.4} & \multicolumn{3}{c}{48.4} & \multicolumn{3}{c}{66.8} & \multicolumn{3}{c}{63.0} \\
IVL3-8B & \multicolumn{3}{c}{87.0} & \multicolumn{3}{c}{89.0} & \multicolumn{3}{c}{87.2} & \multicolumn{3}{c}{71.0} & \multicolumn{3}{c}{74.2} & \multicolumn{3}{c}{91.6} & \multicolumn{3}{c}{80.8} & \multicolumn{3}{c}{73.6} & \multicolumn{3}{c}{89.2} & \multicolumn{3}{c}{82.6} \\
IVL3-8B-I & \multicolumn{3}{c}{78.8} & \multicolumn{3}{c}{85.8} & \multicolumn{3}{c}{77.2} & \multicolumn{3}{c}{63.4} & \multicolumn{3}{c}{62.0} & \multicolumn{3}{c}{84.4} & \multicolumn{3}{c}{66.0} & \multicolumn{3}{c}{56.0} & \multicolumn{3}{c}{79.6} & \multicolumn{3}{c}{72.6} \\
IVL3-9B & \multicolumn{3}{c}{84.4} & \multicolumn{3}{c}{88.2} & \multicolumn{3}{c}{82.4} & \multicolumn{3}{c}{69.4} & \multicolumn{3}{c}{70.8} & \multicolumn{3}{c}{92.0} & \multicolumn{3}{c}{74.4} & \multicolumn{3}{c}{66.0} & \multicolumn{3}{c}{87.2} & \multicolumn{3}{c}{79.4} \\
IVL3-9B-I & \multicolumn{3}{c}{79.4} & \multicolumn{3}{c}{88.4} & \multicolumn{3}{c}{83.2} & \multicolumn{3}{c}{67.8} & \multicolumn{3}{c}{65.8} & \multicolumn{3}{c}{88.4} & \multicolumn{3}{c}{70.4} & \multicolumn{3}{c}{62.0} & \multicolumn{3}{c}{86.8} & \multicolumn{3}{c}{76.9} \\
IVL2.5-4B-M & \multicolumn{3}{c}{81.0} & \multicolumn{3}{c}{85.4} & \multicolumn{3}{c}{86.0} & \multicolumn{3}{c}{62.8} & \multicolumn{3}{c}{63.4} & \multicolumn{3}{c}{90.4} & \multicolumn{3}{c}{71.2} & \multicolumn{3}{c}{64.4} & \multicolumn{3}{c}{92.0} & \multicolumn{3}{c}{77.4} \\
IVL2.5-8B-M & \multicolumn{3}{c}{87.8} & \multicolumn{3}{c}{88.8} & \multicolumn{3}{c}{78.0} & \multicolumn{3}{c}{76.8} & \multicolumn{3}{c}{73.6} & \multicolumn{3}{c}{90.8} & \multicolumn{3}{c}{77.2} & \multicolumn{3}{c}{69.6} & \multicolumn{3}{c}{92.0} & \multicolumn{3}{c}{81.6} \\
IVL2.5-4B & \multicolumn{3}{c}{78.4} & \multicolumn{3}{c}{83.4} & \multicolumn{3}{c}{80.0} & \multicolumn{3}{c}{59.4} & \multicolumn{3}{c}{64.4} & \multicolumn{3}{c}{85.6} & \multicolumn{3}{c}{64.8} & \multicolumn{3}{c}{62.8} & \multicolumn{3}{c}{82.4} & \multicolumn{3}{c}{73.5} \\
IVL2.5-8B & \multicolumn{3}{c}{83.4} & \multicolumn{3}{c}{85.4} & \multicolumn{3}{c}{78.8} & \multicolumn{3}{c}{68.6} & \multicolumn{3}{c}{68.0} & \multicolumn{3}{c}{91.2} & \multicolumn{3}{c}{68.8} & \multicolumn{3}{c}{66.0} & \multicolumn{3}{c}{82.0} & \multicolumn{3}{c}{76.9} \\
ID3-8B & \multicolumn{3}{c}{75.0} & \multicolumn{3}{c}{79.0} & \multicolumn{3}{c}{79.2} & \multicolumn{3}{c}{62.0} & \multicolumn{3}{c}{55.6} & \multicolumn{3}{c}{93.2} & \multicolumn{3}{c}{64.4} & \multicolumn{3}{c}{32.4} & \multicolumn{3}{c}{74.4} & \multicolumn{3}{c}{68.4} \\
ID2-8B & \multicolumn{3}{c}{63.6} & \multicolumn{3}{c}{81.2} & \multicolumn{3}{c}{81.2} & \multicolumn{3}{c}{57.2} & \multicolumn{3}{c}{54.2} & \multicolumn{3}{c}{87.2} & \multicolumn{3}{c}{58.4} & \multicolumn{3}{c}{42.8} & \multicolumn{3}{c}{84.0} & \multicolumn{3}{c}{67.8} \\
P4M-I & \multicolumn{3}{c}{76.2} & \multicolumn{3}{c}{85.6} & \multicolumn{3}{c}{77.6} & \multicolumn{3}{c}{64.2} & \multicolumn{3}{c}{66.4} & \multicolumn{3}{c}{82.4} & \multicolumn{3}{c}{73.2} & \multicolumn{3}{c}{50.0} & \multicolumn{3}{c}{71.2} & \multicolumn{3}{c}{71.9} \\
P3.5V-I & \multicolumn{3}{c}{65.6} & \multicolumn{3}{c}{75.0} & \multicolumn{3}{c}{69.6} & \multicolumn{3}{c}{56.8} & \multicolumn{3}{c}{53.0} & \multicolumn{3}{c}{80.0} & \multicolumn{3}{c}{54.4} & \multicolumn{3}{c}{42.0} & \multicolumn{3}{c}{71.2} & \multicolumn{3}{c}{63.1} \\
LLV-I-QW-7B & \multicolumn{3}{c}{72.0} & \multicolumn{3}{c}{87.4} & \multicolumn{3}{c}{85.2} & \multicolumn{3}{c}{72.0} & \multicolumn{3}{c}{61.4} & \multicolumn{3}{c}{90.0} & \multicolumn{3}{c}{73.2} & \multicolumn{3}{c}{46.0} & \multicolumn{3}{c}{85.6} & \multicolumn{3}{c}{74.8} \\
LLV-I-QW-0.5B & \multicolumn{3}{c}{54.0} & \multicolumn{3}{c}{62.4} & \multicolumn{3}{c}{52.8} & \multicolumn{3}{c}{43.0} & \multicolumn{3}{c}{44.0} & \multicolumn{3}{c}{67.2} & \multicolumn{3}{c}{50.0} & \multicolumn{3}{c}{37.2} & \multicolumn{3}{c}{48.4} & \multicolumn{3}{c}{51.0} \\
XG-MM-P3-I & \multicolumn{3}{c}{72.2} & \multicolumn{3}{c}{79.6} & \multicolumn{3}{c}{77.6} & \multicolumn{3}{c}{68.0} & \multicolumn{3}{c}{59.8} & \multicolumn{3}{c}{90.0} & \multicolumn{3}{c}{66.8} & \multicolumn{3}{c}{51.2} & \multicolumn{3}{c}{64.8} & \multicolumn{3}{c}{70.0} \\
XG-MM-P3-B & \multicolumn{3}{c}{15.4} & \multicolumn{3}{c}{10.2} & \multicolumn{3}{c}{16.4} & \multicolumn{3}{c}{12.0} & \multicolumn{3}{c}{9.4} & \multicolumn{3}{c}{26.8} & \multicolumn{3}{c}{14.8} & \multicolumn{3}{c}{15.2} & \multicolumn{3}{c}{22.4} & \multicolumn{3}{c}{15.8} \\

\midrule
       \multicolumn{31}{c}{\emph{4-Shot Setting}} \\
\midrule
         \textbf{{Model}} & \multicolumn{3}{@{}c@{}}{\textbf{{Attribute}}} & \multicolumn{3}{@{}c@{}}{\textbf{{Object}}} & \multicolumn{3}{@{}c@{}}{\textbf{{Plurality}}} & \multicolumn{3}{@{}c@{}}{\textbf{{Counting}}} & \multicolumn{3}{@{}c@{}}{\textbf{{Relations}}} & \multicolumn{3}{@{}c@{}}{\textbf{{Action}}} &  \multicolumn{3}{@{}c@{}}{\textbf{{Commonsense}}} & \multicolumn{3}{@{}c@{}}{\textbf{{Temporal}}} & \multicolumn{3}{@{}c@{}}{\textbf{{Coreference}}} & \multicolumn{3}{@{}c@{}}{\textbf{{Avg.}}}\\
\midrule
    & {R} & {S} & {S+C} & {R} & {S} & {S+C} & {R} & {S} & {S+C} & {R} & {S} & {S+C} & {R} & {S} & {S+C} & {R} & {S} & {S+C} & {R} & {S} & {S+C} & {R} & {S} & {S+C} & {R} & {S} & {S+C} & {R} & {S} & {S+C}\\
\cmidrule{2-31}
QW2.5VL-3B-I & 82.6 & 83.2 & 73.6 & 86.6 & 86.2 & 85.0 & 87.2 & 87.6 & 84.0 & 65.0 & 65.0 & 67.2 & 71.0 & 66.4 & 65.8 & 95.2 & 94.4 & 91.2 & 69.2 & 77.6 & 75.2 & 63.2 & 62.0 & 52.0 & 90.8 & 89.2 & 84.0 & 79.0 & 79.1 & 74.2 \\
QW2.5VL-7B-I & 86.4 & 85.8 & 81.4 & 88.8 & 88.8 & 89.0 & 88.8 & 91.6 & 91.6 & 75.6 & 74.6 & 74.0 & 77.6 & 79.4 & 75.6 & 92.8 & 94.0 & 90.4 & 76.4 & 78.4 & 76.4 & 63.6 & 68.8 & 62.8 & 94.0 & 94.0 & 94.8 & 82.7 & 83.9 & 80.1 \\
QW2VL-2B & 48.8 & 54.4 & 46.8 & 75.0 & 75.2 & 63.8 & 70.0 & 72.4 & 52.8 & 55.2 & 54.4 & 43.8 & 51.0 & 51.0 & 45.2 & 88.4 & 86.8 & 70.0 & 50.8 & 56.0 & 49.6 & 50.4 & 54.8 & 37.2 & 70.0 & 66.8 & 54.4 & 62.2 & 63.5 & 51.1 \\
QW2VL-2B-I & 63.0 & 56.4 & 63.8 & 79.4 & 77.2 & 72.8 & 77.2 & 74.8 & 73.2 & 59.4 & 55.2 & 57.4 & 56.4 & 55.0 & 52.0 & 90.0 & 89.6 & 86.8 & 62.8 & 58.4 & 56.8 & 54.0 & 50.0 & 49.6 & 72.8 & 66.8 & 72.8 & 68.3 & 64.8 & 64.0 \\
QW2VL-7B & 82.0 & 77.4 & 74.6 & 87.8 & 87.2 & 84.2 & 85.6 & 84.4 & 80.0 & 71.2 & 72.2 & 74.4 & 69.8 & 65.0 & 68.0 & 94.4 & 91.6 & 91.6 & 70.8 & 70.0 & 74.0 & 59.6 & 63.2 & 56.4 & 90.8 & 90.0 & 87.2 & 79.1 & 77.9 & 75.4 \\
QW2VL-7B-I & 83.6 & 83.8 & 81.2 & 87.0 & 86.6 & 83.8 & 85.6 & 87.2 & 88.4 & 74.4 & 76.2 & 73.0 & 72.2 & 71.4 & 67.4 & 92.8 & 92.4 & 92.4 & 72.4 & 72.4 & 78.0 & 65.2 & 63.6 & 53.2 & 88.8 & 88.8 & 86.8 & 80.2 & 80.3 & 77.2 \\
IVL3-1B & 44.8 & 38.8 & 36.8 & 55.4 & 55.4 & 46.8 & 50.4 & 49.2 & 43.6 & 36.2 & 39.8 & 35.2 & 42.2 & 39.0 & 31.0 & 53.2 & 56.0 & 45.2 & 41.2 & 45.2 & 39.2 & 34.0 & 28.4 & 24.8 & 44.0 & 51.6 & 41.6 & 44.6 & 44.8 & 37.8 \\
IVL3-1B-I & 45.2 & 37.0 & 33.6 & 55.4 & 51.6 & 40.0 & 54.4 & 54.4 & 42.4 & 37.2 & 38.8 & 31.0 & 41.8 & 37.6 & 28.0 & 56.0 & 53.2 & 41.2 & 43.2 & 45.2 & 36.8 & 31.2 & 28.0 & 21.6 & 48.0 & 44.8 & 36.4 & 45.8 & 43.4 & 34.3 \\
IVL3-2B & 73.0 & 75.4 & 68.8 & 80.6 & 82.4 & 77.0 & 73.2 & 76.4 & 74.4 & 61.0 & 61.6 & 53.4 & 56.4 & 56.4 & 55.2 & 83.6 & 85.2 & 81.2 & 62.0 & 69.6 & 63.2 & 48.8 & 45.6 & 44.0 & 78.4 & 76.0 & 73.2 & 68.6 & 69.8 & 64.7 \\
IVL3-2B-I & 74.0 & 71.6 & 59.0 & 79.6 & 82.0 & 70.2 & 74.4 & 71.6 & 66.4 & 55.6 & 54.8 & 47.8 & 56.6 & 55.6 & 47.8 & 84.0 & 82.4 & 70.8 & 61.6 & 65.6 & 59.6 & 50.8 & 44.8 & 39.6 & 74.8 & 71.6 & 65.2 & 67.9 & 66.7 & 57.7 \\
IVL3-8B & 85.4 & 83.8 & 83.2 & 88.8 & 89.6 & 84.8 & 88.8 & 87.2 & 88.8 & 74.6 & 75.4 & 71.6 & 71.4 & 69.0 & 72.2 & 89.2 & 93.6 & 89.2 & 80.4 & 79.2 & 84.4 & 70.4 & 69.2 & 64.8 & 93.6 & 94.4 & 92.4 & 82.5 & 82.4 & 79.9 \\
IVL3-8B-I & 83.0 & 81.8 & 78.8 & 87.2 & 88.6 & 86.2 & 89.2 & 90.0 & 88.4 & 71.0 & 68.0 & 63.4 & 68.4 & 71.0 & 65.6 & 90.8 & 89.2 & 82.8 & 74.4 & 75.2 & 78.4 & 64.4 & 62.8 & 58.4 & 90.8 & 92.4 & 90.8 & 79.9 & 79.9 & 75.2 \\
IVL3-9B & 81.2 & 83.6 & 80.0 & 88.0 & 88.2 & 83.2 & 82.0 & 79.2 & 88.0 & 66.6 & 68.4 & 66.2 & 68.2 & 67.8 & 68.6 & 89.2 & 91.6 & 88.4 & 71.2 & 74.4 & 78.4 & 62.0 & 59.6 & 58.8 & 91.6 & 87.6 & 89.6 & 77.8 & 77.8 & 76.4 \\
IVL3-9B-I & 82.2 & 79.2 & 78.4 & 86.0 & 87.2 & 82.0 & 80.0 & 79.2 & 82.4 & 64.6 & 64.8 & 63.6 & 65.6 & 61.8 & 65.8 & 89.2 & 86.4 & 86.0 & 70.0 & 72.8 & 75.2 & 58.4 & 57.2 & 52.8 & 84.8 & 84.8 & 83.2 & 75.6 & 74.8 & 73.3 \\
IVL2.5-4B-M & 78.6 & 73.6 & 54.6 & 83.6 & 84.0 & 64.0 & 80.8 & 80.8 & 54.4 & 62.6 & 64.6 & 48.4 & 65.8 & 62.4 & 49.0 & 86.4 & 83.2 & 66.4 & 70.4 & 74.8 & 52.4 & 54.4 & 51.2 & 38.4 & 83.6 & 81.6 & 65.6 & 74.0 & 72.9 & 53.4 \\
IVL2.5-8B-M & 84.8 & 80.8 & 81.2 & 87.2 & 87.6 & 85.2 & 86.4 & 83.6 & 80.8 & 72.2 & 70.4 & 68.8 & 72.4 & 68.6 & 72.2 & 90.8 & 90.8 & 88.4 & 75.2 & 75.2 & 73.2 & 58.4 & 62.8 & 54.0 & 90.8 & 90.8 & 88.4 & 79.8 & 79.0 & 75.5 \\
IVL2.5-4B & 75.0 & 74.8 & 47.8 & 84.4 & 83.4 & 57.6 & 79.6 & 82.4 & 48.4 & 59.6 & 58.2 & 44.6 & 62.2 & 60.8 & 40.8 & 88.0 & 87.2 & 53.2 & 70.8 & 70.8 & 42.0 & 58.0 & 50.0 & 30.0 & 81.6 & 82.4 & 51.6 & 73.2 & 72.2 & 45.5 \\
IVL2.5-8B & 80.8 & 81.0 & 74.8 & 87.4 & 85.8 & 81.8 & 83.2 & 80.8 & 78.4 & 68.0 & 65.6 & 63.6 & 64.4 & 65.6 & 66.2 & 90.0 & 91.2 & 86.0 & 71.6 & 66.8 & 73.2 & 62.4 & 52.4 & 44.8 & 87.6 & 84.8 & 85.6 & 77.3 & 74.9 & 71.1 \\
ID3-8B & 77.2 & 76.8 & 73.2 & 82.6 & 81.2 & 81.6 & 80.0 & 83.2 & 85.6 & 63.6 & 63.4 & 62.0 & 59.0 & 56.0 & 60.6 & 94.4 & 96.0 & 92.0 & 68.0 & 67.2 & 70.4 & 47.2 & 47.2 & 42.4 & 78.0 & 79.6 & 82.4 & 72.2 & 72.3 & 71.0 \\
ID2-8B & 66.2 & 65.0 & 60.6 & 80.2 & 81.4 & 82.0 & 79.2 & 77.6 & 74.8 & 58.8 & 60.8 & 57.0 & 53.0 & 52.6 & 55.4 & 82.8 & 85.2 & 88.4 & 64.0 & 64.4 & 58.4 & 43.6 & 45.6 & 39.6 & 87.2 & 86.0 & 80.8 & 68.3 & 68.7 & 64.5 \\
P4M-I & 69.2 & 68.2 & 62.4 & 73.2 & 78.2 & 67.6 & 75.6 & 76.4 & 69.6 & 54.2 & 55.4 & 48.4 & 57.8 & 58.4 & 56.6 & 82.0 & 86.0 & 79.2 & 70.4 & 66.0 & 64.4 & 41.6 & 42.4 & 41.2 & 75.6 & 75.2 & 78.8 & 66.6 & 67.4 & 61.2 \\
P3.5V-I & 60.0 & 56.6 & 50.6 & 59.4 & 65.2 & 68.8 & 59.2 & 69.2 & 67.6 & 51.8 & 49.4 & 45.4 & 48.0 & 47.8 & 45.4 & 84.0 & 84.0 & 74.4 & 61.2 & 58.4 & 51.2 & 39.2 & 33.2 & 34.0 & 71.2 & 74.8 & 68.8 & 59.3 & 59.8 & 54.7 \\
LLV-I-QW-7B & 70.6 & 64.0 & 62.8 & 81.0 & 81.0 & 74.8 & 84.0 & 78.4 & 72.0 & 63.0 & 57.2 & 56.4 & 49.0 & 52.0 & 44.2 & 81.2 & 79.6 & 72.8 & 63.6 & 62.4 & 59.2 & 34.8 & 33.6 & 35.2 & 85.2 & 82.8 & 81.2 & 68.0 & 65.7 & 59.7 \\
LLV-I-QW-0.5B & 48.8 & 50.6 & 47.6 & 58.6 & 56.2 & 26.4 & 47.2 & 44.0 & 32.8 & 26.0 & 27.0 & 14.6 & 43.2 & 40.0 & 33.6 & 63.2 & 63.6 & 41.6 & 52.8 & 50.8 & 48.8 & 31.2 & 18.4 & 15.6 & 44.8 & 46.8 & 40.4 & 46.2 & 44.2 & 32.6 \\
XG-MM-P3-I & 65.2 & 57.2 & 65.0 & 78.2 & 80.2 & 78.4 & 78.0 & 77.2 & 64.0 & 63.2 & 62.6 & 59.2 & 60.4 & 55.2 & 55.6 & 89.2 & 90.4 & 72.0 & 70.4 & 66.4 & 69.2 & 44.0 & 50.4 & 39.6 & 70.8 & 71.6 & 52.0 & 68.8 & 67.9 & 62.9 \\
XG-MM-P3-B & 34.0 & 33.8 & 16.6 & 34.4 & 34.0 & 22.6 & 48.4 & 46.8 & 13.2 & 31.2 & 28.8 & 15.8 & 37.2 & 32.4 & 17.0 & 47.2 & 44.4 & 22.4 & 31.6 & 31.6 & 26.0 & 28.0 & 22.8 & 15.2 & 54.0 & 56.8 & 11.6 & 38.4 & 36.8 & 18.6 \\

\midrule
       \multicolumn{31}{c}{\emph{8-Shot Setting}} \\
\midrule
\textbf{{Model}} & \multicolumn{3}{@{}c@{}}{\textbf{{Attribute}}} & \multicolumn{3}{@{}c@{}}{\textbf{{Object}}} & \multicolumn{3}{@{}c@{}}{\textbf{{Plurality}}} & \multicolumn{3}{@{}c@{}}{\textbf{{Counting}}} & \multicolumn{3}{@{}c@{}}{\textbf{{Relations}}} & \multicolumn{3}{@{}c@{}}{\textbf{{Action}}} &  \multicolumn{3}{@{}c@{}}{\textbf{{Commonsense}}} & \multicolumn{3}{@{}c@{}}{\textbf{{Temporal}}} & \multicolumn{3}{@{}c@{}}{\textbf{{Coreference}}} & \multicolumn{3}{@{}c@{}}{\textbf{{Avg.}}}\\
\midrule
        & {R} & {S} & {S+C} & {R} & {S} & {S+C} & {R} & {S} & {S+C} & {R} & {S} & {S+C} & {R} & {S} & {S+C} & {R} & {S} & {S+C} & {R} & {S} & {S+C} & {R} & {S} & {S+C} & {R} & {S} & {S+C} & {R} & {S} & {S+C}\\

\cmidrule{2-31}
QW2.5VL-3B-I & 82.6 & 82.0 & 75.0 & 85.2 & 86.2 & 85.6 & 85.2 & 86.4 & 80.8 & 65.6 & 66.6 & 62.8 & 69.4 & 67.4 & 66.0 & 92.8 & 94.4 & 90.4 & 68.8 & 78.8 & 76.0 & 61.6 & 54.4 & 50.0 & 90.8 & 89.6 & 79.2 & 78.0 & 78.4 & 73.3 \\
QW2.5VL-7B-I & 86.2 & 86.8 & 81.6 & 89.2 & 88.8 & 88.2 & 89.6 & 91.2 & 90.4 & 76.6 & 77.2 & 72.0 & 78.0 & 78.2 & 76.6 & 94.0 & 95.2 & 94.0 & 78.4 & 80.8 & 80.4 & 65.2 & 68.4 & 62.0 & 93.6 & 94.0 & 94.0 & 83.4 & 84.5 & 80.7 \\
QW2VL-2B & 48.8 & 48.8 & 47.0 & 74.0 & 74.4 & 67.2 & 69.2 & 70.4 & 59.2 & 54.8 & 54.6 & 48.0 & 54.6 & 47.8 & 49.6 & 88.8 & 85.6 & 75.6 & 52.4 & 51.2 & 49.2 & 50.4 & 49.6 & 41.2 & 65.6 & 68.4 & 50.4 & 62.1 & 61.2 & 54.6 \\
QW2VL-2B-I & 59.6 & 53.0 & 63.6 & 77.4 & 77.8 & 74.0 & 77.2 & 74.4 & 76.4 & 57.8 & 55.0 & 55.2 & 54.6 & 54.4 & 53.0 & 91.6 & 91.6 & 85.2 & 59.2 & 56.8 & 57.2 & 52.4 & 49.2 & 47.6 & 71.6 & 66.8 & 74.8 & 66.8 & 64.3 & 64.0 \\
QW2VL-7B & 80.0 & 78.4 & 73.4 & 87.6 & 86.6 & 87.2 & 85.2 & 87.2 & 85.6 & 71.8 & 72.6 & 70.2 & 69.0 & 64.6 & 64.8 & 96.4 & 91.6 & 91.2 & 72.0 & 71.6 & 76.8 & 59.6 & 62.8 & 61.2 & 92.4 & 91.2 & 88.0 & 79.3 & 78.5 & 76.3 \\
QW2VL-7B-I & 84.6 & 82.6 & 78.8 & 87.2 & 87.4 & 85.4 & 86.4 & 87.6 & 88.0 & 74.8 & 74.8 & 75.6 & 71.8 & 72.2 & 68.8 & 94.0 & 92.0 & 90.8 & 71.2 & 71.6 & 74.4 & 62.4 & 61.2 & 55.2 & 90.8 & 90.4 & 86.8 & 80.4 & 80.0 & 77.1 \\
IVL3-1B & 45.8 & 40.8 & 38.0 & 53.8 & 54.2 & 48.8 & 58.4 & 54.4 & 47.6 & 39.2 & 38.2 & 34.2 & 39.0 & 43.0 & 32.8 & 55.6 & 55.6 & 57.2 & 43.2 & 46.0 & 37.2 & 36.4 & 31.2 & 29.6 & 48.4 & 53.2 & 39.6 & 46.6 & 46.3 & 40.7 \\
IVL3-1B-I & 43.2 & 39.4 & 32.4 & 52.2 & 52.6 & 41.8 & 56.4 & 51.6 & 42.0 & 39.0 & 40.4 & 33.6 & 41.4 & 40.8 & 32.0 & 53.6 & 52.4 & 44.4 & 46.4 & 38.8 & 30.0 & 34.0 & 31.6 & 25.2 & 45.2 & 43.2 & 35.6 & 45.7 & 43.4 & 35.2 \\
IVL3-2B & 74.8 & 72.2 & 66.8 & 82.0 & 81.0 & 78.2 & 73.2 & 73.6 & 72.0 & 59.8 & 61.2 & 50.8 & 58.8 & 54.2 & 55.0 & 86.4 & 84.4 & 78.0 & 69.6 & 67.2 & 64.0 & 48.8 & 50.0 & 40.8 & 76.8 & 76.0 & 76.4 & 70.0 & 68.9 & 63.2 \\
IVL3-2B-I & 69.0 & 70.6 & 60.2 & 79.6 & 80.4 & 69.0 & 72.0 & 71.2 & 64.0 & 54.6 & 58.4 & 48.2 & 58.2 & 54.4 & 47.4 & 80.4 & 82.8 & 70.8 & 67.6 & 66.0 & 60.8 & 54.0 & 43.2 & 34.0 & 70.0 & 72.4 & 64.0 & 67.3 & 66.6 & 56.8 \\
IVL3-8B & 86.0 & 84.0 & 83.8 & 88.0 & 88.6 & 84.8 & 92.0 & 92.8 & 89.2 & 72.8 & 75.0 & 72.0 & 70.8 & 68.8 & 72.0 & 91.6 & 88.8 & 85.6 & 78.4 & 75.2 & 82.4 & 69.2 & 67.6 & 68.8 & 92.8 & 93.6 & 91.2 & 82.4 & 81.6 & 79.8 \\
IVL3-8B-I & 85.4 & 82.0 & 76.4 & 89.0 & 86.2 & 84.4 & 87.6 & 90.8 & 82.4 & 69.8 & 68.6 & 67.2 & 69.4 & 66.8 & 68.8 & 90.0 & 89.6 & 84.0 & 73.2 & 73.6 & 75.6 & 64.8 & 61.6 & 54.8 & 91.6 & 91.6 & 90.0 & 80.1 & 79.0 & 74.2 \\
IVL3-9B & 85.2 & 80.0 & 83.6 & 85.6 & 87.6 & 80.0 & 83.6 & 84.8 & 88.8 & 70.0 & 66.4 & 66.4 & 68.6 & 67.4 & 71.4 & 89.2 & 91.6 & 81.2 & 73.6 & 74.4 & 79.2 & 58.8 & 55.6 & 54.8 & 84.8 & 86.4 & 88.0 & 77.7 & 77.1 & 75.7 \\
IVL3-9B-I & 78.6 & 79.4 & 78.0 & 86.4 & 84.8 & 76.0 & 73.6 & 82.8 & 83.6 & 64.6 & 65.8 & 60.4 & 65.6 & 64.2 & 65.0 & 89.2 & 88.0 & 84.8 & 70.8 & 71.6 & 74.0 & 56.4 & 52.4 & 58.4 & 82.4 & 80.0 & 84.4 & 74.2 & 74.3 & 72.5 \\
IVL2.5-4B-M & 72.4 & 73.2 & 53.0 & 82.2 & 81.0 & 65.2 & 79.2 & 76.4 & 57.6 & 61.4 & 61.4 & 50.4 & 66.2 & 61.8 & 45.2 & 88.8 & 83.2 & 59.6 & 72.4 & 72.8 & 51.2 & 48.8 & 47.2 & 30.0 & 85.2 & 82.0 & 59.2 & 73.0 & 71.0 & 51.5 \\
IVL2.5-8B-M & 83.2 & 82.4 & 81.8 & 84.8 & 87.4 & 85.2 & 84.8 & 81.2 & 85.2 & 70.6 & 69.4 & 66.4 & 68.8 & 68.4 & 67.6 & 88.0 & 92.4 & 87.2 & 73.2 & 74.4 & 76.4 & 55.6 & 54.0 & 52.4 & 90.4 & 90.0 & 88.0 & 77.7 & 77.7 & 75.3 \\
IVL2.5-4B & 71.8 & 74.4 & 44.2 & 82.6 & 81.6 & 56.4 & 79.6 & 81.6 & 54.8 & 58.4 & 57.0 & 39.0 & 61.8 & 56.2 & 40.0 & 87.2 & 81.2 & 48.4 & 68.4 & 70.4 & 42.4 & 53.2 & 45.2 & 30.4 & 84.4 & 83.2 & 58.0 & 71.9 & 70.1 & 44.4 \\
IVL2.5-8B & 77.6 & 78.8 & 75.6 & 86.2 & 85.6 & 76.6 & 81.2 & 82.8 & 80.4 & 69.0 & 69.6 & 62.8 & 65.0 & 63.4 & 59.6 & 92.0 & 88.0 & 80.0 & 68.8 & 70.8 & 68.4 & 55.6 & 52.8 & 42.0 & 88.4 & 89.6 & 80.4 & 76.0 & 75.7 & 68.2 \\
ID3-8B & 76.6 & 77.8 & 71.6 & 81.0 & 80.8 & 82.0 & 82.0 & 82.8 & 80.4 & 64.2 & 62.4 & 62.8 & 57.4 & 59.0 & 59.4 & 95.6 & 96.0 & 88.8 & 67.6 & 66.0 & 71.2 & 44.8 & 49.2 & 44.4 & 77.6 & 78.0 & 79.2 & 71.9 & 72.4 & 70.1 \\
ID2-8B & 64.2 & 68.4 & 59.2 & 79.0 & 80.0 & 79.8 & 76.4 & 81.2 & 79.2 & 58.8 & 60.6 & 60.8 & 52.4 & 50.4 & 53.6 & 82.8 & 86.8 & 81.2 & 60.4 & 64.4 & 60.8 & 42.0 & 44.4 & 40.4 & 84.4 & 85.2 & 76.8 & 66.7 & 69.0 & 64.4 \\
P4M-I & 63.6 & 70.2 & 65.4 & 72.8 & 75.4 & 66.2 & 74.8 & 74.4 & 74.4 & 49.4 & 49.2 & 43.6 & 54.0 & 54.0 & 58.0 & 81.6 & 87.2 & 82.4 & 66.4 & 65.6 & 67.6 & 36.4 & 49.6 & 36.4 & 78.0 & 74.0 & 72.8 & 64.1 & 66.6 & 61.8 \\
P3.5V-I & 55.2 & 55.6 & 50.0 & 66.0 & 67.4 & 65.6 & 59.2 & 66.4 & 60.4 & 49.6 & 49.2 & 45.0 & 46.8 & 45.6 & 43.4 & 83.2 & 84.4 & 74.0 & 58.0 & 54.0 & 52.8 & 37.6 & 36.8 & 33.6 & 68.8 & 67.2 & 59.2 & 58.3 & 58.5 & 53.1 \\
LLV-I-QW-7B & 64.4 & 61.6 & 62.8 & 80.4 & 80.0 & 78.4 & 84.4 & 78.4 & 76.0 & 58.0 & 55.4 & 59.6 & 53.4 & 53.0 & 49.2 & 88.4 & 82.8 & 80.0 & 66.0 & 66.0 & 57.6 & 33.6 & 37.6 & 30.4 & 80.4 & 81.2 & 79.6 & 67.7 & 66.2 & 61.8 \\
LLV-I-QW-0.5B & 48.4 & 48.0 & 44.0 & 51.0 & 51.8 & 29.0 & 45.2 & 47.6 & 34.8 & 27.8 & 29.6 & 18.2 & 42.6 & 39.4 & 34.0 & 60.4 & 62.8 & 44.8 & 53.6 & 48.4 & 47.6 & 28.4 & 18.0 & 17.6 & 46.0 & 46.8 & 47.6 & 44.8 & 43.6 & 33.8 \\
XG-MM-P3-I & 66.8 & 62.6 & 60.0 & 78.0 & 78.0 & 72.4 & 75.6 & 79.2 & 63.2 & 63.0 & 63.2 & 56.8 & 59.2 & 56.0 & 54.4 & 88.0 & 88.0 & 72.4 & 70.8 & 68.4 & 65.2 & 49.2 & 48.4 & 43.6 & 74.4 & 72.0 & 47.6 & 69.4 & 68.4 & 61.0 \\
XG-MM-P3-B & 31.8 & 32.8 & 25.6 & 39.0 & 40.0 & 37.6 & 48.0 & 49.6 & 35.2 & 27.2 & 26.4 & 27.6 & 37.0 & 33.2 & 26.4 & 47.6 & 47.2 & 41.6 & 34.0 & 33.2 & 28.8 & 27.2 & 19.6 & 17.6 & 48.8 & 46.8 & 32.4 & 37.8 & 36.5 & 30.1 \\

    \bottomrule
\end{tabular}
    \end{threeparttable}
}
    \caption{\scriptsize Accuracy performance of the evaluated MLLMs on \benchmark, varying in model size and pretraining strategies, evaluated with 0-4-8 shots across three settings: Random (R), Similar (S), and Similar with CoT (S+C) settings. In the R setting, few-shot examples are randomly selected. In the S setting, few-shot examples are selected based on visual and textual similarity. In the S+C setting, examples are also selected based on visual and textual similarity but additionally include a CoT description. Models with the suffix ’I’ indicate instruction-tuned versions.}
    \label{tab:all_results}
\end{table*}

\begin{table*}[!t]
    \centering
    \renewcommand{\arraystretch}{1.15}
    \resizebox{\linewidth}{!}{
    \begin{threeparttable}
    \begin{tabular}{
  @{}l@{\;}
  c@{$\;$}c@{$\;\;\;$}
  c@{$\;$}c@{$\;\;\;$}
  c@{$\;$}c@{$\;\;\;$}
  c@{$\;$}c@{$\;\;\;$}
  c@{$\;$}c@{$\;\;\;$}
  c@{$\;$}c@{$\;\;\;$}
  c@{$\;$}c@{$\;\;\;$}
  c@{$\;$}c@{$\;\;\;$}
  c@{$\;$}c@{$\;\;\;$}
  c@{$\;$}c@{}}

    \toprule
    \multicolumn{21}{c}{\emph{Zero-Shot Setting}} \\
\midrule
  \textbf{Model} & \multicolumn{2}{@{}c@{}}{\textbf{{Attribute}}} & \multicolumn{2}{@{}c@{}}{\textbf{{Object}}} & \multicolumn{2}{@{}c@{}}{\textbf{{Plurality}}} & \multicolumn{2}{@{}c@{}}{\textbf{{Counting}}} & \multicolumn{2}{@{}c@{}}{\textbf{{Relations}}} & \multicolumn{2}{@{}c@{}}{\textbf{{Action}}} &  \multicolumn{2}{@{}c@{}}{\textbf{{Commonsense}}} & \multicolumn{2}{@{}c@{}}{\textbf{{Temporal}}} & \multicolumn{2}{@{}c@{}}{\textbf{{Coreference}}} & \multicolumn{2}{@{}c@{}}{\textbf{{Avg.}}}\\
\midrule      

QW2.5VL-3B-I & \multicolumn{2}{c}{83.0} & \multicolumn{2}{c}{88.2} & \multicolumn{2}{c}{80.4} & \multicolumn{2}{c}{67.2} & \multicolumn{2}{c}{68.4} & \multicolumn{2}{c}{89.6} & \multicolumn{2}{c}{68.0} & \multicolumn{2}{c}{61.2} & \multicolumn{2}{c}{83.6} & \multicolumn{2}{c}{76.6} \\
QW2.5VL-7B-I & \multicolumn{2}{c}{83.8} & \multicolumn{2}{c}{89.0} & \multicolumn{2}{c}{86.4} & \multicolumn{2}{c}{74.2} & \multicolumn{2}{c}{77.8} & \multicolumn{2}{c}{92.8} & \multicolumn{2}{c}{77.6} & \multicolumn{2}{c}{58.8} & \multicolumn{2}{c}{90.8} & \multicolumn{2}{c}{81.2} \\
QW2VL-2B & \multicolumn{2}{c}{41.4} & \multicolumn{2}{c}{72.4} & \multicolumn{2}{c}{68.8} & \multicolumn{2}{c}{51.8} & \multicolumn{2}{c}{48.4} & \multicolumn{2}{c}{78.4} & \multicolumn{2}{c}{44.0} & \multicolumn{2}{c}{32.4} & \multicolumn{2}{c}{52.0} & \multicolumn{2}{c}{54.4} \\
QW2VL-2B-I & \multicolumn{2}{c}{60.0} & \multicolumn{2}{c}{77.0} & \multicolumn{2}{c}{76.4} & \multicolumn{2}{c}{60.2} & \multicolumn{2}{c}{56.0} & \multicolumn{2}{c}{83.2} & \multicolumn{2}{c}{54.4} & \multicolumn{2}{c}{42.8} & \multicolumn{2}{c}{63.6} & \multicolumn{2}{c}{63.7} \\
QW2VL-7B & \multicolumn{2}{c}{73.8} & \multicolumn{2}{c}{85.0} & \multicolumn{2}{c}{80.0} & \multicolumn{2}{c}{68.2} & \multicolumn{2}{c}{63.6} & \multicolumn{2}{c}{92.0} & \multicolumn{2}{c}{69.2} & \multicolumn{2}{c}{52.0} & \multicolumn{2}{c}{84.0} & \multicolumn{2}{c}{74.2} \\
QW2VL-7B-I & \multicolumn{2}{c}{82.4} & \multicolumn{2}{c}{87.4} & \multicolumn{2}{c}{86.4} & \multicolumn{2}{c}{72.8} & \multicolumn{2}{c}{71.2} & \multicolumn{2}{c}{91.6} & \multicolumn{2}{c}{70.8} & \multicolumn{2}{c}{54.4} & \multicolumn{2}{c}{88.4} & \multicolumn{2}{c}{78.4} \\
ID3-8B & \multicolumn{2}{c}{87.8} & \multicolumn{2}{c}{87.4} & \multicolumn{2}{c}{84.8} & \multicolumn{2}{c}{72.2} & \multicolumn{2}{c}{73.6} & \multicolumn{2}{c}{96.0} & \multicolumn{2}{c}{78.8} & \multicolumn{2}{c}{63.2} & \multicolumn{2}{c}{87.2} & \multicolumn{2}{c}{81.2} \\
ID2-8B & \multicolumn{2}{c}{74.2} & \multicolumn{2}{c}{83.0} & \multicolumn{2}{c}{83.2} & \multicolumn{2}{c}{61.0} & \multicolumn{2}{c}{58.4} & \multicolumn{2}{c}{83.6} & \multicolumn{2}{c}{65.2} & \multicolumn{2}{c}{45.6} & \multicolumn{2}{c}{88.8} & \multicolumn{2}{c}{71.4} \\
P3.5V-I & \multicolumn{2}{c}{65.8} & \multicolumn{2}{c}{77.6} & \multicolumn{2}{c}{71.6} & \multicolumn{2}{c}{57.0} & \multicolumn{2}{c}{56.6} & \multicolumn{2}{c}{85.6} & \multicolumn{2}{c}{64.8} & \multicolumn{2}{c}{46.4} & \multicolumn{2}{c}{77.6} & \multicolumn{2}{c}{67.0} \\
LLV-I-QW-7B & \multicolumn{2}{c}{37.2} & \multicolumn{2}{c}{25.8} & \multicolumn{2}{c}{45.2} & \multicolumn{2}{c}{39.6} & \multicolumn{2}{c}{34.2} & \multicolumn{2}{c}{36.0} & \multicolumn{2}{c}{33.6} & \multicolumn{2}{c}{25.2} & \multicolumn{2}{c}{44.4} & \multicolumn{2}{c}{35.7} \\
LLV-I-QW-0.5B & \multicolumn{2}{c}{38.6} & \multicolumn{2}{c}{41.2} & \multicolumn{2}{c}{52.0} & \multicolumn{2}{c}{37.2} & \multicolumn{2}{c}{37.2} & \multicolumn{2}{c}{51.6} & \multicolumn{2}{c}{30.8} & \multicolumn{2}{c}{28.0} & \multicolumn{2}{c}{49.6} & \multicolumn{2}{c}{40.7} \\

\midrule
       \multicolumn{21}{c}{\emph{4-Shot Setting}} \\
\midrule
         \textbf{{Model}} & \multicolumn{2}{@{}c@{}}{\textbf{{Attribute}}} & \multicolumn{2}{@{}c@{}}{\textbf{{Object}}} & \multicolumn{2}{@{}c@{}}{\textbf{{Plurality}}} & \multicolumn{2}{@{}c@{}}{\textbf{{Counting}}} & \multicolumn{2}{@{}c@{}}{\textbf{{Relations}}} & \multicolumn{2}{@{}c@{}}{\textbf{{Action}}} &  \multicolumn{2}{@{}c@{}}{\textbf{{Commonsense}}} & \multicolumn{2}{@{}c@{}}{\textbf{{Temporal}}} & \multicolumn{2}{@{}c@{}}{\textbf{{Coreference}}} & \multicolumn{2}{@{}c@{}}{\textbf{{Avg.}}}\\
\midrule
    & {R} & {S} & {R} & {S} & {R} & {S} & {R} & {S} & {R} & {S} & {R} & {S} & {R} & {S} & {R} & {S} & {R} & {S} & {R} & {S}\\
\cmidrule{2-21}

QW2.5VL-3B-I & 84.0 & 82.4 & 86.8 & 86.8 & 85.2 & 84.8 & 64.4 & 63.2 & 69.8 & 67.6 & 90.0 & 95.6 & 67.2 & 76.4 & 58.4 & 56.0 & 92.0 & 87.2 & 77.5 & 76.6 \\
QW2.5VL-7B-I & 84.0 & 84.4 & 89.0 & 86.6 & 90.8 & 90.8 & 73.6 & 71.8 & 78.4 & 78.2 & 94.0 & 94.4 & 77.2 & 80.8 & 56.8 & 60.0 & 94.0 & 94.0 & 82.0 & 80.9 \\
QW2VL-2B & 52.4 & 51.4 & 71.6 & 74.0 & 71.6 & 72.4 & 54.0 & 53.8 & 51.2 & 50.0 & 87.2 & 86.0 & 52.0 & 55.6 & 45.2 & 44.8 & 67.2 & 65.6 & 61.4 & 61.0 \\
QW2VL-2B-I & 62.4 & 56.4 & 76.8 & 77.4 & 75.6 & 73.2 & 56.2 & 53.8 & 56.6 & 55.4 & 91.2 & 90.0 & 62.0 & 58.8 & 48.4 & 47.2 & 73.2 & 70.0 & 66.9 & 64.0 \\
QW2VL-7B & 79.0 & 73.4 & 87.0 & 86.4 & 86.0 & 85.2 & 72.6 & 69.8 & 69.4 & 62.6 & 94.4 & 90.8 & 70.4 & 71.2 & 56.0 & 55.2 & 92.8 & 90.8 & 78.6 & 74.3 \\
QW2VL-7B-I & 81.6 & 81.6 & 86.8 & 85.2 & 86.8 & 86.0 & 71.8 & 71.6 & 72.0 & 71.0 & 91.6 & 90.0 & 71.6 & 72.4 & 54.8 & 58.8 & 91.2 & 90.8 & 78.7 & 77.1 \\
ID3-8B & 96.4 & 96.2 & 96.0 & 94.6 & 97.2 & 96.0 & 92.0 & 90.8 & 92.4 & 92.6 & 98.0 & 99.6 & 94.0 & 93.2 & 91.6 & 89.6 & 94.4 & 95.2 & 94.7 & 94.1 \\
ID2-8B & 85.6 & 86.4 & 88.6 & 89.8 & 88.8 & 88.8 & 75.0 & 78.8 & 74.4 & 72.6 & 90.8 & 90.8 & 82.4 & 85.6 & 65.6 & 71.6 & 95.6 & 93.6 & 83.0 & 83.0 \\
P3.5V-I & 61.0 & 57.6 & 75.4 & 75.4 & 72.0 & 69.6 & 56.6 & 55.8 & 52.4 & 53.4 & 88.4 & 86.8 & 60.8 & 60.0 & 43.2 & 39.6 & 69.2 & 77.6 & 64.3 & 62.3 \\
LLV-I-QW-7B & 35.0 & 37.8 & 51.2 & 50.4 & 49.6 & 50.4 & 37.2 & 37.6 & 36.4 & 37.0 & 48.8 & 50.0 & 35.2 & 34.4 & 32.4 & 30.4 & 56.0 & 56.0 & 42.4 & 41.0 \\
LLV-I-QW-0.5B & 37.2 & 37.4 & 43.4 & 39.8 & 52.4 & 52.0 & 39.6 & 42.0 & 38.0 & 38.8 & 56.4 & 54.8 & 32.8 & 32.8 & 30.4 & 24.0 & 50.8 & 49.6 & 42.3 & 40.2 \\

\midrule
       \multicolumn{21}{c}{\emph{8-Shot Setting}} \\
\midrule
\textbf{{Model}} & \multicolumn{2}{@{}c@{}}{\textbf{{Attribute}}} & \multicolumn{2}{@{}c@{}}{\textbf{{Object}}} & \multicolumn{2}{@{}c@{}}{\textbf{{Plurality}}} & \multicolumn{2}{@{}c@{}}{\textbf{{Counting}}} & \multicolumn{2}{@{}c@{}}{\textbf{{Relations}}} & \multicolumn{2}{@{}c@{}}{\textbf{{Action}}} &  \multicolumn{2}{@{}c@{}}{\textbf{{Commonsense}}} & \multicolumn{2}{@{}c@{}}{\textbf{{Temporal}}} & \multicolumn{2}{@{}c@{}}{\textbf{{Coreference}}} & \multicolumn{2}{@{}c@{}}{\textbf{{Avg.}}}\\
\midrule
        & {R} & {S} & {R} & {S} & {R} & {S} & {R} & {S} & {R} & {S} & {R} & {S} & {R} & {S} & {R} & {S} & {R} & {S} & {R} & {S}\\

\cmidrule{2-21}

QW2.5VL-3B-I & 84.0 & 82.4 & 86.8 & 86.8 & 85.2 & 84.8 & 64.4 & 63.2 & 69.8 & 67.6 & 90.0 & 95.6 & 67.2 & 76.4 & 58.4 & 56.0 & 92.0 & 87.2 & 77.5 & 76.6 \\
QW2.5VL-7B-I & 84.0 & 84.4 & 89.0 & 86.6 & 90.8 & 90.8 & 73.6 & 71.8 & 78.4 & 78.2 & 94.0 & 94.4 & 77.2 & 80.8 & 56.8 & 60.0 & 94.0 & 94.0 & 82.0 & 80.9 \\
QW2VL-2B & 52.4 & 51.4 & 71.6 & 74.0 & 71.6 & 72.4 & 54.0 & 53.8 & 51.2 & 50.0 & 87.2 & 86.0 & 52.0 & 55.6 & 45.2 & 44.8 & 67.2 & 65.6 & 61.4 & 61.0 \\
QW2VL-2B-I & 62.4 & 56.4 & 76.8 & 77.4 & 75.6 & 73.2 & 56.2 & 53.8 & 56.6 & 55.4 & 91.2 & 90.0 & 62.0 & 58.8 & 48.4 & 47.2 & 73.2 & 70.0 & 66.9 & 64.0 \\
QW2VL-7B & 79.0 & 73.4 & 87.0 & 86.4 & 86.0 & 85.2 & 72.6 & 69.8 & 69.4 & 62.6 & 94.4 & 90.8 & 70.4 & 71.2 & 56.0 & 55.2 & 92.8 & 90.8 & 78.6 & 74.3 \\
QW2VL-7B-I & 81.6 & 81.6 & 86.8 & 85.2 & 86.8 & 86.0 & 71.8 & 71.6 & 72.0 & 71.0 & 91.6 & 90.0 & 71.6 & 72.4 & 54.8 & 58.8 & 91.2 & 90.8 & 78.7 & 77.1 \\
ID3-8B & 96.4 & 96.2 & 96.0 & 94.6 & 97.2 & 96.0 & 92.0 & 90.8 & 92.4 & 92.6 & 98.0 & 99.6 & 94.0 & 93.2 & 91.6 & 89.6 & 94.4 & 95.2 & 94.7 & 94.1 \\
ID2-8B & 85.6 & 86.4 & 88.6 & 89.8 & 88.8 & 88.8 & 75.0 & 78.8 & 74.4 & 72.6 & 90.8 & 90.8 & 82.4 & 85.6 & 65.6 & 71.6 & 95.6 & 93.6 & 83.0 & 83.0 \\
P3.5V-I & 61.0 & 57.6 & 75.4 & 75.4 & 72.0 & 69.6 & 56.6 & 55.8 & 52.4 & 53.4 & 88.4 & 86.8 & 60.8 & 60.0 & 43.2 & 39.6 & 69.2 & 77.6 & 64.3 & 62.3 \\
LLV-I-QW-7B & 35.0 & 37.8 & 51.2 & 50.4 & 49.6 & 50.4 & 37.2 & 37.6 & 36.4 & 37.0 & 48.8 & 50.0 & 35.2 & 34.4 & 32.4 & 30.4 & 56.0 & 56.0 & 42.4 & 41.0 \\
LLV-I-QW-0.5B & 37.2 & 37.4 & 43.4 & 39.8 & 52.4 & 52.0 & 39.6 & 42.0 & 38.0 & 38.8 & 56.4 & 54.8 & 32.8 & 32.8 & 30.4 & 24.0 & 50.8 & 49.6 & 42.3 & 40.2 \\

    \bottomrule
\end{tabular}
    \end{threeparttable}
}
    \caption{Pairwise accuracy results on \benchmark. For each question, models are presented with multiple answer options. We compute the perplexity score for each option and select the one with the lowest perplexity as the model’s prediction. Accuracy is then calculated based on the correctness of these selections.}
    \label{tab:perp_results}
\end{table*}

\begin{figure*}[t!]
    \centering
    \includegraphics[width=\linewidth]{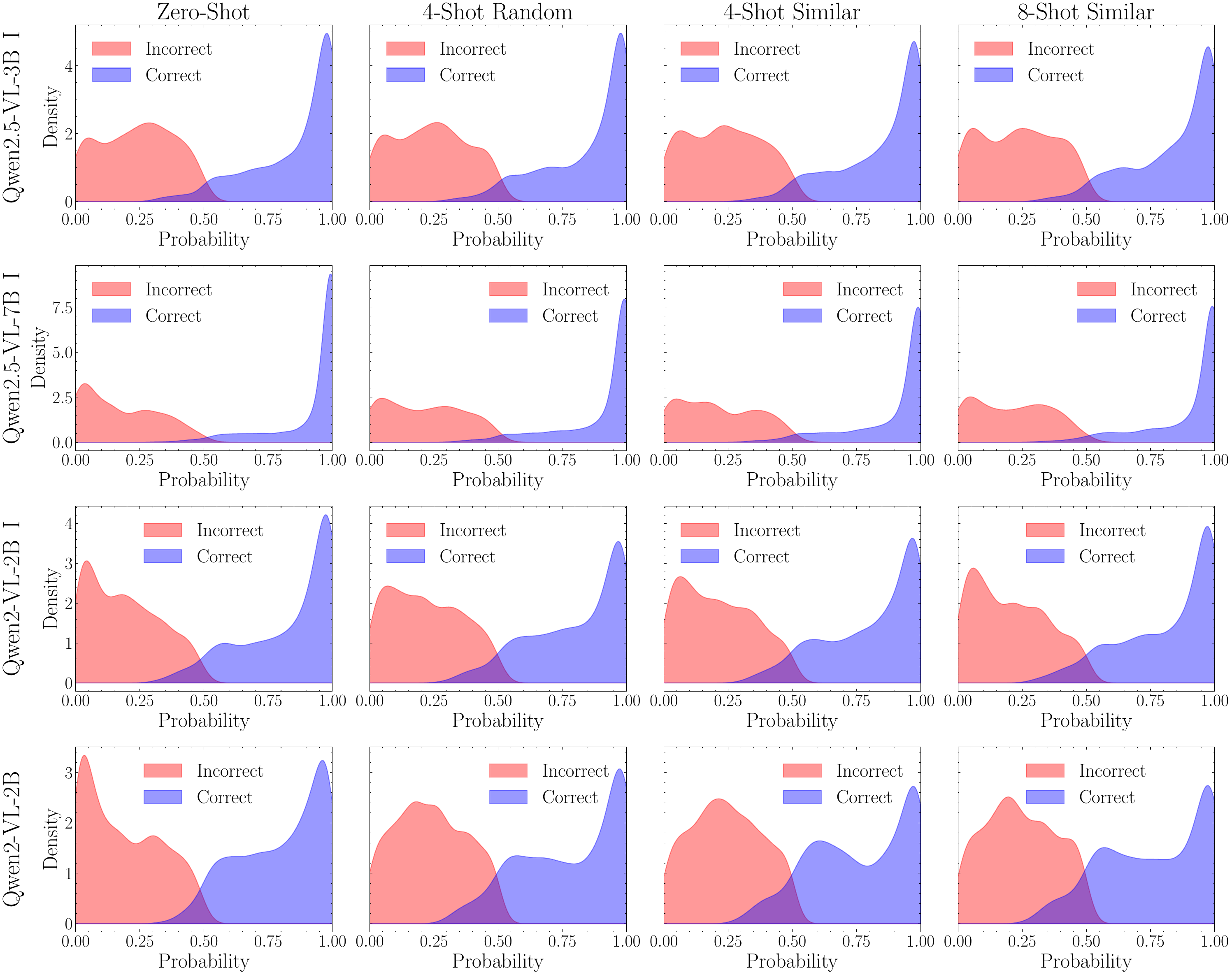}
    \caption{Uncertainty quantification for other Qwen models, showing the probability distributions assigned to correct (blue) and incorrect (red) predictions across different prompting strategies.}
    \label{fig:uncertainity_quantification_appd}
\end{figure*}

\subsection{How Hard \benchmark?}
\begin{figure*}[t!]
    \centering
    \includegraphics[width=\linewidth]{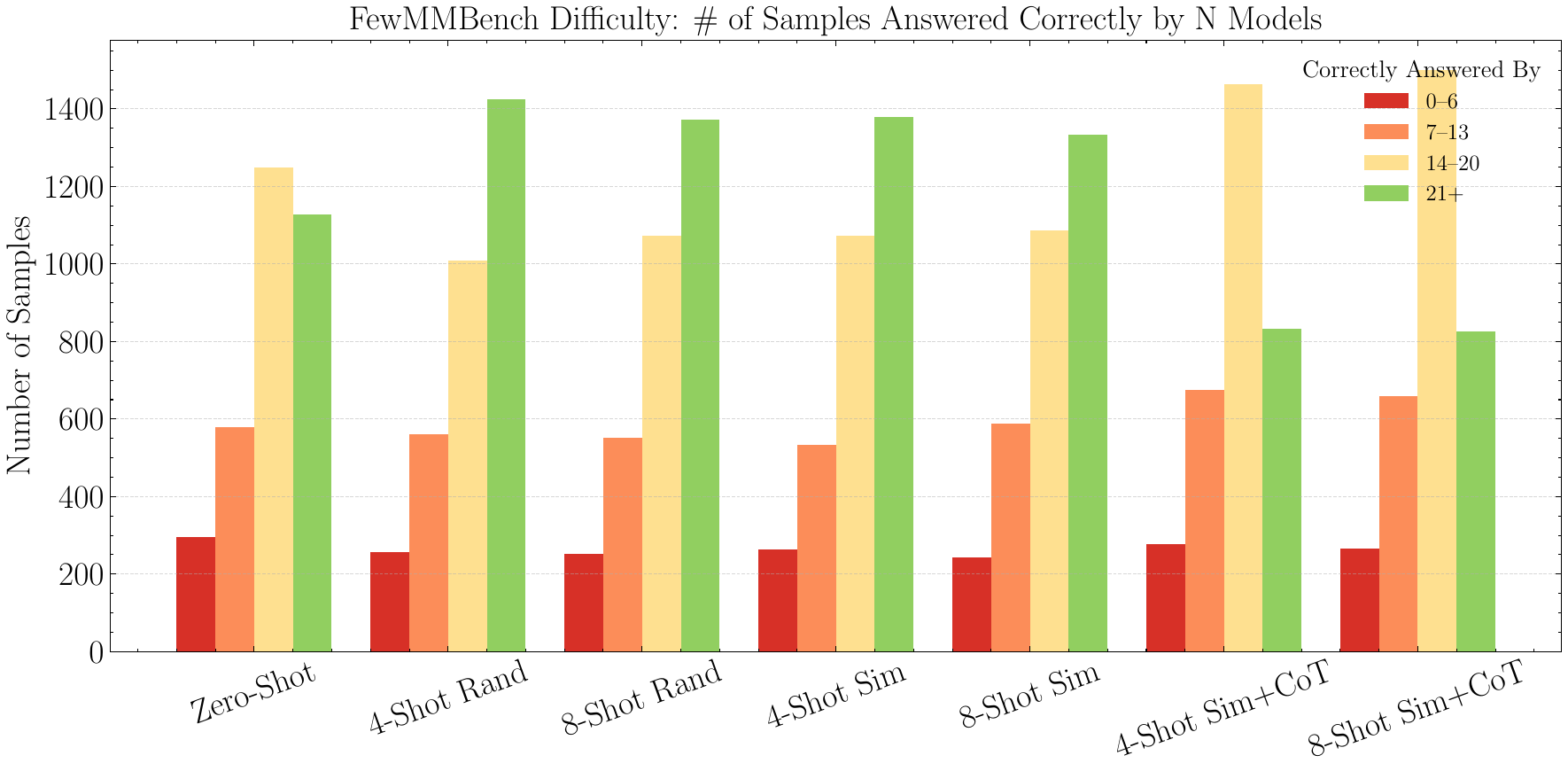}
    \caption{\textbf{Distribution of correctly answered samples across models in \benchmark}. Each bar represents the number of benchmark samples that were correctly answered by a specific number of models, grouped into four buckets: [0–6], [7–13], [14–20], and [21+] out of 26 total models. Results are shown across seven different evaluation settings, including zero-shot and various few-shot configurations with random or semantically similar demonstrations, with and without chain-of-thought (CoT) prompting. The distribution illustrates the challenging nature of the benchmark, with a substantial portion of samples answered correctly by only a few models.}
    \label{fig:how_hard_dataset}
\end{figure*}

Figure \ref{fig:how_hard_dataset} illustrates the distribution of correctly answered samples in \benchmark, categorized by how many of the evaluated MLLMs answered each instance correctly. Each sample is grouped into one of four buckets—[0–6], [7–13], [14–20], and [21+]—based on the number of models that produced a correct answer. The analysis is conducted across seven evaluation settings, ranging from zero-shot to increasingly supportive few-shot contexts that incorporate semantically similar examples and chain-of-thought prompting.

The results clearly demonstrate the challenging nature of \benchmark. In the zero-shot setting, a significant portion of the samples falls into the [14–20] and [21+] buckets, indicating that most models struggle to perform well without any contextual support. As expected, few-shot prompting provides some improvement, particularly when demonstrations are semantically similar to the test input or when reasoning is scaffolded with CoT. However, the overall shift toward lower agreement buckets (i.e., [0–6] and [7-13]) remains modest across these conditions.

Even in the most supportive configuration—8-shot similar demonstrations with CoT—the number of samples answered correctly by a large majority of models remains limited. This suggests that current MLLMs do not yet fully capitalize on the advantages provided by prompting strategies. In other words, while similarity-based demonstrations and CoT prompting are directionally helpful, their practical impact is constrained by the models’ limited ability to generalize from few examples or follow multi-step reasoning cues.

These findings highlight a critical gap: the prompting techniques that have shown success in language-only LLMs are not yet being leveraged effectively in the multimodal setting. Bridging this gap requires architectural or training improvements that allow MLLMs to more robustly integrate context and demonstrations, particularly in tasks that require complex grounding or reasoning. FewMMBench thus serves not only as a challenging testbed, but also as a diagnostic tool to expose such underutilized capabilities.

\subsection{Uncertainity Quantification}
We conduct an uncertainty quantification analysis to assess how different prompting strategies affect model confidence and calibration. In Figure~\ref{fig:uncertainity_quantification_appd}, we visualize the probability distributions assigned by various Qwen model variants to their correct (blue) and incorrect (red) predictions under Zero-Shot, 4-Shot Random, 4-Shot Similar, and 8-Shot Similar prompting conditions.

Interestingly, our findings reveal that few-shot prompting does not consistently improve model calibration or performance. Across both instruction-tuned and non-instruction-tuned variants, the average probability assigned to correct predictions decreases in the few-shot setting compared to zero-shot. This trend suggests that the added demonstrations may introduce noise or uncertainty rather than provide useful guidance.

Moreover, we observe that non-instruction-tuned models (e.g., Qwen2-VL-7B and Qwen2-VL-2B) appear to benefit slightly more from random rather than similar in-context examples—a surprising outcome given the intuition that semantically similar demonstrations should be more helpful. This implies that these models may not yet be reliably leveraging semantic similarity in few-shot setups, possibly due to the absence of alignment or instruction-following fine-tuning.

Overall, these results challenge the assumption that few-shot prompting universally enhances performance and highlight the importance of understanding model behavior under uncertainty.
\subsection{Attribute Recognition}
\label{appd_attribute_r}
The Attribute Recognition task in FewMMBench assesses a model’s ability to identify fine-grained visual properties of objects, such as color, material, texture, or shape. Each instance consists of an image and a multiple-choice question focused on a specific attribute, requiring the model to distinguish subtle visual cues. This task isolates attribute understanding from object recognition by controlling object identity across examples. It serves as a focused probe of visual grounding and compositional reasoning within multimodal models.
\subsubsection{Example Selection}
\begin{figure*}[ht!]
    \centering
    \begin{subfigure}[b]{0.3\linewidth}
        \centering
        \includegraphics[width=\textwidth]{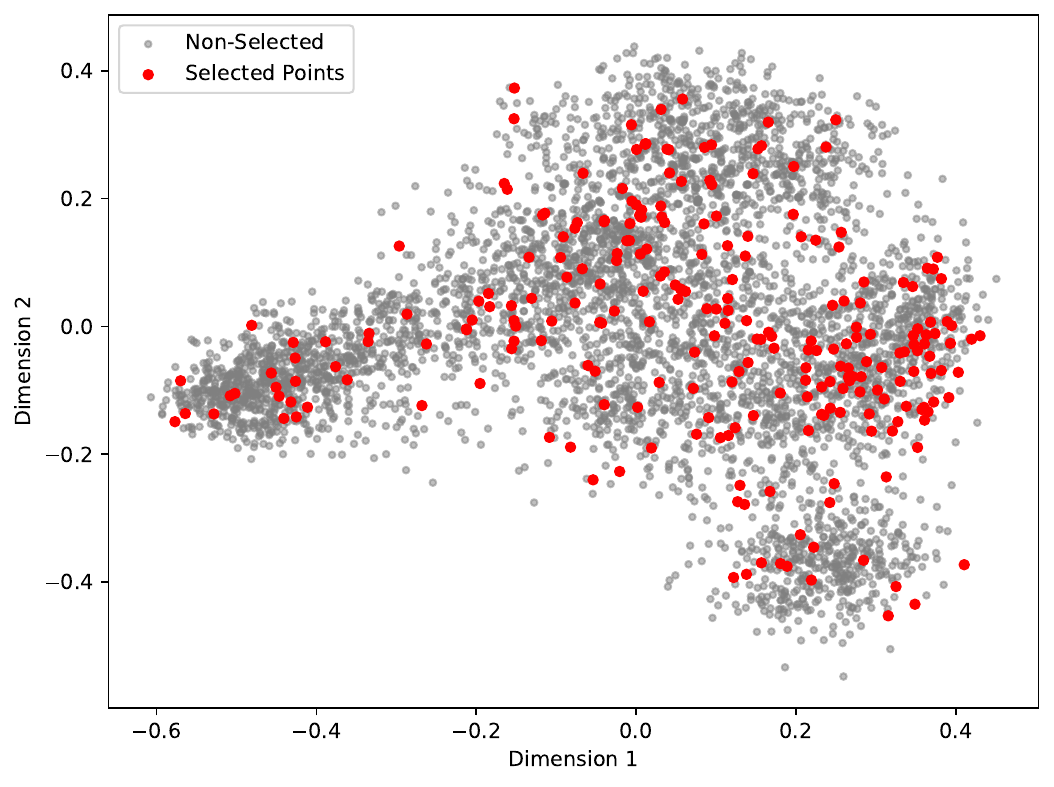}
        \caption{PCA Projection}
        
    \end{subfigure}
    \begin{subfigure}[b]{0.3\linewidth}
        \centering
        \includegraphics[width=\textwidth]{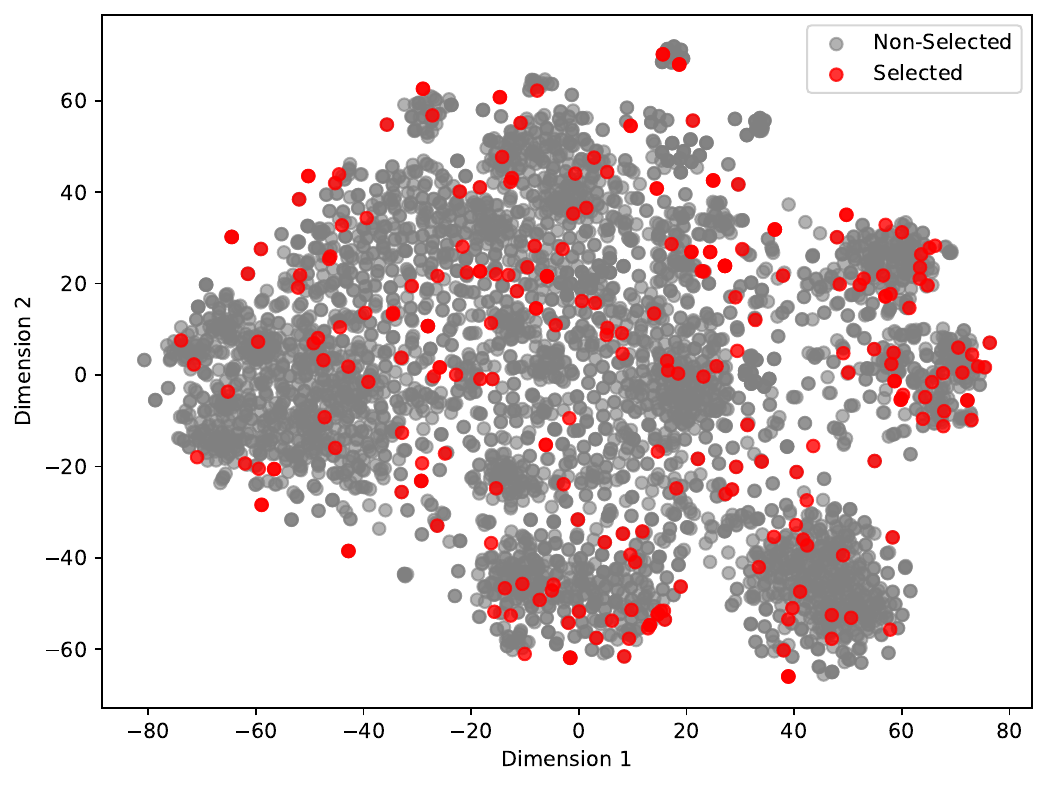}
        \caption{t-SNE Visualization}
        
    \end{subfigure}
    \begin{subfigure}[b]{0.3\linewidth}
        \centering
        \includegraphics[width=\textwidth]{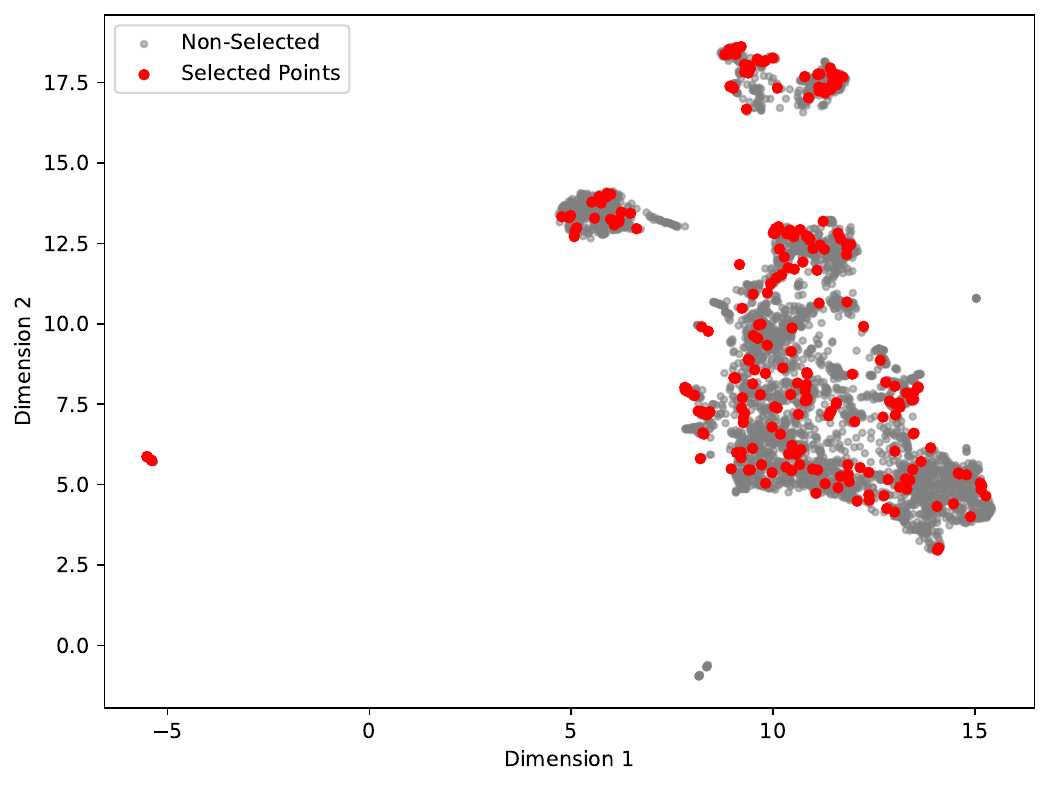}
        \caption{UMAP Visualization}
        
    \end{subfigure}

    \vspace{0.5cm} %

    \begin{subfigure}[b]{0.3\linewidth}
        \centering
        \includegraphics[width=\textwidth]{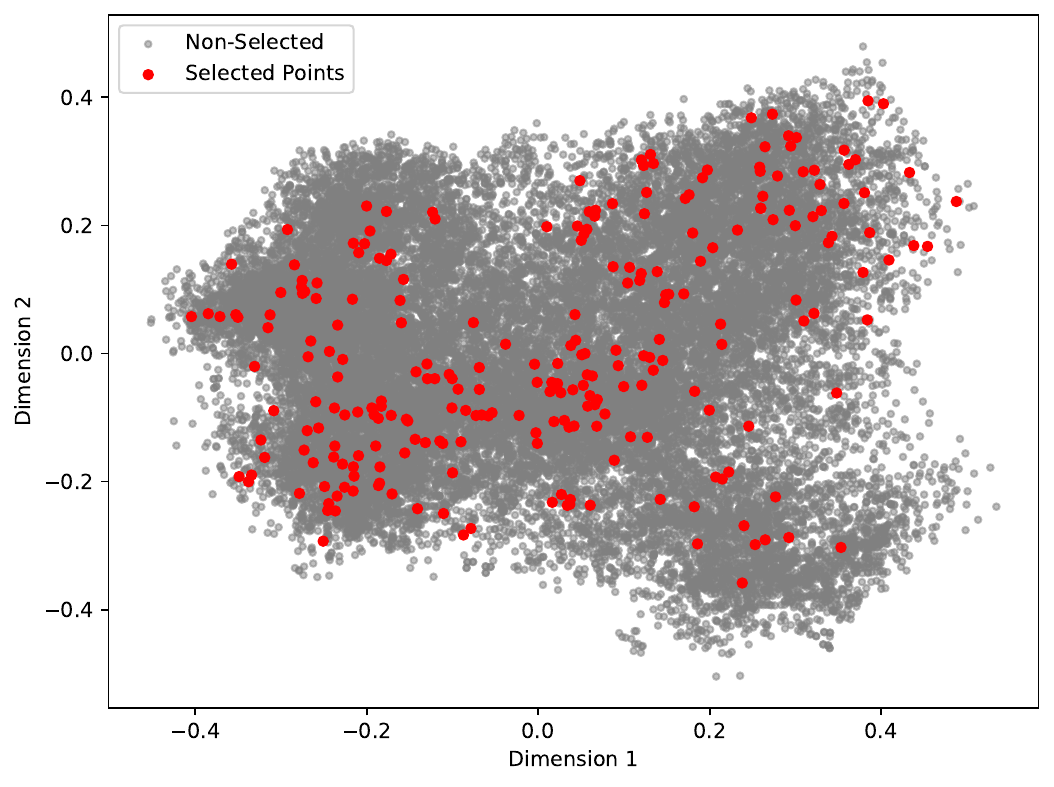}
        \caption{PCA Projection}
        
    \end{subfigure}
    \begin{subfigure}[b]{0.3\linewidth}
        \centering
        \includegraphics[width=\textwidth]{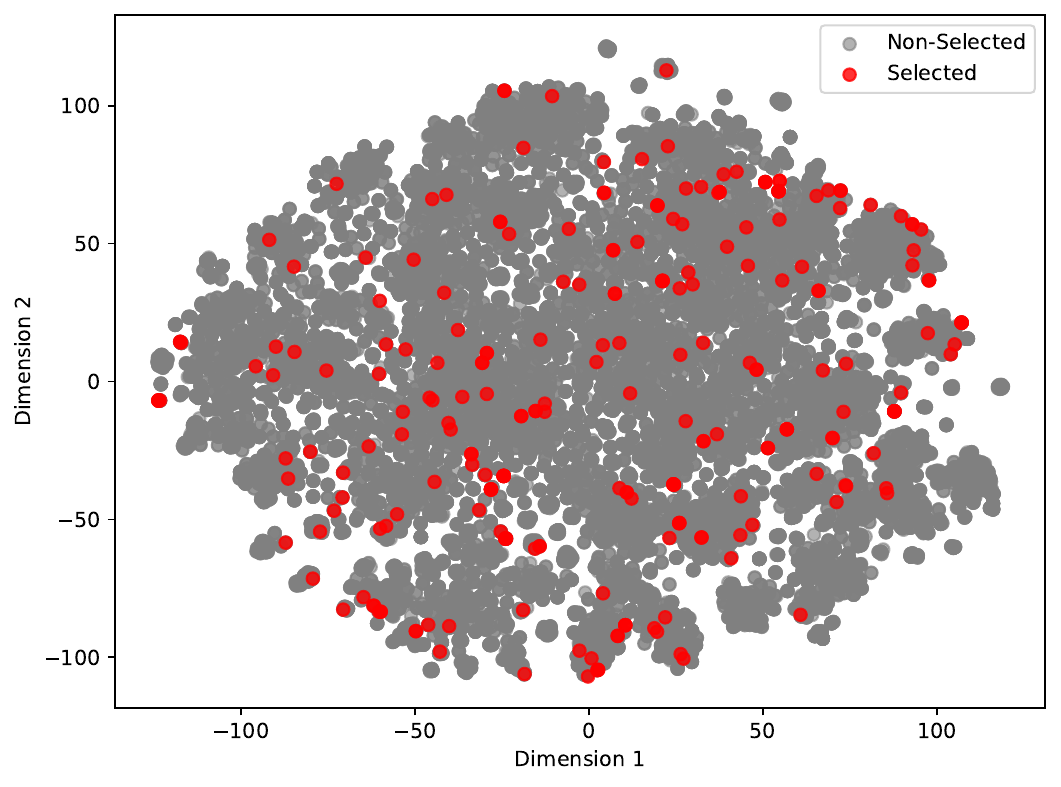}
        \caption{t-SNE Visualization}
        
    \end{subfigure}
    \begin{subfigure}[b]{0.3\linewidth}
        \centering
        \includegraphics[width=\textwidth]{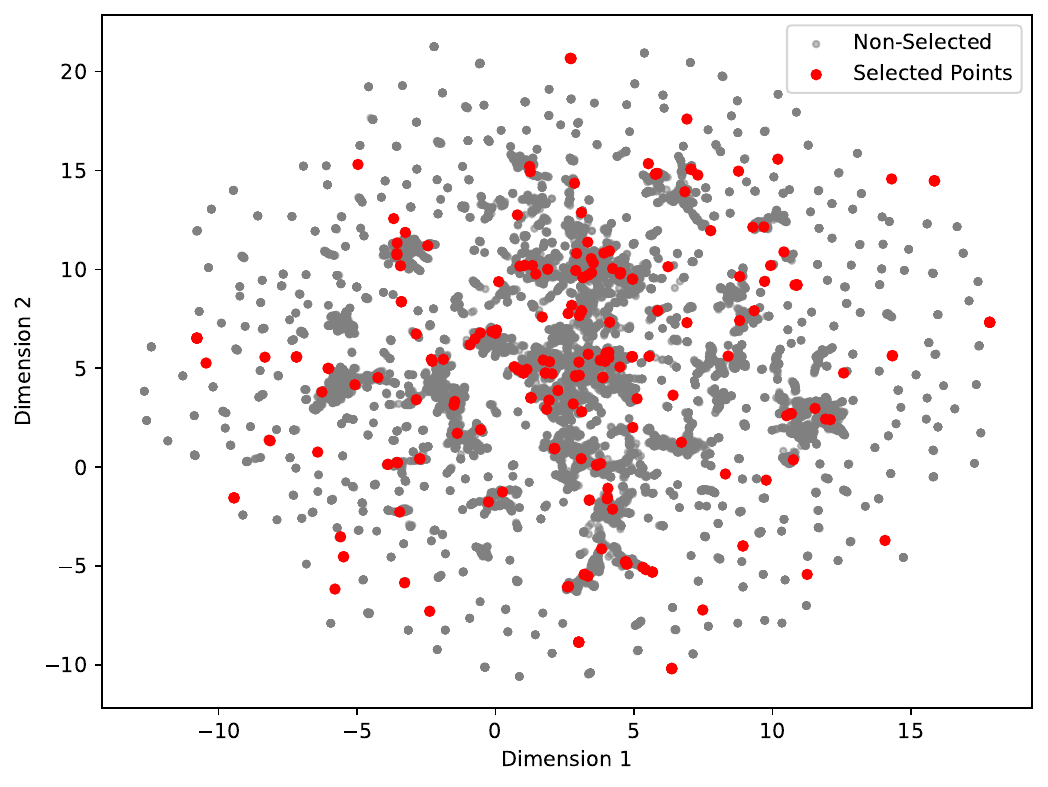}
        \caption{UMAP Visualization}
        
    \end{subfigure}

    \caption{Visualization of PCA, t-SNE, and UMAP projections for the Attribute Recognition task. The top row presents representations from the Multiple-Choice QA setting, while the bottom row corresponds to the Caption/Foil Classification setting.}
    \label{fig:sixfigsAR}
\end{figure*}

Figure \ref{fig:sixfigsAR} presents PCA, t-SNE, and UMAP projections for the Attribute Recognition task. In the MCQA setting, the Graph-Cut $\lambda$ was set to $50$. Due to the larger number of source samples in the Caption/Foil classification setting, $\lambda$ was increased to $200$.

\subsubsection{In-Depth Results}
The Attribute Recognition task in \benchmark proves moderately challenging for current multimodal large language models. While top models like InternVL2 and Qwen2.5 achieve strong performance in the 0-shot setting (up to 87.8\% accuracy), overall results highlight significant variation across models, particularly among smaller variants, indicating that the task requires nuanced visual-linguistic grounding. This difficulty is further confirmed by the pairwise accuracy evaluation, where models must rely on perplexity-based comparisons rather than classification, resulting in consistently lower scores.

Few-shot prompting does not consistently improve performance and often leads to degradation. Adding more examples (from 4-shot to 8-shot) yields minimal or even negative gains for many models, suggesting that longer prompts might introduce noise or distract from the core task. CoT prompting similarly fails to enhance performance; in several cases, it reduces accuracy. This indicates that current MLLMs are not well-optimized for multi-step visual reasoning and may struggle with reasoning chains that dilute focus from fine-grained attribute cues.

The effect of example selection—random versus semantically similar—appears minimal and model-dependent. While some models show slight improvements with similar examples, the gains are not systematic, highlighting limitations in models' ability to leverage contextual similarity effectively. These trends suggest that more examples or reasoning steps alone are insufficient; rather, improved representation learning and prompt understanding are necessary to better tackle fine-grained attribute recognition in multimodal settings.
\subsection{Visual Object Recognition}
\label{appd_visual_or}
The Visual Object Recognition task evaluates a model’s ability to identify objects depicted in an image. Each instance presents an image and a multiple-choice question targeting a specific object, with distractors carefully selected to test the model’s visual discrimination. This task serves as a baseline for assessing general visual recognition capabilities in multimodal models.
\subsubsection{Example Selection}
\begin{figure*}[ht!]
    \centering
    \begin{subfigure}[b]{0.3\linewidth}
        \centering
        \includegraphics[width=\textwidth]{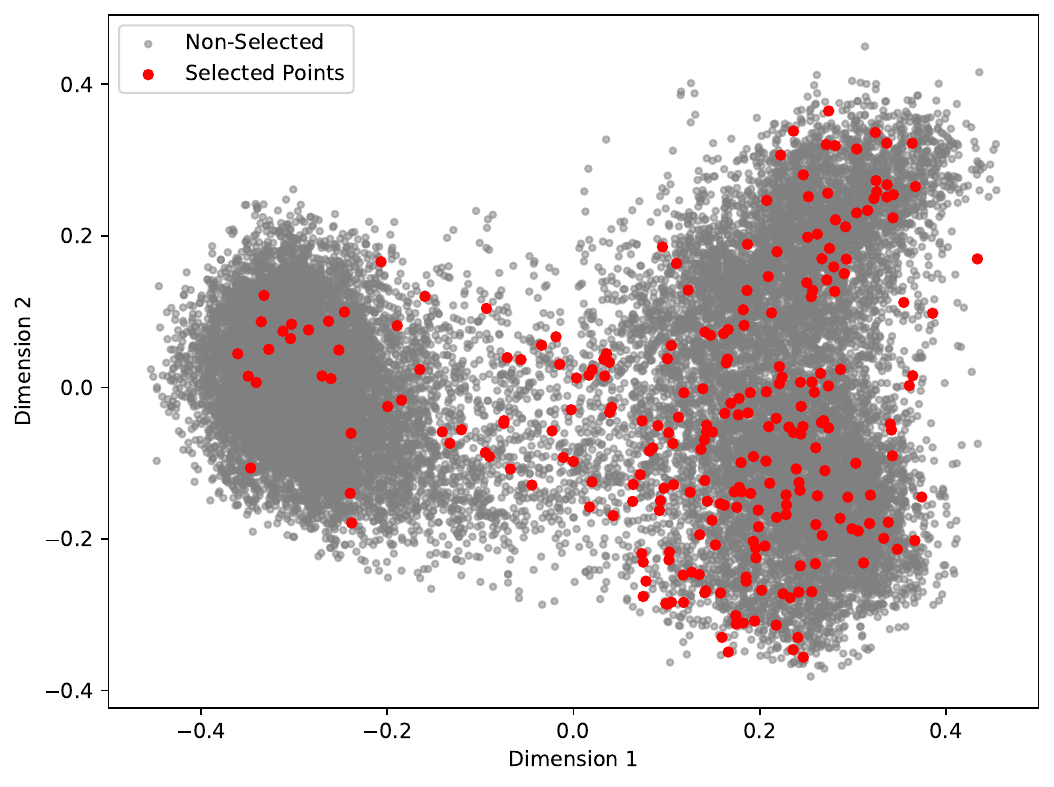}
        \caption{PCA Projection}
        
    \end{subfigure}
    \begin{subfigure}[b]{0.3\linewidth}
        \centering
        \includegraphics[width=\textwidth]{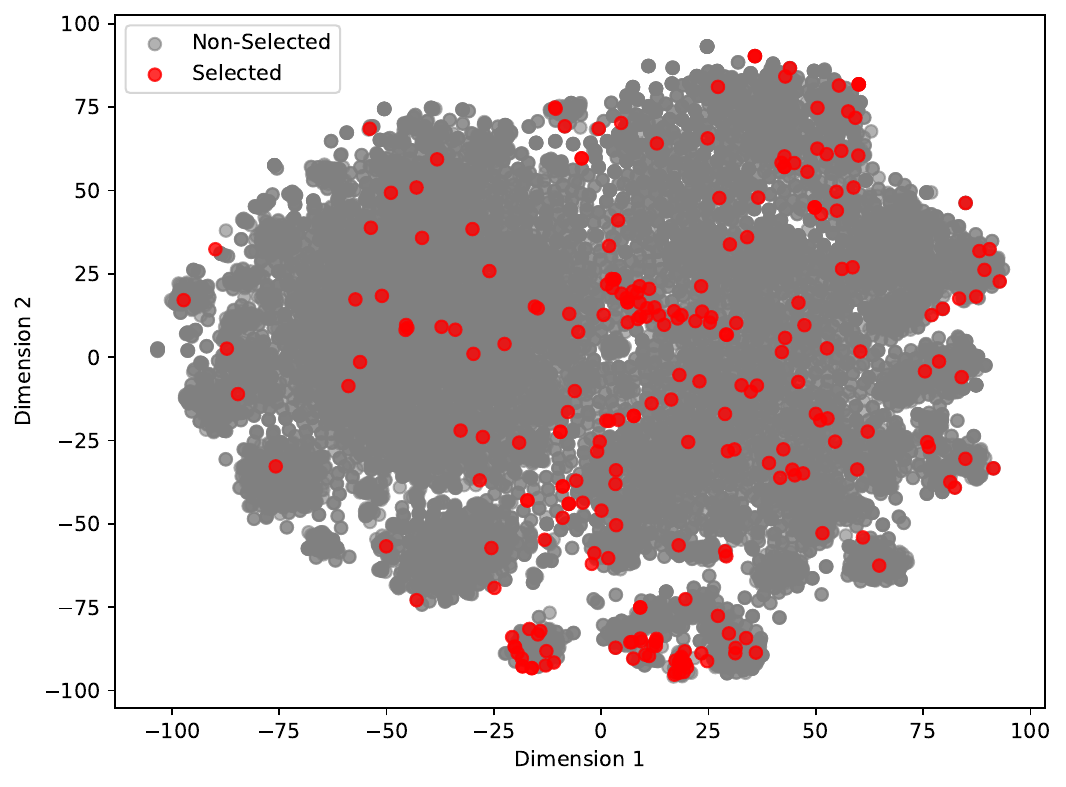}
        \caption{t-SNE Visualization}
        
    \end{subfigure}
    \begin{subfigure}[b]{0.3\linewidth}
        \centering
        \includegraphics[width=\textwidth]{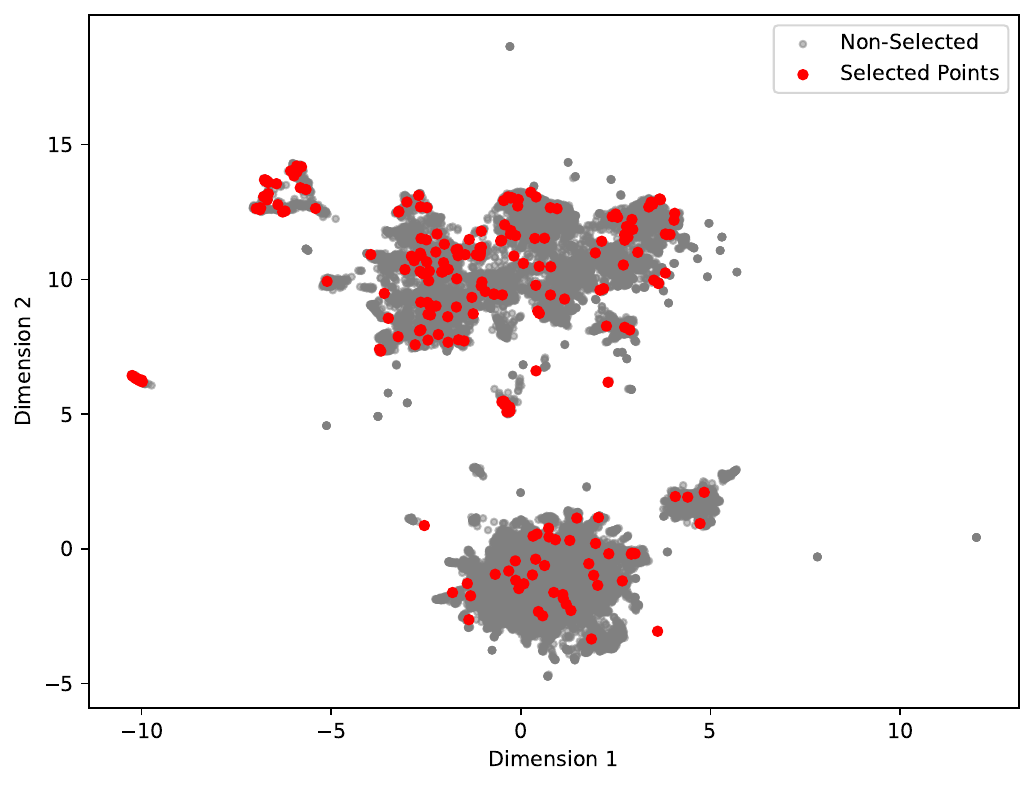}
        \caption{UMAP Visualization}
        
    \end{subfigure}

    \vspace{0.5cm} %

    \begin{subfigure}[b]{0.3\linewidth}
        \centering
        \includegraphics[width=\textwidth]{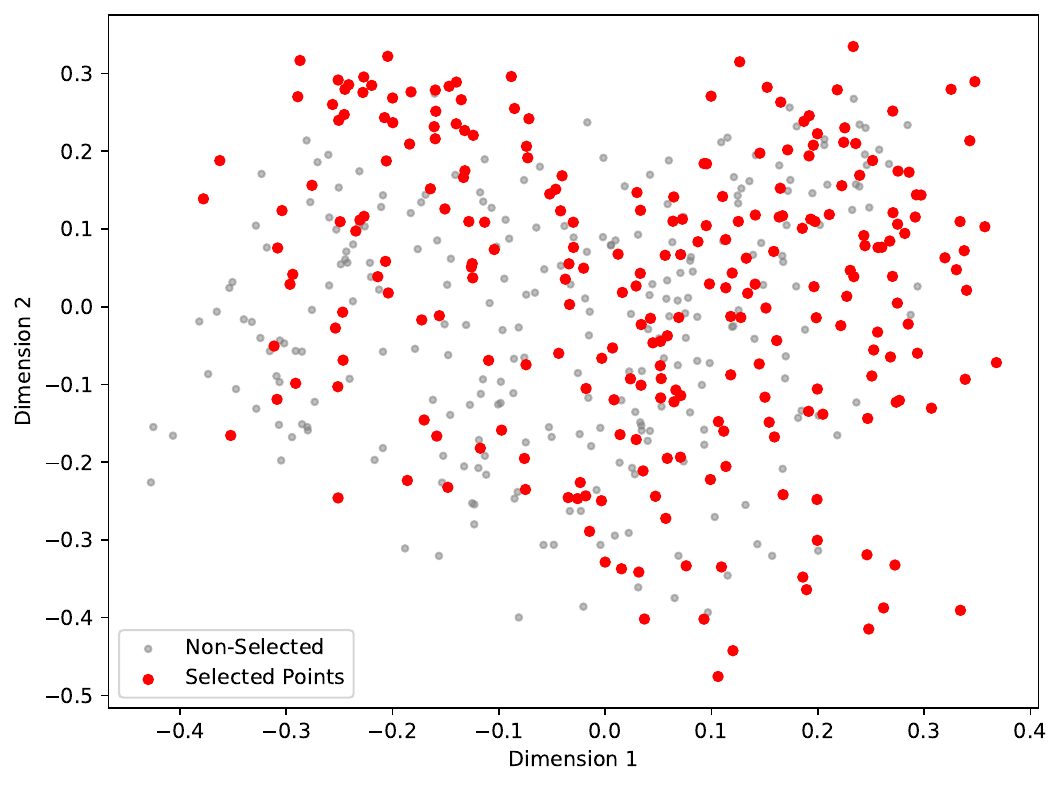}
        \caption{PCA Projection}
        
    \end{subfigure}
    \begin{subfigure}[b]{0.3\linewidth}
        \centering
        \includegraphics[width=\textwidth]{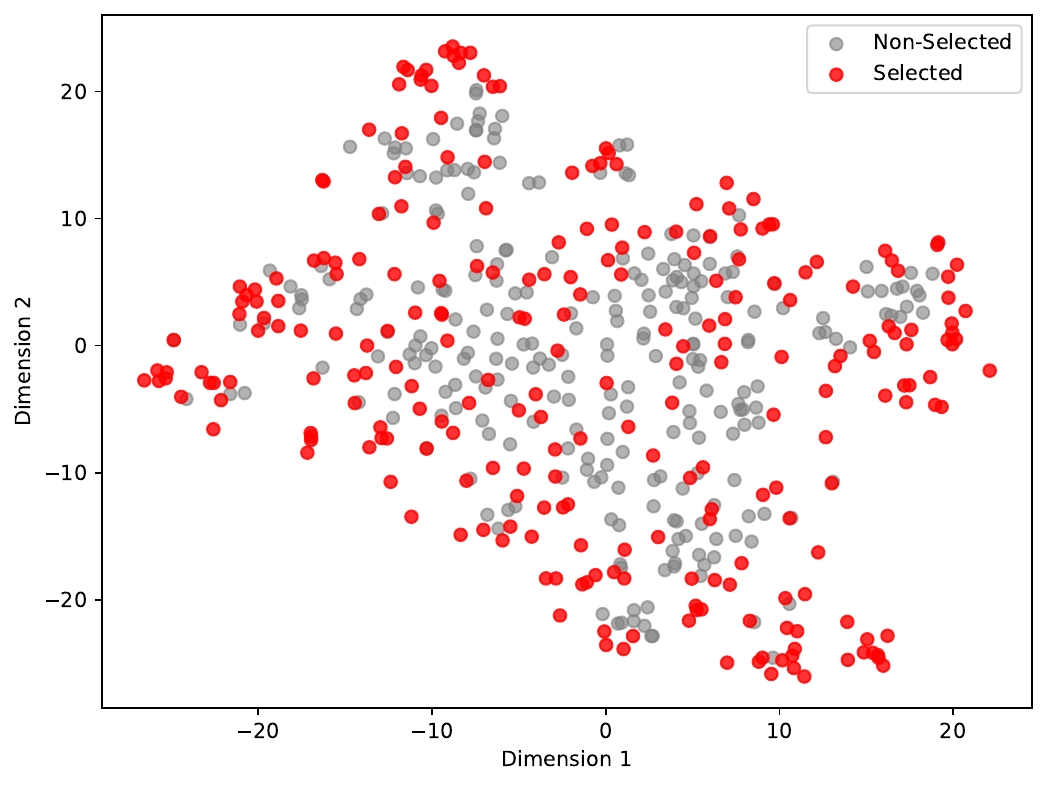}
        \caption{t-SNE Visualization}
        
    \end{subfigure}
    \begin{subfigure}[b]{0.3\linewidth}
        \centering
        \includegraphics[width=\textwidth]{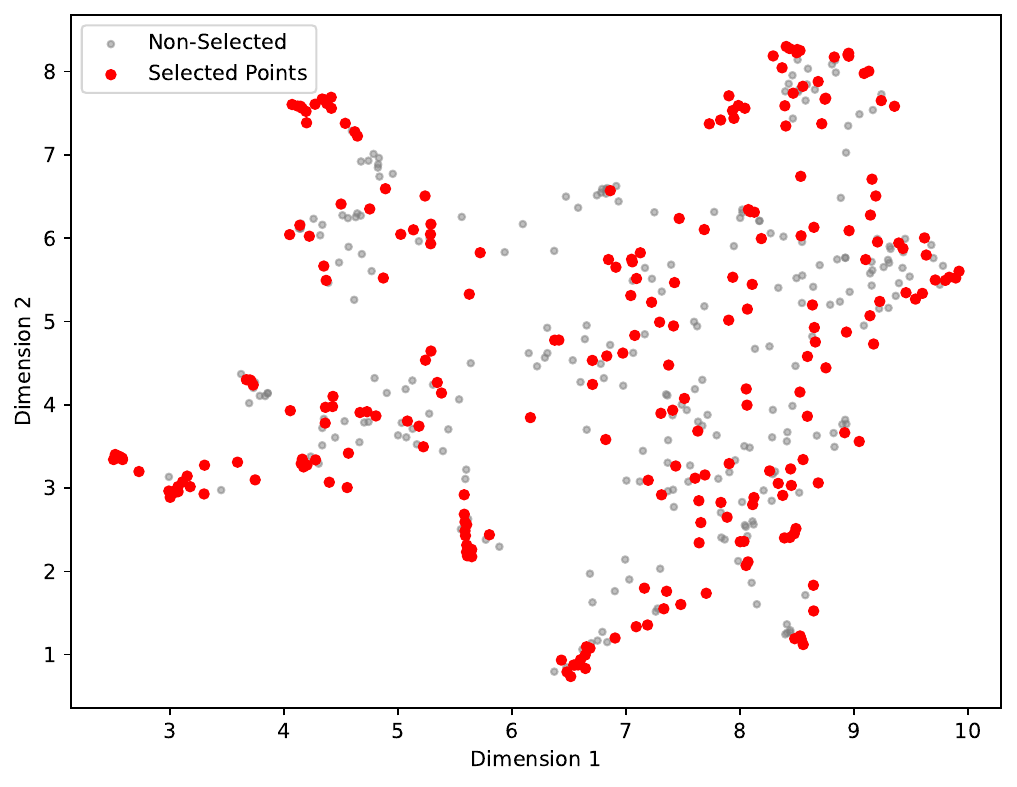}
        \caption{UMAP Visualization}
        
    \end{subfigure}

    \caption{Visualization of PCA, t-SNE, and UMAP projections for the Visual Object Recognition task. The top row presents representations from the Multiple-Choice QA setting, while the bottom row corresponds to the Caption/Foil classification setting.}
    \label{fig:sixfigsVOR}
\end{figure*}

Figure \ref{fig:sixfigsVOR} presents PCA, t-SNE, and UMAP projections for the Visual Object Recognition task. In the Caption/Foil classification, the Graph-Cut $\lambda$ was set to $50$. Due to the larger number of source samples in the MCQA setting, $\lambda$ was increased to $200$.

\subsubsection{In-Depth Results}
The Visual Object Recognition task in \benchmark appears more tractable for current multimodal LLMs compared to Attribute Recognition. Most high-capacity models achieve strong 0-shot performance—e.g., InternVL2 variants and Qwen2.5 series consistently exceed 85\%—indicating that object categorization is well-aligned with pretraining objectives of these models. However, performance gains from few-shot prompting are limited. While marginal improvements are observed for some models in 4-shot and 8-shot random settings, similar-shot prompting offers no consistent advantage, and in some cases, leads to slight regressions. This suggests that while the task is easier, models do not robustly benefit from additional in-context examples.

The lack of significant gains with increasing number of shots likely stems from redundancy and suboptimal utilization of added context. Many models (e.g., Qwen2-VL-3B, Idefics3-8B) exhibit plateauing or declining performance from 4-shot to 8-shot, hinting at a saturation point beyond which additional examples may introduce noise or context overload. Furthermore, the minimal difference between random and similar example selection indicates that models may not be leveraging semantic similarity effectively. This can be attributed to a lack of targeted training on few-shot reasoning or insufficient sensitivity to fine-grained visual similarity cues in demonstrations.

Unlike more abstract or relational tasks, object recognition typically relies on identifying concrete and discriminative visual features, which current MLLMs handle well in a zero-shot setting. However, the failure of CoT reasoning to be tested here is consistent with the general trend seen in Attribute Recognition—extended reasoning does not align with this task’s visual nature and may distract rather than aid. Overall, while object recognition is a relatively easy task for MLLMs, their limited ability to capitalize on prompting strategies highlights a gap in generalization and in-context learning capabilities that warrants further investigation.

\subsection{Plurality Recognition}
\label{appd_plurality_r}
The Plurality Recognition task probes a model’s understanding of number and count-based distinctions. Given an image and a question referring to objects in either singular or plural form, the model must correctly identify whether the linguistic reference matches the visual content. This task is designed to test models’ grounding of quantificational language in visual scenes.
\subsubsection{Example Selection}
\begin{figure*}[ht!]
    \centering
    \begin{subfigure}[b]{0.3\linewidth}
        \centering
        \includegraphics[width=\textwidth]{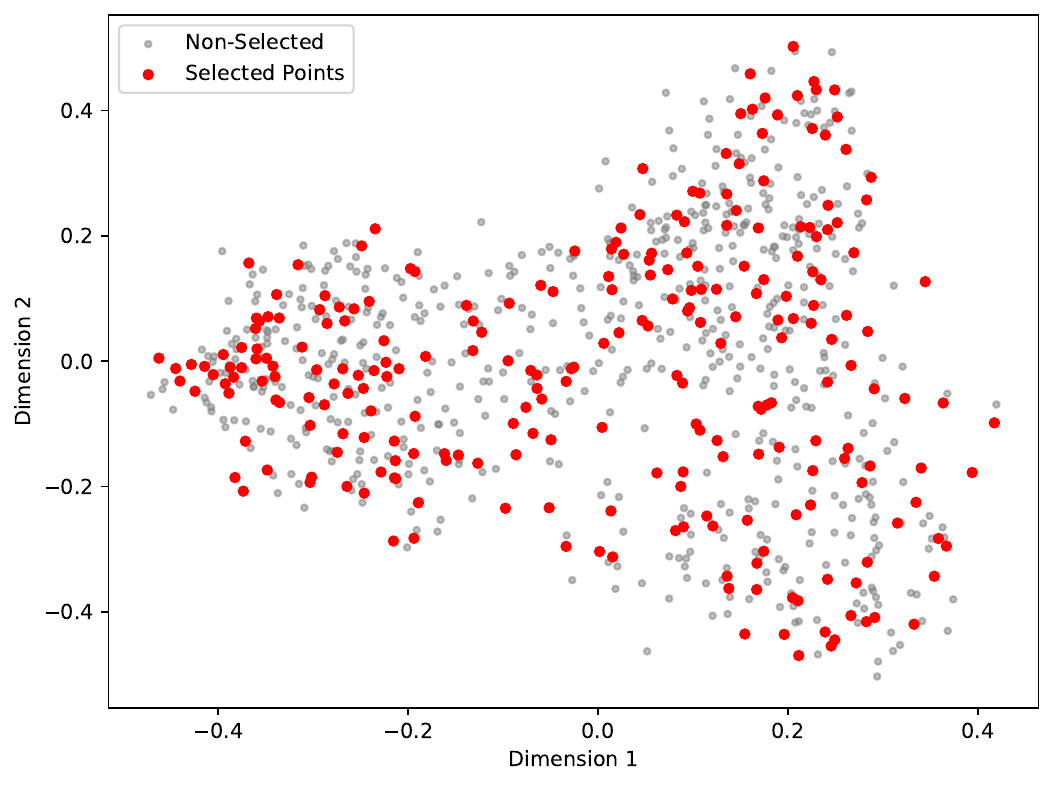}
        \caption{PCA Projection}
        
    \end{subfigure}
    \begin{subfigure}[b]{0.3\linewidth}
        \centering
        \includegraphics[width=\textwidth]{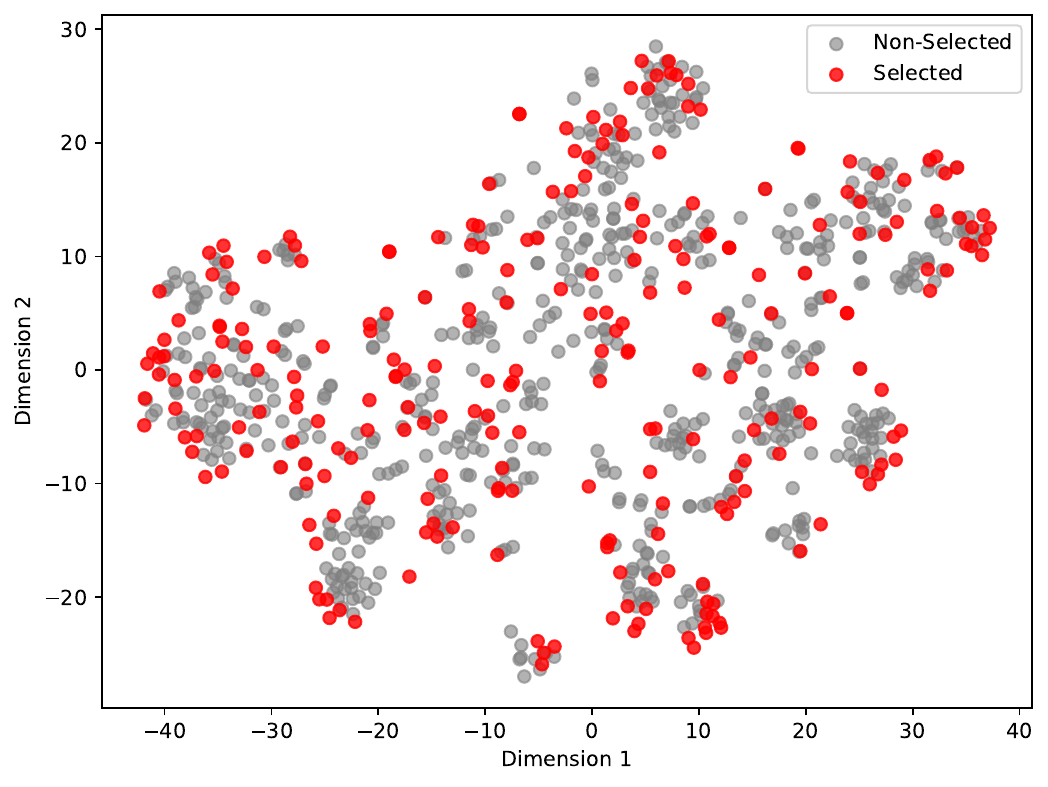}
        \caption{t-SNE Visualization}
        
    \end{subfigure}
    \begin{subfigure}[b]{0.3\linewidth}
        \centering
        \includegraphics[width=\textwidth]{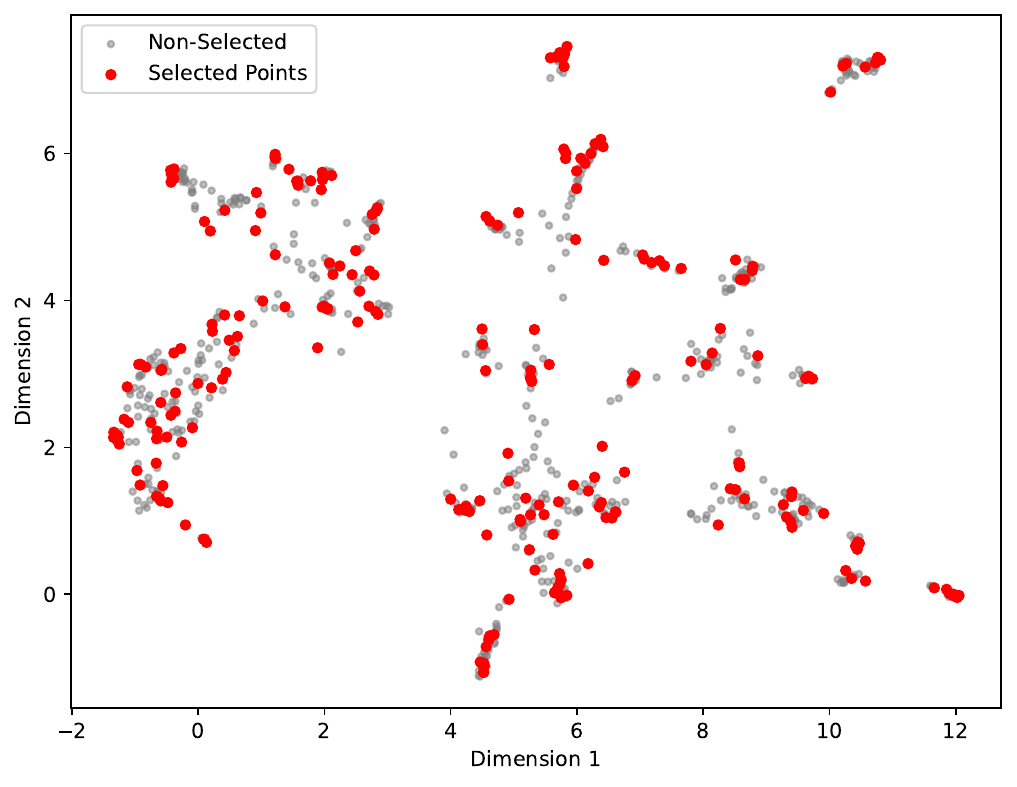}
        \caption{UMAP Visualization}
        
    \end{subfigure}

    \caption{Visualization of PCA, t-SNE, and UMAP projections for the Plurality Recognition task. }
    \label{fig:sixfigsPR}
\end{figure*}

Figure \ref{fig:sixfigsPR} presents PCA, t-SNE, and UMAP projections for the Plurality Recognition task. The Graph-Cut $\lambda$ was set to $50$.

\subsubsection{In-Depth Results}
The Plurality Recognition task in \benchmark appears relatively easy. For example, Qwen2.5-VL-7B-Instruct achieves a high accuracy of ~91\% in the few-shot setting, and a wide range of other models—including Phi-3.5-Vision-Instruct and LLaVA-NeXT-Interleave-Qwen-7B—also demonstrate strong performance. These results suggest that current MLLMs are generally capable of handling plural reasoning effectively, with limited room for further improvement on this task.

Few-shot prompting helps, but its impact varies with example selection and model strength. Using similar examples provides clear gains over random selection, especially in larger instruction-tuned models. However, increasing the number of examples from 4 to 8 often leads to stagnation or slight regression, suggesting potential information overload or reduced prompt relevance. Importantly, newer or larger models do not always outperform older or smaller ones across settings, contradicting expected scaling trends and highlighting that plural understanding remains a weak point even in recent MLLMs.

CoT reasoning does not reliably improve performance and often leads to degradation, particularly in mid- and low-tier models. This suggests current MLLMs struggle to maintain coherent step-by-step reasoning in multimodal contexts. Pairwise accuracy results—based on perplexity comparisons—reinforce these findings, with strong models like InternVL2.5-8B maintaining consistency, while others show erratic scoring behavior. Overall, these results emphasize that Plurality Recognition is a challenging task, and solving it requires improvements not only in scale or instruction tuning, but also in the core reasoning and alignment mechanisms of MLLMs.
\subsection{Object Counting}
\label{appd_object_c}
The Object Counting task requires models to infer the exact number of target objects in an image. Each instance includes a visual input and a counting-based multiple-choice question, where distractors reflect common counting errors. The task evaluates fine-grained object localization, subitizing, and the ability to resist visual distractors or occlusions.
\subsubsection{Example Selection}
\begin{figure*}[ht!]
    \centering
    \begin{subfigure}[b]{0.3\linewidth}
        \centering
        \includegraphics[width=\textwidth]{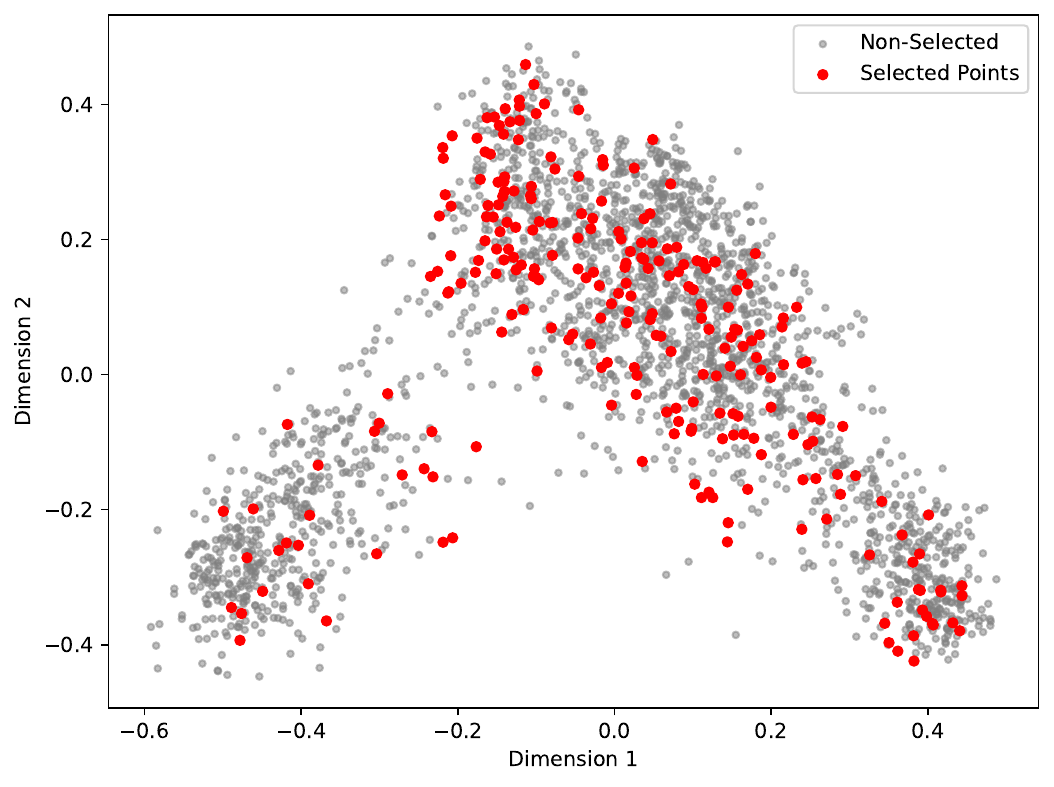}
        \caption{PCA Projection}
        
    \end{subfigure}
    \begin{subfigure}[b]{0.3\linewidth}
        \centering
        \includegraphics[width=\textwidth]{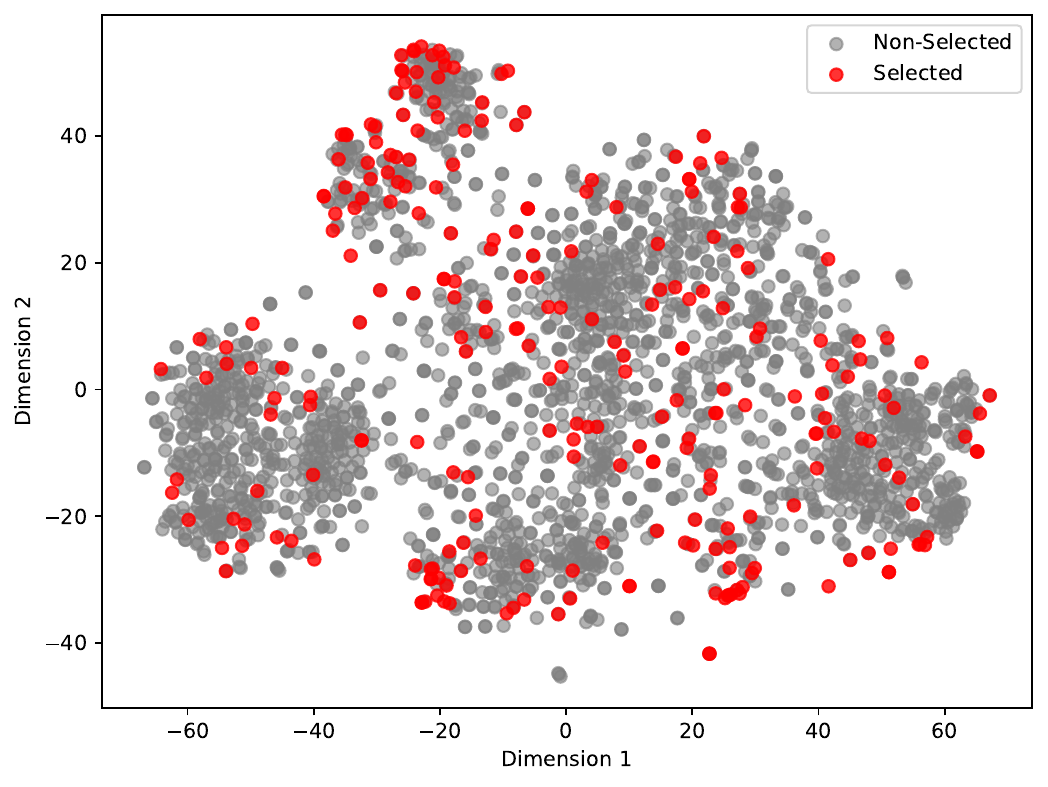}
        \caption{t-SNE Visualization}
        
    \end{subfigure}
    \begin{subfigure}[b]{0.3\linewidth}
        \centering
        \includegraphics[width=\textwidth]{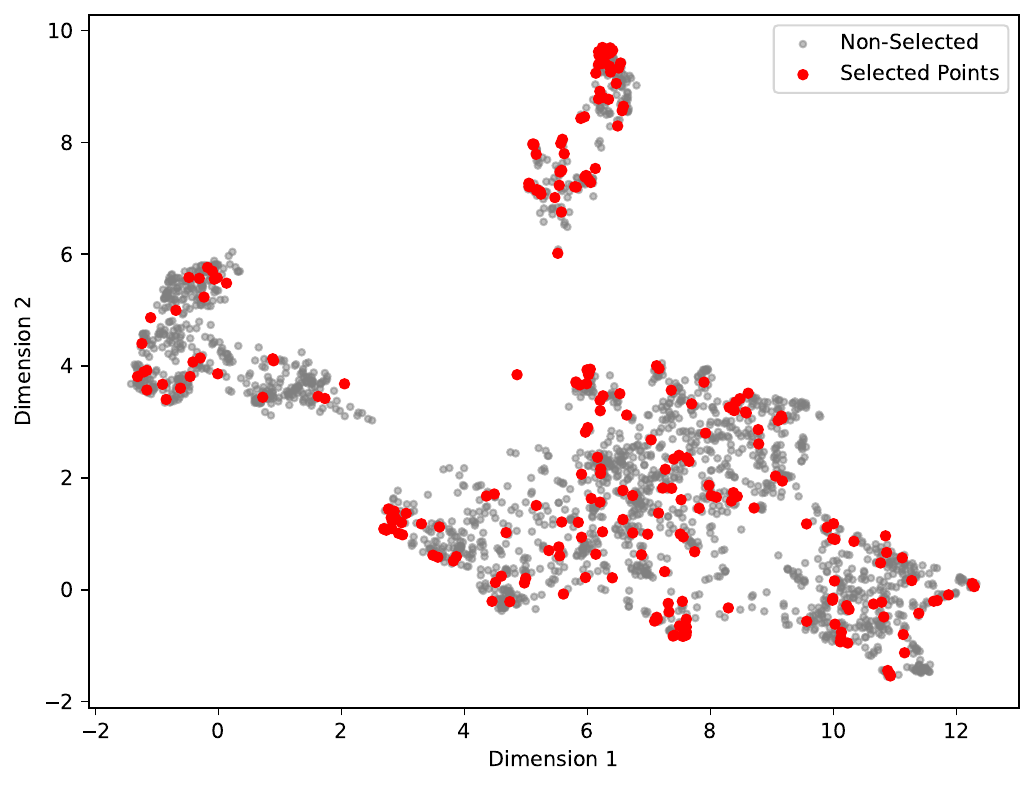}
        \caption{UMAP Visualization}
        
    \end{subfigure}

    \vspace{0.5cm} %

    \begin{subfigure}[b]{0.3\linewidth}
        \centering
        \includegraphics[width=\textwidth]{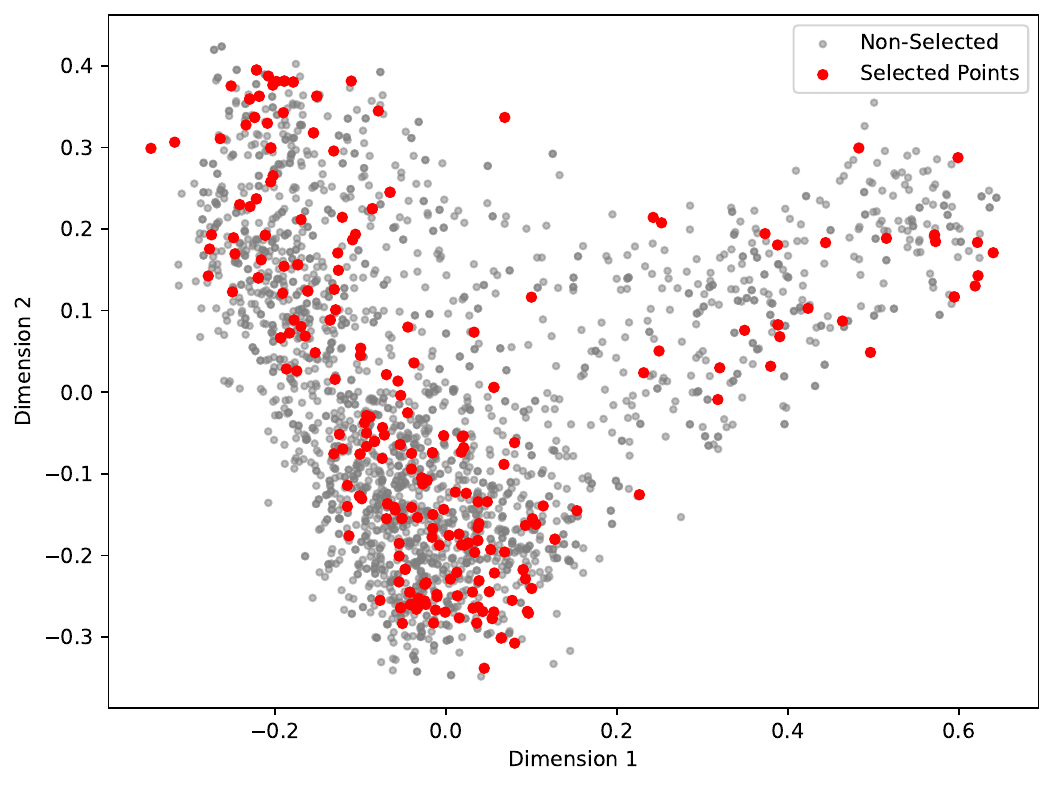}
        \caption{PCA Projection}
        
    \end{subfigure}
    \begin{subfigure}[b]{0.3\linewidth}
        \centering
        \includegraphics[width=\textwidth]{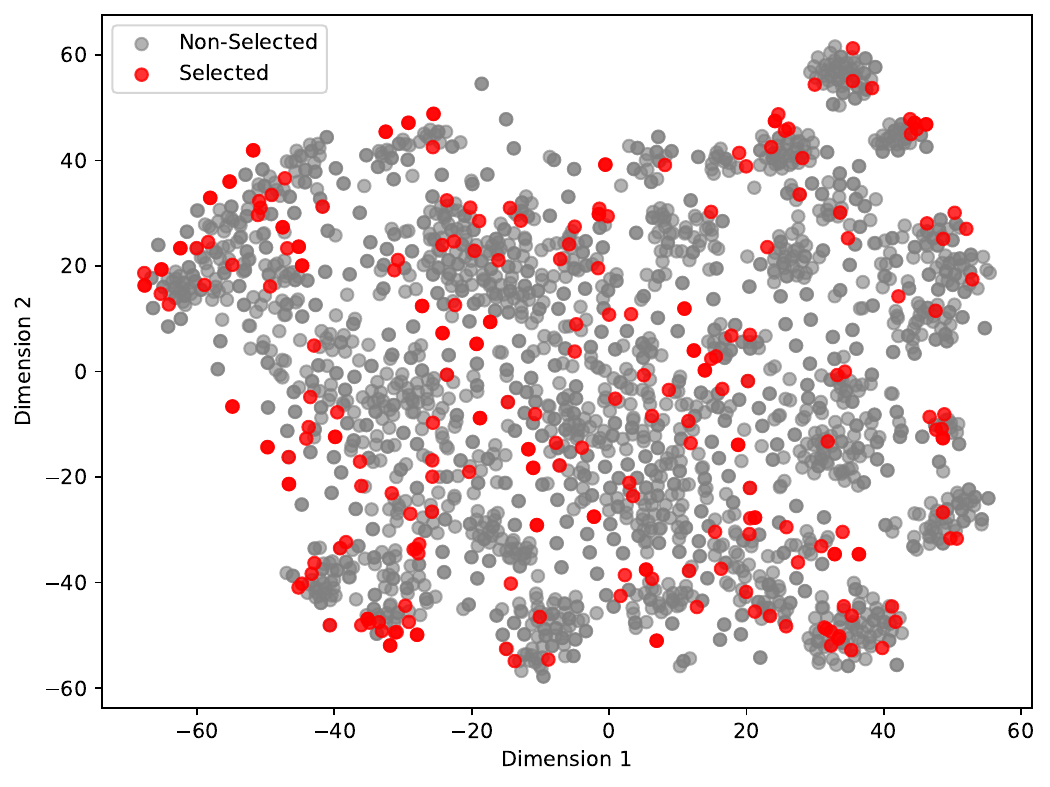}
        \caption{t-SNE Visualization}
        
    \end{subfigure}
    \begin{subfigure}[b]{0.3\linewidth}
        \centering
        \includegraphics[width=\textwidth]{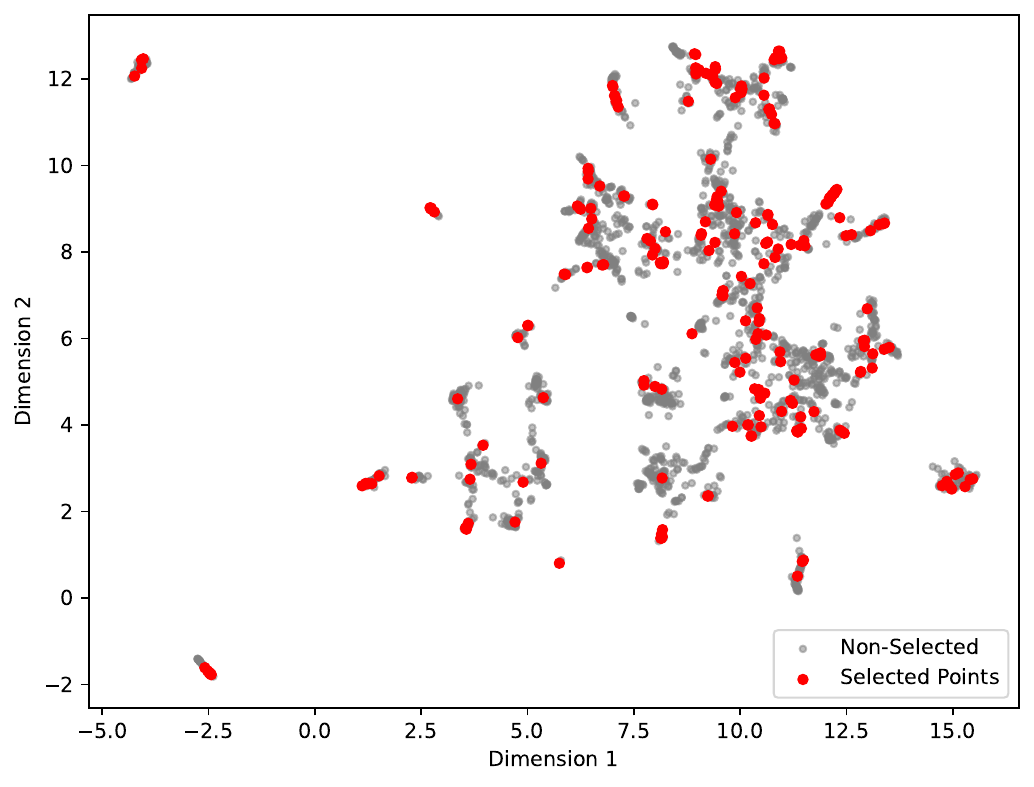}
        \caption{UMAP Visualization}
        
    \end{subfigure}

    \caption{Visualization of PCA, t-SNE, and UMAP projections for the Object Counting task. The top row presents representations from the Multiple-Choice QA setting, while the bottom row corresponds to the Caption/Foil Classification setting.}
    \label{fig:sixfigsOC}
\end{figure*}

Figure \ref{fig:sixfigsOC} presents PCA, t-SNE, and UMAP projections for the Object Counting task. The Graph-Cut $\lambda$ was set to $50$.

\subsubsection{In-Depth Results}
The Object Counting task in FewMMBench should be considered hard. While some high-capacity models like Qwen2.5-VL-7B-Instruct and InternVL2.5-8B-MPO reach moderate accuracies (~74–77\%) in favorable settings, most other models—including newer or scaled variants—fail to surpass 65\%, with smaller or untuned models like Qwen2-VL-2B and LLaVA-NeXT-Interleave-Qwen-0.5B performing close to random. These patterns violate expected trends where newer or more powerful models should outperform older or smaller ones, suggesting that object counting remains a fundamentally challenging task for current MLLMs.

Few-shot prompting provides limited and inconsistent gains. While a small number of models benefit slightly from 4-shot or 8-shot setups (e.g., Qwen2.5-VL-7B-Instruct improves marginally from 74.6\% to 77.2\%), many models experience stagnant or even declining performance as the number of examples increases. This decay is particularly pronounced in models such as Phi-3.5-Vision-Instruct and InternVL2.5-4B, possibly due to context window noise, prompt overload, or lack of alignment between examples and the target query. Moreover, example selection plays a limited role: similar examples occasionally offer marginal improvements, but their effect is not consistent across model scales or architectures.

CoT prompting largely fails to enhance performance and often reduces it, particularly in mid-sized or instruction-tuned models. For instance, InternVL2.5-8B-MPO and Qwen2.5-VL-7B-Instruct show drops of 2–3 percentage points when CoT is added, highlighting a recurring issue with multimodal CoT—namely, that it can introduce irrelevant reasoning or hallucinations when visual understanding is weak. Pairwise accuracy scores support this conclusion, as most models—including those with instruction tuning—show low confidence and poor ranking consistency across choices. Taken together, these findings demonstrate that Object Counting remains unsolved for current MLLMs and points to a need for improved visual grounding, discrete quantity reasoning, and multimodal alignment strategies.
\subsection{Spatial Relations Understanding}
\label{appd_spatial_r_u}
This task assesses the model’s ability to reason about spatial configurations between entities in an image. Questions focus on relational terms such as “next to,” “above,” or “between,” and require accurate interpretation of spatial arrangements. The task isolates spatial reasoning from object recognition by varying object types and positions across instances.
\subsubsection{Example Selection}
Figure \ref{fig:sixfigsAR} presents PCA, t-SNE, and UMAP projections for the Spatial Relations Understanding task. In the MCQA setting, the Graph-Cut $\lambda$ was set to $50$. Due to the larger number of source samples in the Caption/Foil classification setting, $\lambda$ was increased to $200$.
\begin{figure*}[ht!]
    \centering
    \begin{subfigure}[b]{0.3\linewidth}
        \centering
        \includegraphics[width=\textwidth]{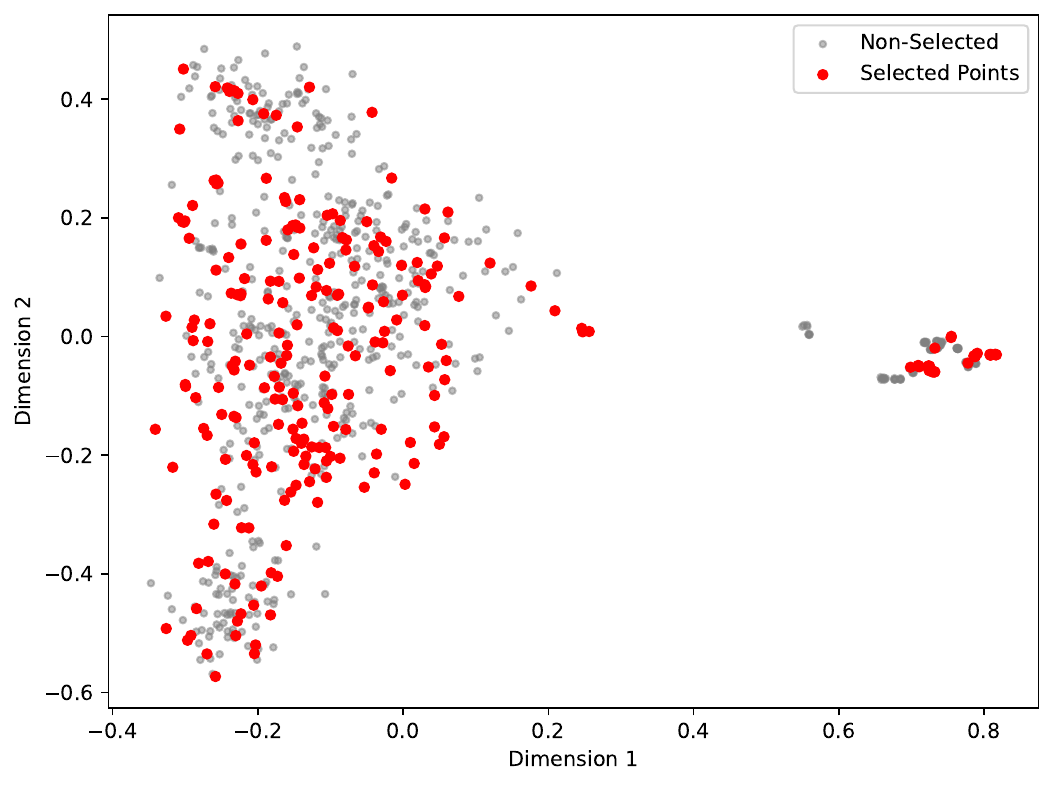}
        \caption{PCA Projection}
        
    \end{subfigure}
    \begin{subfigure}[b]{0.3\linewidth}
        \centering
        \includegraphics[width=\textwidth]{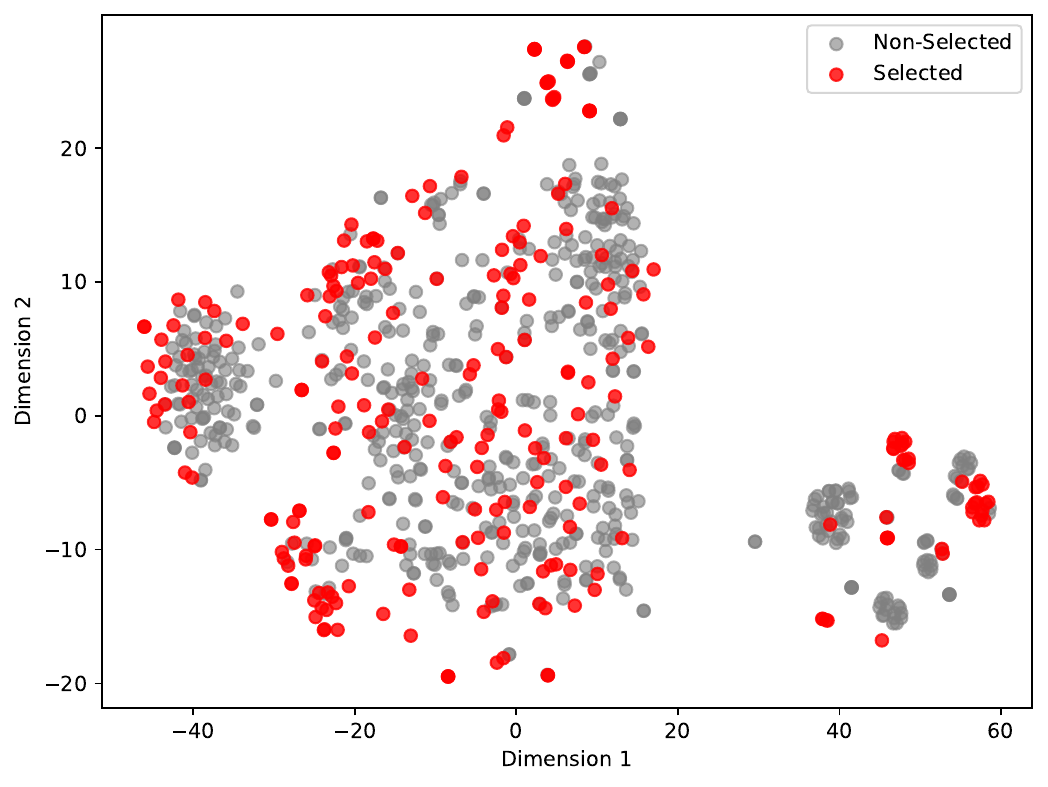}
        \caption{t-SNE Visualization}
        
    \end{subfigure}
    \begin{subfigure}[b]{0.3\linewidth}
        \centering
        \includegraphics[width=\textwidth]{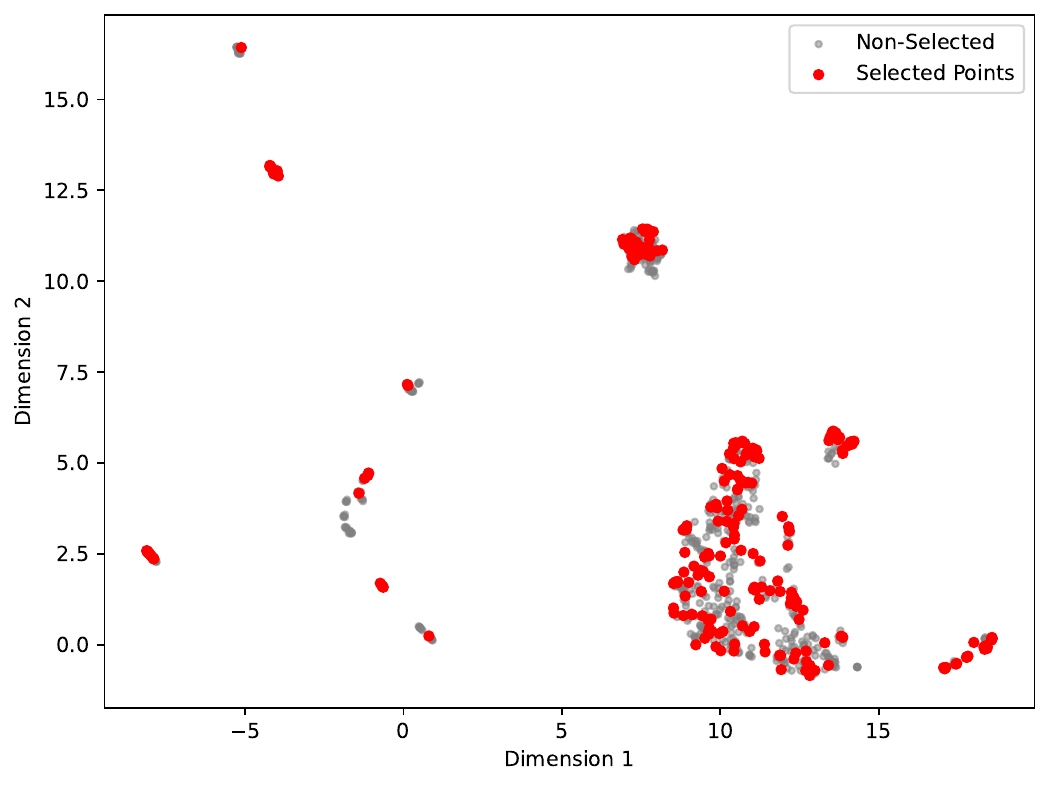}
        \caption{UMAP Visualization}
        
    \end{subfigure}

    \vspace{0.5cm} %

    \begin{subfigure}[b]{0.3\linewidth}
        \centering
        \includegraphics[width=\textwidth]{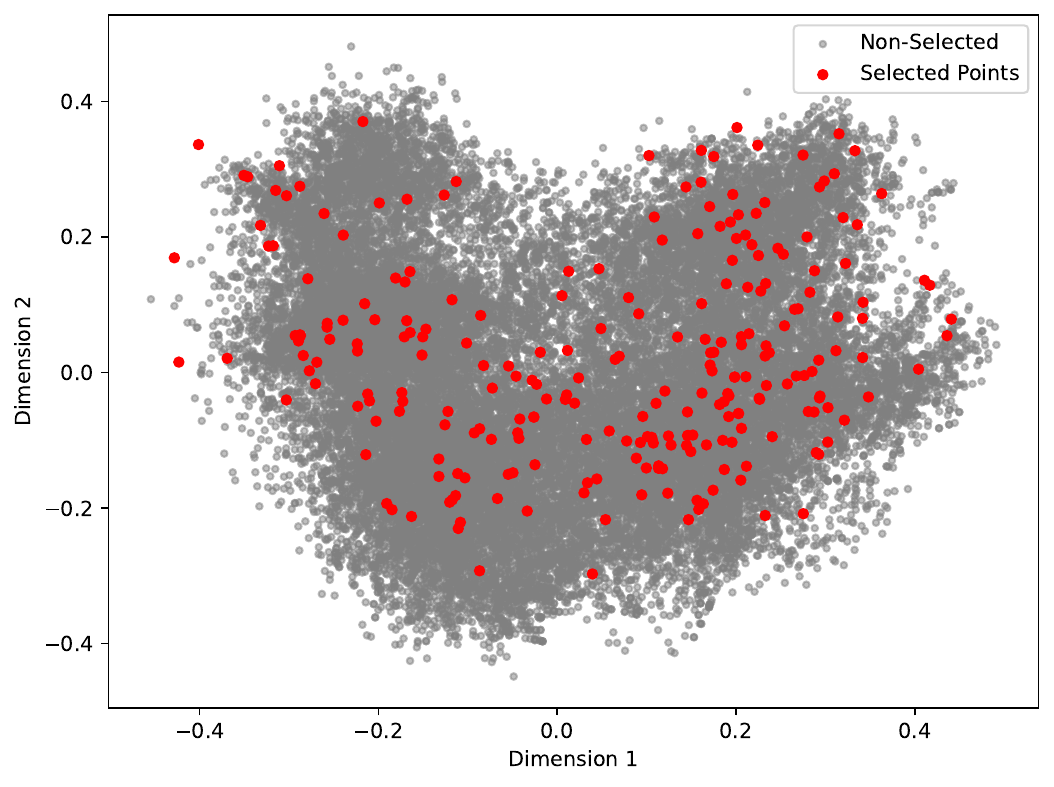}
        \caption{PCA Projection}
        
    \end{subfigure}
    \begin{subfigure}[b]{0.3\linewidth}
        \centering
        \includegraphics[width=\textwidth]{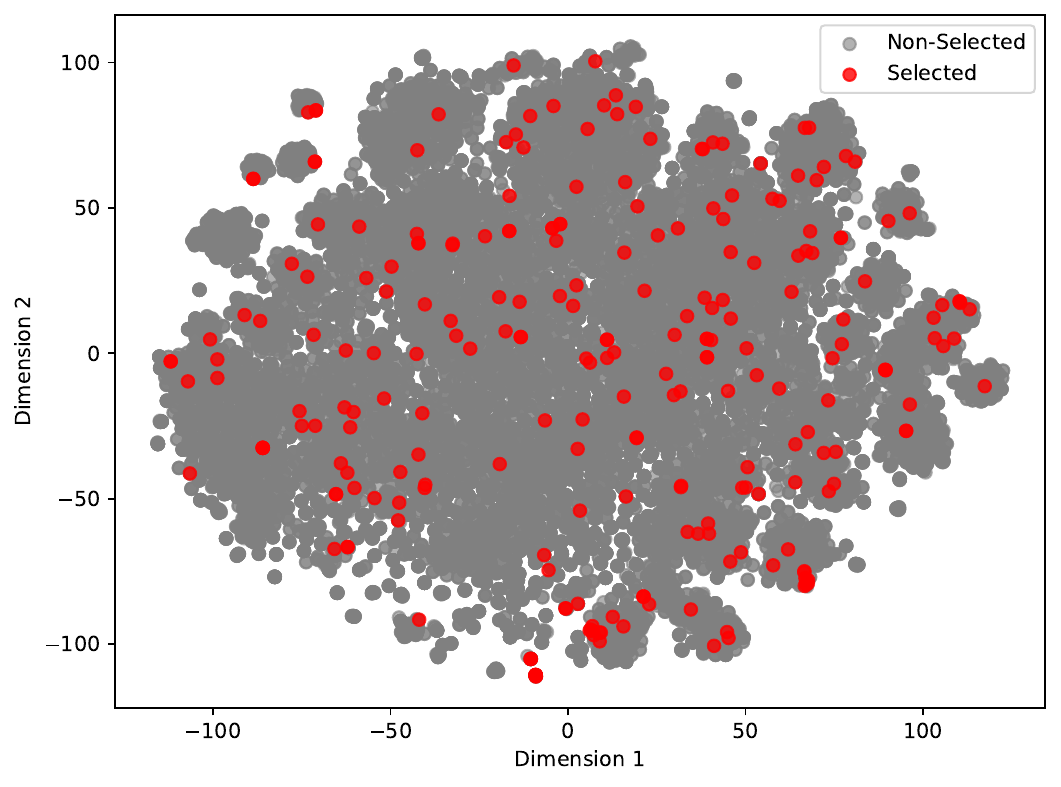}
        \caption{t-SNE Visualization}
        
    \end{subfigure}
    \begin{subfigure}[b]{0.3\linewidth}
        \centering
        \includegraphics[width=\textwidth]{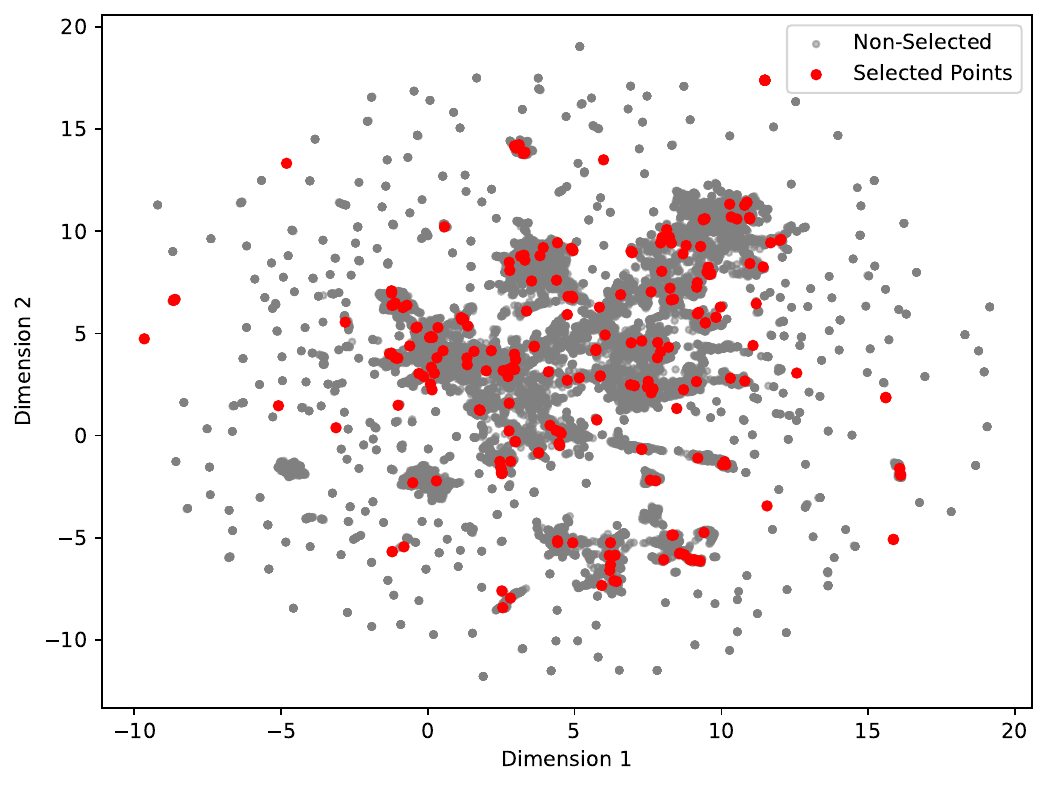}
        \caption{UMAP Visualization}
        
    \end{subfigure}

    \caption{Visualization of PCA, t-SNE, and UMAP projections for the Spatial Relations Understanding task. The top row presents representations from the Multiple-Choice QA setting, while the bottom row corresponds to the Caption/Foil classification setting.}
    \label{fig:sixfigsSRU}
\end{figure*}

\subsubsection{In-Depth Results}
The Spatial Relations Understanding task in \benchmark reveals moderate difficulty for current Multimodal Large Language Models (MLLMs). While top-tier models like Qwen2.5-VL-7B-Instruct and InternVL2.5-8B-MPO perform relatively well—reaching around 78–79\% accuracy—many other models, including some larger or newer variants, show inconsistent results or fail to generalize across prompting conditions. Models such as LLaVA-NeXT-Interleave-Qwen-0.5B and Qwen2-VL-2B, for instance, struggle to exceed 45\%, indicating that accurate spatial reasoning is not yet a broadly solved capability across architectures and scales.

Few-shot prompting provides variable gains. For stronger models, 4-shot and 8-shot settings offer some improvement, particularly under similar example selection. However, these gains are not guaranteed and often plateau or decline with additional shots, likely due to context saturation or less relevant support examples. Notably, example similarity generally outperforms random selection, though the margin is modest. This suggests that spatial reasoning may depend more on inherent model capabilities and less on example-driven induction, especially when support samples fail to capture fine-grained relational cues present in the target queries.

CoT reasoning does not reliably enhance performance and in some cases degrades it. Even in larger models, such as Qwen2.5-VL-7B-Instruct, CoT adds minimal value or introduces noise that disrupts spatial interpretation. Pairwise accuracy results reinforce this trend: while leading models show consistent ranking behaviors, others—regardless of scale or tuning—frequently misrank options, indicating weak internal representations of spatial structure. These observations highlight that although some MLLMs approach strong performance, spatial reasoning remains brittle and sensitive to prompting design, suggesting the need for more structured visual-linguistic training and better inductive alignment.
\subsection{Action Recognition}
\label{appd_action_r}
The Action Recognition task tests a model’s understanding of dynamic or implied events in static images. Each instance includes a question about what action is being performed, with options representing plausible but distinct actions. The task challenges models to integrate visual cues such as posture, object affordances, and scene context to infer actions.
\subsubsection{Example Selection}
\begin{figure*}[ht!]
    \centering
    \begin{subfigure}[b]{0.3\linewidth}
        \centering
        \includegraphics[width=\textwidth]{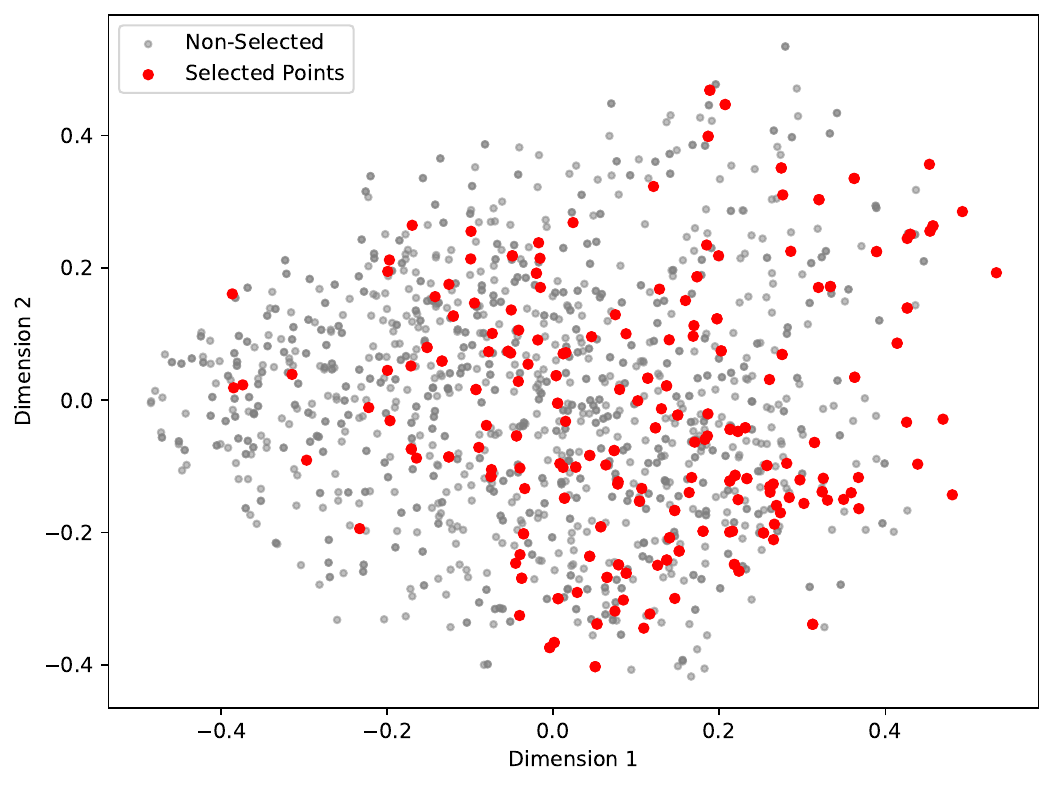}
        \caption{PCA Projection}
        
    \end{subfigure}
    \begin{subfigure}[b]{0.3\linewidth}
        \centering
        \includegraphics[width=\textwidth]{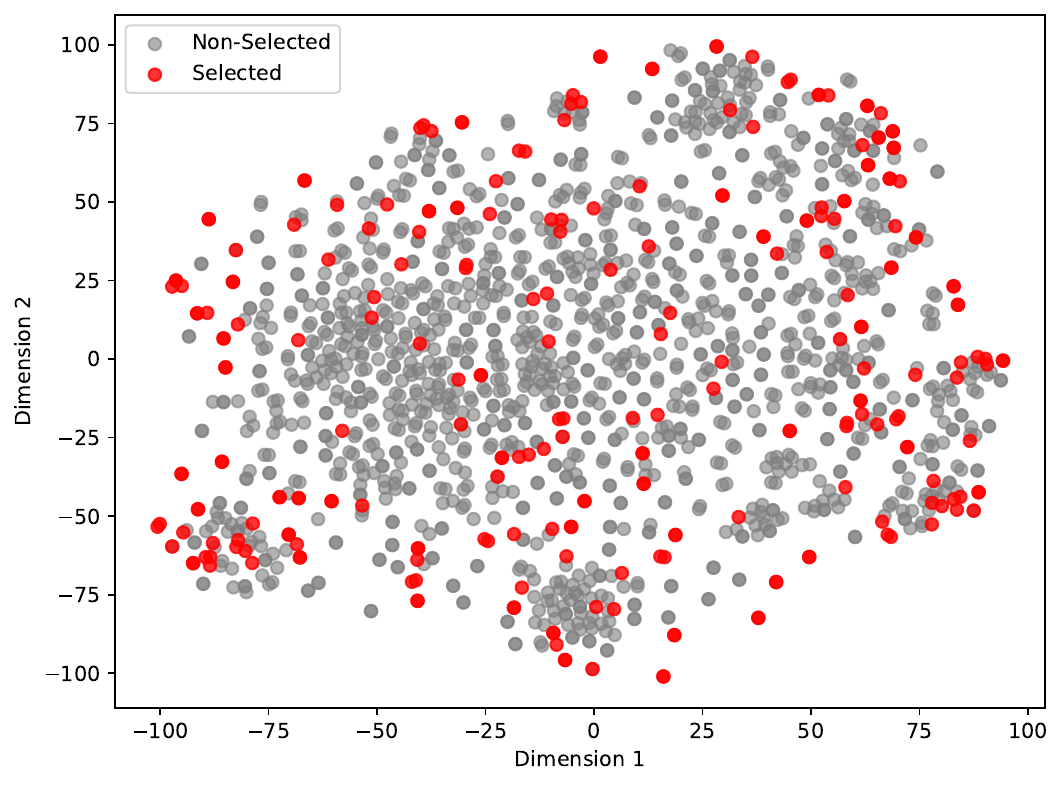}
        \caption{t-SNE Visualization}
        
    \end{subfigure}
    \begin{subfigure}[b]{0.3\linewidth}
        \centering
        \includegraphics[width=\textwidth]{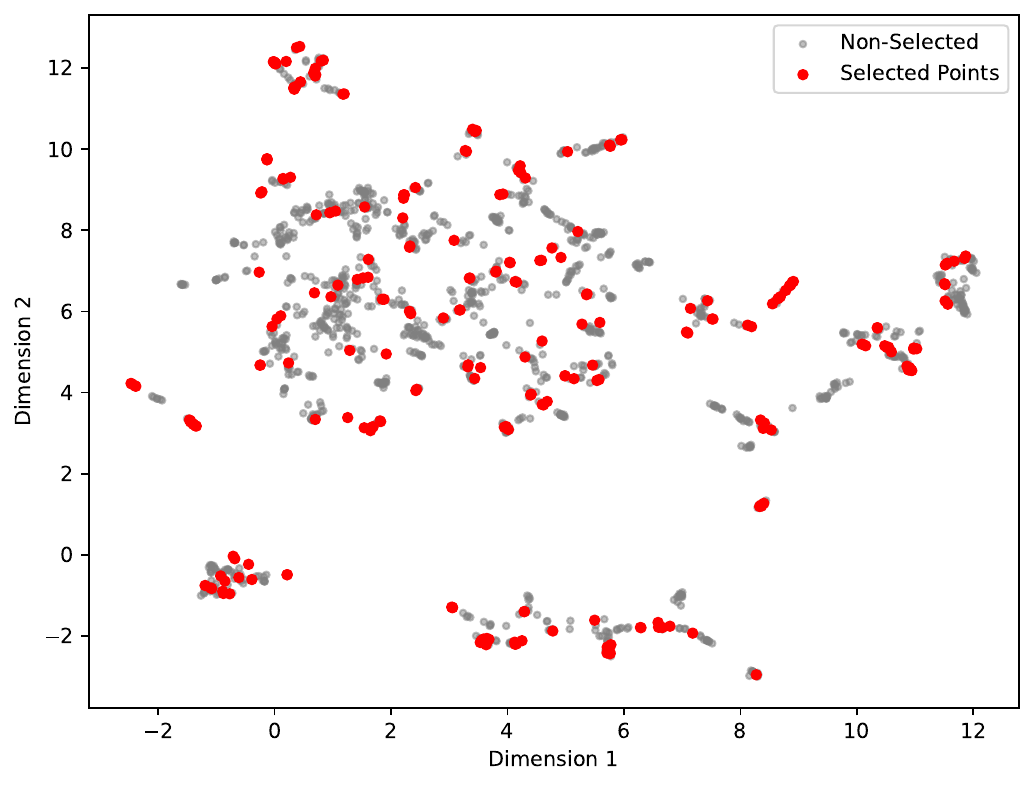}
        \caption{UMAP Visualization}
        
    \end{subfigure}

    \caption{Visualization of PCA, t-SNE, and UMAP projections for the Action Recognition task. }
    \label{fig:sixfigsAcR}
\end{figure*}

Figure \ref{fig:sixfigsAcR} presents PCA, t-SNE, and UMAP projections for the Action Recognition task. The Graph-Cut $\lambda$ was set to $50$.

\subsubsection{In-Depth Results}
The Action Recognition task in \benchmark appears to be relatively easy for well-aligned and instruction-tuned MLLMs. A broad range of high- and mid-capacity models—including Qwen2.5-VL-7B-Instruct, InternVL2.5-8B, Idefics2-8B, and Idefics3-8B—achieve over 90\% accuracy across various prompting conditions, with several models exceeding 96\%. These consistently strong results indicate that recognizing actions from images is well within the reach of current state-of-the-art MLLMs, provided they possess adequate visual grounding and instruction alignment.

Few-shot prompting, particularly when using representative and similar examples, provides additional gains, often enhancing already high-performing baselines. The Idefics model family notably benefits from this approach. In particular, Idefics3-8B, under the 8-shot setting, correctly answered all action recognition questions and achieved 100\% accuracy, demonstrating the effectiveness of well-designed demonstrations in reinforcing model reliability on visual reasoning tasks.

However, this improvement is not universal. Several models with limited visual grounding or suboptimal alignment—such as LLaVA-NeXT-Interleave-Qwen-0.5B and Qwen2-VL-2B—continue to struggle, even under favorable few-shot prompting, achieving only ~40–60\% accuracy. This highlights the persistent variability in model robustness and visual-textual understanding, even among models with comparable parameter counts.

Interestingly, the use of CoT prompting generally leads to performance degradation, especially in weaker models. CoT may introduce irrelevant or misleading reasoning steps in tasks that primarily require visual pattern recognition. Pairwise accuracy results further support these observations: strong models show consistent option ranking, while weaker ones exhibit erratic behaviors. Overall, while Action Recognition is tractable for advanced MLLMs, meaningful performance differences persist depending on training, instruction tuning, and visual grounding quality.

\subsection{Commonsense Reasoning}
\label{appd_commonsense_r}
This task evaluates whether models can combine visual understanding with external commonsense knowledge. Questions are grounded in the image but require inference beyond what is explicitly visible, such as typical object uses or human behavior patterns. The task tests multimodal models’ ability to incorporate prior knowledge in visually grounded reasoning.

\subsubsection{Example Selection}
\begin{figure*}[ht!]
    \centering
    \begin{subfigure}[b]{0.3\linewidth}
        \centering
        \includegraphics[width=\textwidth]{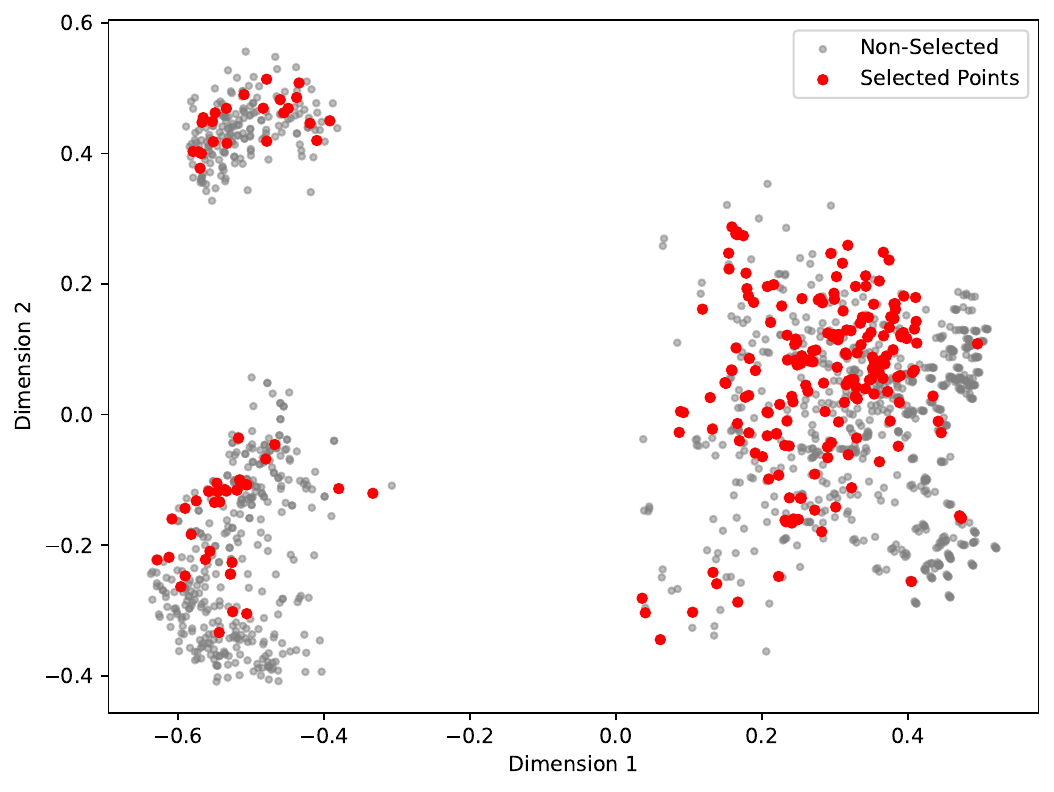}
        \caption{PCA Projection}
        
    \end{subfigure}
    \begin{subfigure}[b]{0.3\linewidth}
        \centering
        \includegraphics[width=\textwidth]{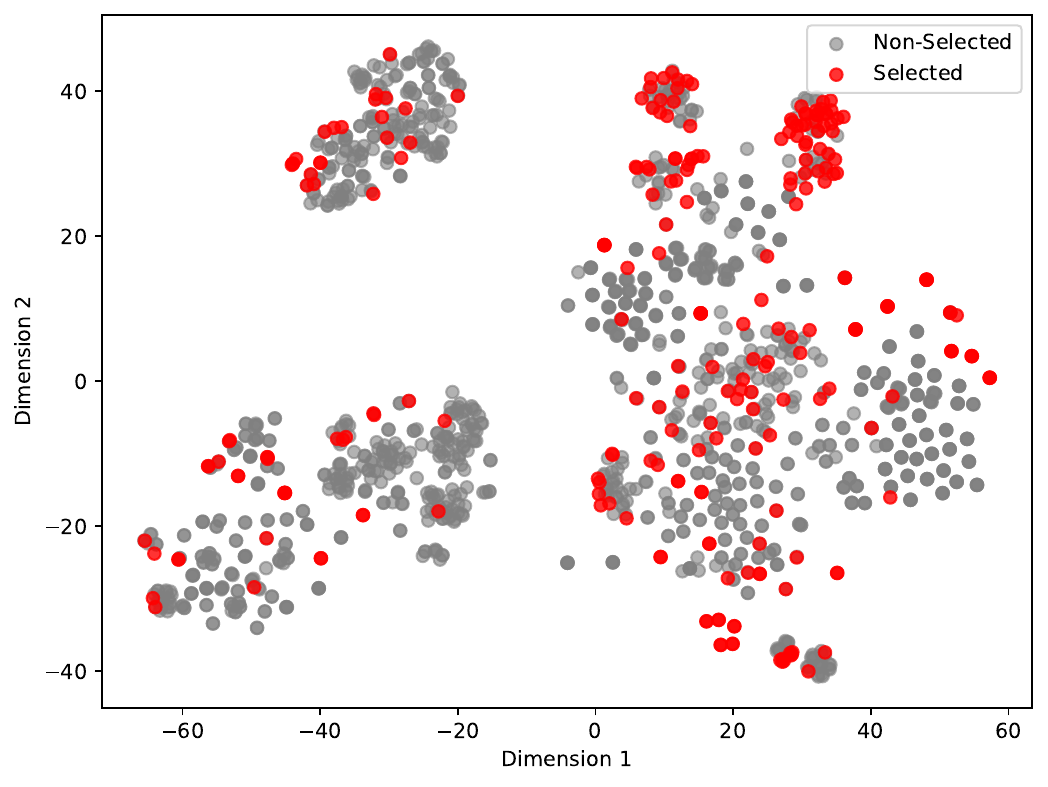}
        \caption{t-SNE Visualization}
        
    \end{subfigure}
    \begin{subfigure}[b]{0.3\linewidth}
        \centering
        \includegraphics[width=\textwidth]{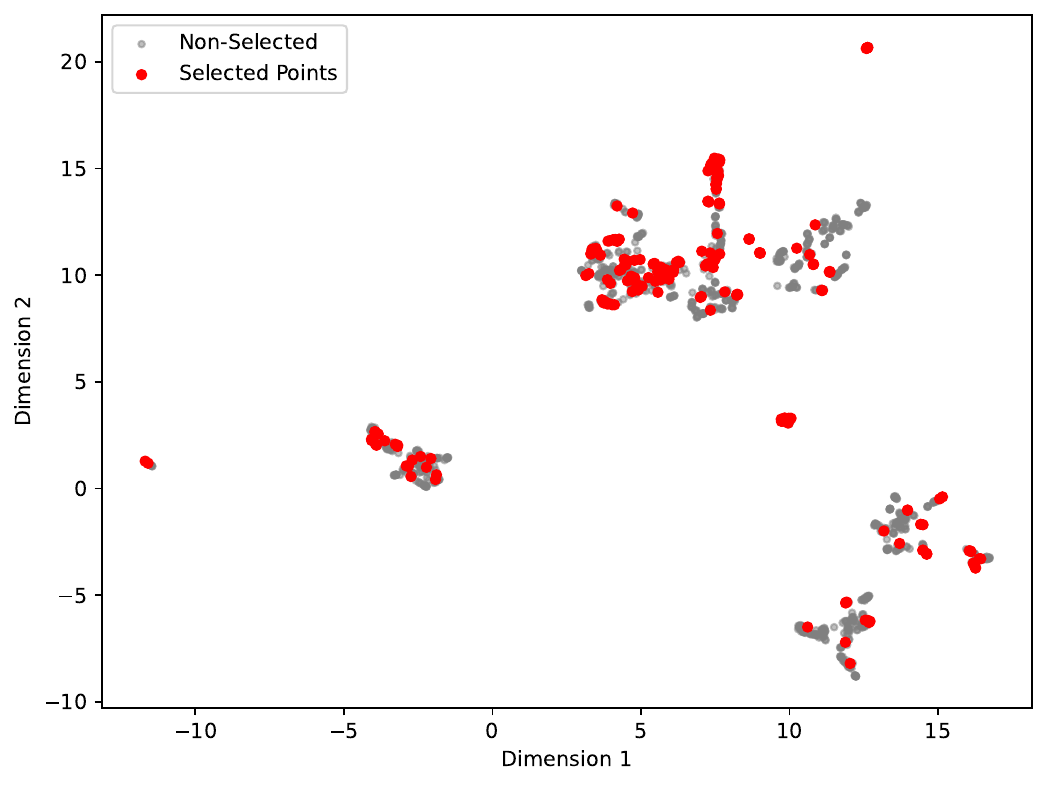}
        \caption{UMAP Visualization}
        
    \end{subfigure}

    \caption{Visualization of PCA, t-SNE, and UMAP projections for the Commonsense Reasoning task. }
    \label{fig:sixfigsCR}
\end{figure*}

Figure \ref{fig:sixfigsCR} presents PCA, t-SNE, and UMAP projections for the Commonsense Reasoning task. The Graph-Cut $\lambda$ was set to $50$.

\subsubsection{In-Depth Results}
Commonsense Reasoning in \benchmark presents a moderate challenge for current MLLMs. Top-performing models like Qwen2.5-VL-7B-Instruct and InternVL2.5-8B achieve around 80\% accuracy in favorable prompting conditions, but the performance varies widely across models. Notably, several newer or instruction-tuned models, such as Phi-3.5-Vision-Instruct and LLaVA-NeXT-Interleave-Qwen-7B, show only marginal gains over less capable baselines, suggesting that commonsense reasoning is not yet a reliably solved capability across architectures. This is further evidenced by persistent performance gaps between models of similar scale and design, pointing to uneven handling of grounded reasoning tasks.

Few-shot prompting in the generation setting provides inconsistent and often limited benefits. While a few models improve slightly with more examples, others plateau or even regress—likely due to prompt misalignment, increased input complexity, or difficulty integrating abstract reasoning from support instances. Interestingly, similar example selection yields minimal advantage over random sampling, indicating that contextual proximity alone is insufficient for improving commonsense grounding. Moreover, adding CoT rarely helps and sometimes reduces performance, as seen in models like InternVL2.5-4B and LLaVA-NeXT-Interleave-Qwen-0.5B, likely due to verbose or unstructured reasoning disrupting answer accuracy.

A striking observation emerges from the pairwise accuracy results: the Idefics family—particularly Idefics2-8B—demonstrates significant few-shot gains, reaching over 90\% accuracy, despite showing minimal improvement in the generative setting. This divergence suggests that Idefics models may internalize few-shot cues more effectively when used to score options rather than generate answers directly. Their architecture or training approach may better align with likelihood-based ranking under in-context examples, rather than free-form reasoning. These findings point to a broader insight: commonsense reasoning benefits not just from stronger models, but also from task-specific prompting formats and inference strategies that better exploit the models’ internal representations.

\subsection{Temporal Reasoning}
\label{appd_temporal_r}
The Temporal Reasoning task assesses a model’s ability to understand temporal order and progression based on visual sequences. Each instance presents a concatenated series of images arranged in a single composite image, depicting different stages of an event or process. The model must answer a multiple-choice question about temporal relationships such as "what happened before" or "what happens next." This task probes the model’s ability to integrate sequential visual information and infer temporal dynamics from static representations.
\subsubsection{Example Selection}
\begin{figure*}[ht!]
    \centering
    \begin{subfigure}[b]{0.3\linewidth}
        \centering
        \includegraphics[width=\textwidth]{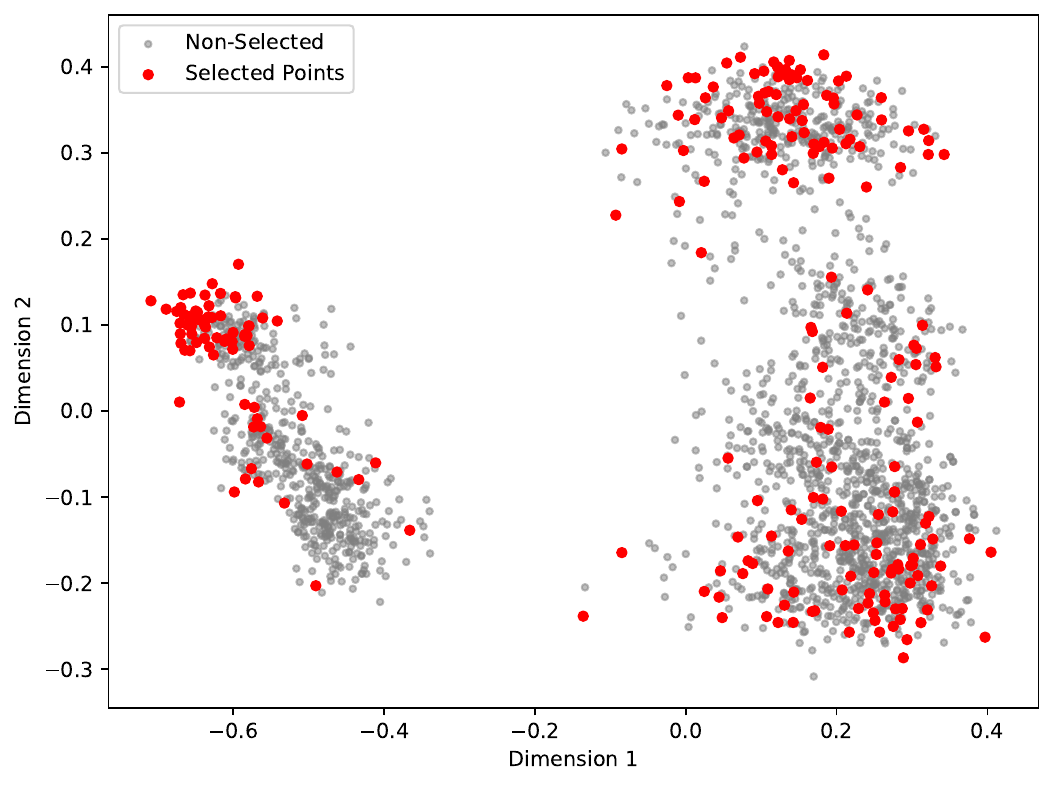}
        \caption{PCA Projection}
        
    \end{subfigure}
    \begin{subfigure}[b]{0.3\linewidth}
        \centering
        \includegraphics[width=\textwidth]{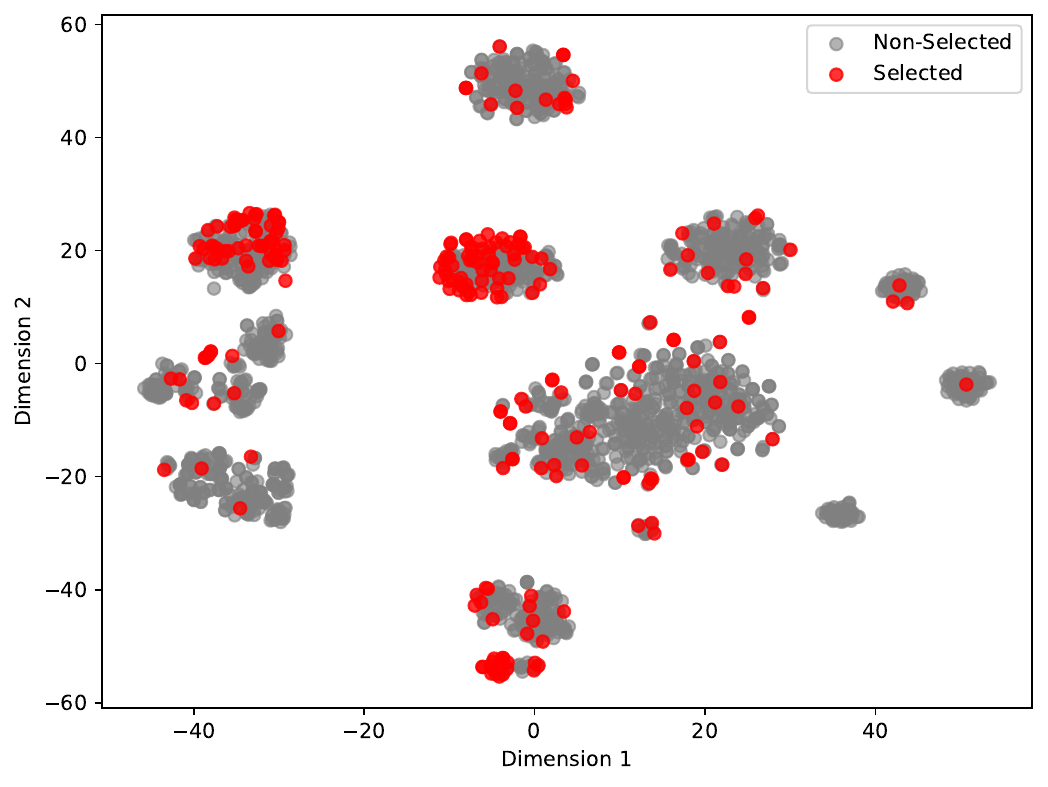}
        \caption{t-SNE Visualization}
        
    \end{subfigure}
    \begin{subfigure}[b]{0.3\linewidth}
        \centering
        \includegraphics[width=\textwidth]{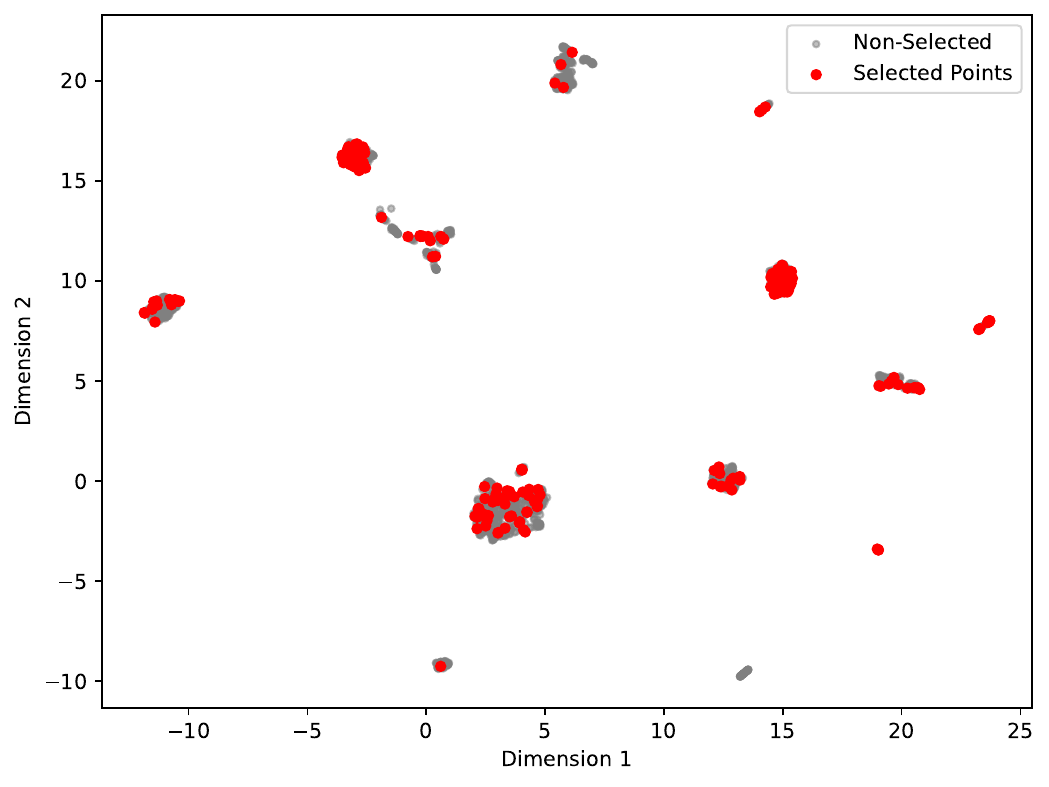}
        \caption{UMAP Visualization}
        
    \end{subfigure}

    \caption{Visualization of PCA, t-SNE, and UMAP projections for the Temporal Reasoning task. }
    \label{fig:sixfigsTR}
\end{figure*}

Figure \ref{fig:sixfigsTR} presents PCA, t-SNE, and UMAP projections for the Temporal Reasoning task. The Graph-Cut $\lambda$ was set to $50$.

\subsubsection{In-Depth Results}
Temporal Reasoning in \benchmark emerges as one of the hardest tasks for current MLLMs. Most models—including those with instruction tuning or larger capacities—fail to achieve robust performance, with generative accuracy generally hovering around or below 60\%. Even strong performers like Qwen2.5-VL-7B-Instruct and InternVL2.5-8B struggle to maintain consistency across prompting conditions. Models such as Phi-3.5-Vision-Instruct and Qwen2-VL-2B show minimal improvement over baseline scores, highlighting a broader limitation in temporal comprehension that persists across scales and model families.

Few-shot prompting provides limited or negative impact in generative settings. Additional examples, whether randomly sampled or semantically similar, do not consistently improve performance and often introduce prompt interference or distract from temporal focus. CoT fails to alleviate this challenge and frequently reduces accuracy, particularly in mid- and small-sized models. The instability of these prompting strategies suggests that temporal reasoning requires more than in-context pattern induction—it likely demands explicit modeling of event order, duration, and causality, which remains weakly represented in current MLLM architectures.

Interestingly, the Idefics family shows a sharp contrast between generation and pairwise settings. While their generative performance remains moderate, Idefics models achieve the highest pairwise accuracies across all prompting conditions (e.g., Idefics2-8B exceeds 95\% in 8-shot similar), indicating strong internal ranking capabilities despite weak output generation. This points to an architectural or decoding advantage in option-based inference, but the broader finding holds: Temporal Reasoning remains unsolved for most models, underscoring a critical gap in temporal understanding that is not easily bridged by scale, prompting, or tuning alone.

\subsection{Coreference Resolution}
\label{appd_coreference_r}
The Coreference Resolution task tests the model’s ability to resolve pronouns or noun phrases to the correct visual referents. Each question requires grounding linguistic expressions in specific entities within the image, often under ambiguity. This task evaluates multimodal understanding of referential language, a key challenge for aligning visual and textual modalities.
\subsubsection{Example Selection}
\begin{figure*}[ht!]
    \centering
    \begin{subfigure}[b]{0.3\linewidth}
        \centering
        \includegraphics[width=\textwidth]{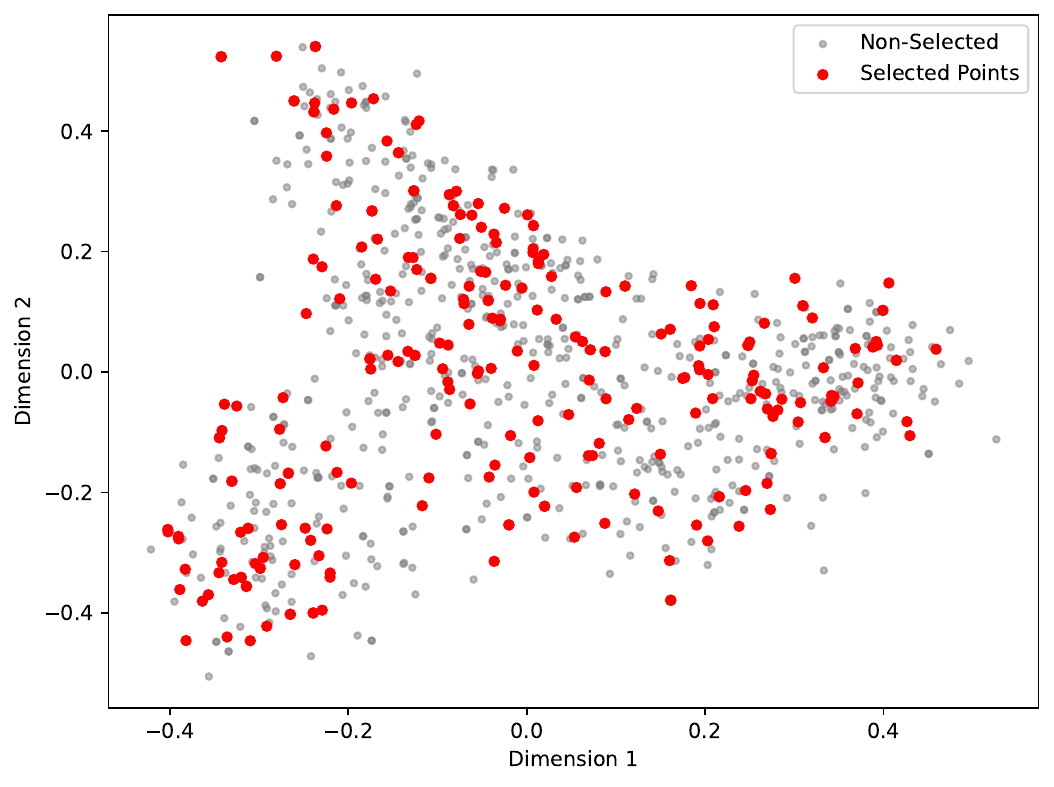}
        \caption{PCA Projection}
        
    \end{subfigure}
    \begin{subfigure}[b]{0.3\linewidth}
        \centering
        \includegraphics[width=\textwidth]{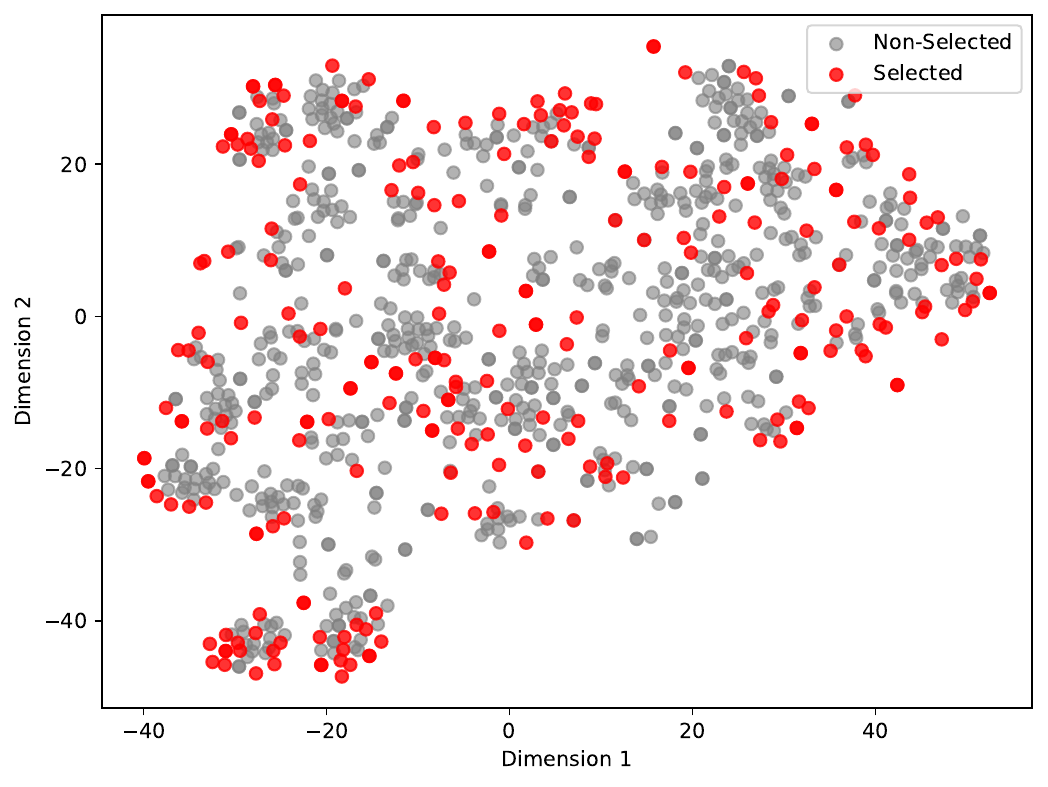}
        \caption{t-SNE Visualization}
        
    \end{subfigure}
    \begin{subfigure}[b]{0.3\linewidth}
        \centering
        \includegraphics[width=\textwidth]{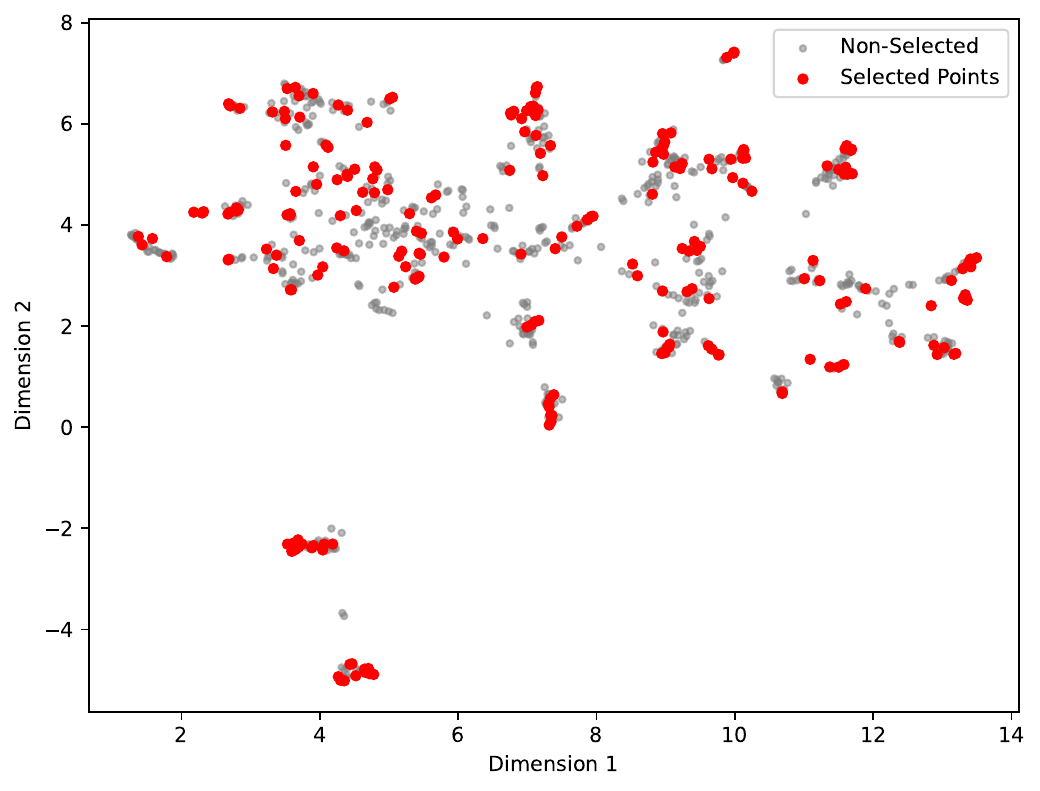}
        \caption{UMAP Visualization}
        
    \end{subfigure}

    \caption{Visualization of PCA, t-SNE, and UMAP projections for the Coreference Resolution task. }
    \label{fig:sixfigsCorR}
\end{figure*}

Figure \ref{fig:sixfigsCorR} presents PCA, t-SNE, and UMAP projections for the Coreference Resolution task. The Graph-Cut $\lambda$ was set to $50$.

\subsubsection{In-Depth Results}
Coreference Resolution appears to be a relatively easier task within the \benchmark suite, as evidenced by the strong performance of several advanced models. Instruction-tuned and larger models such as Qwen2.5-VL-7B-Instruct, InternVL2.5-8B-MPO, and the Idefics family consistently achieve high accuracy, often surpassing 90\% in both generative and pairwise evaluations. This indicates that many models have effectively internalized coreference patterns, allowing them to resolve pronouns and noun phrases with high reliability across varied prompting conditions.

Few-shot prompting generally benefits model performance, with random and similar example selections both contributing positively. However, the advantage of example similarity is less pronounced here than in other reasoning tasks, suggesting that coreference resolution relies more on robust language understanding than on contextual priming from closely related examples. Increasing the number of shots typically yields marginal gains, indicating diminishing returns beyond a small set of demonstrations. Interestingly, Chain of Thought prompting does not consistently enhance performance and may even slightly reduce accuracy for some models, potentially due to the relatively straightforward nature of coreference resolution where stepwise reasoning is less critical.

Overall, the high baseline accuracies and stable gains from few-shot prompting demonstrate that coreference resolution is a well-captured capability among recent multimodal language models. This contrasts with more challenging reasoning tasks where models struggle despite scaling and advanced prompting. The findings highlight that coreference resolution benefits substantially from both model capacity and instruction tuning, and the task’s linguistic regularity allows models to generalize well without heavy dependence on example similarity or complex prompting strategies.
\section{How CoT Fails?}
\label{appd_why_cot_fails}
\begin{figure*}[t!]
    \centering
    \includegraphics[width=6.2in]{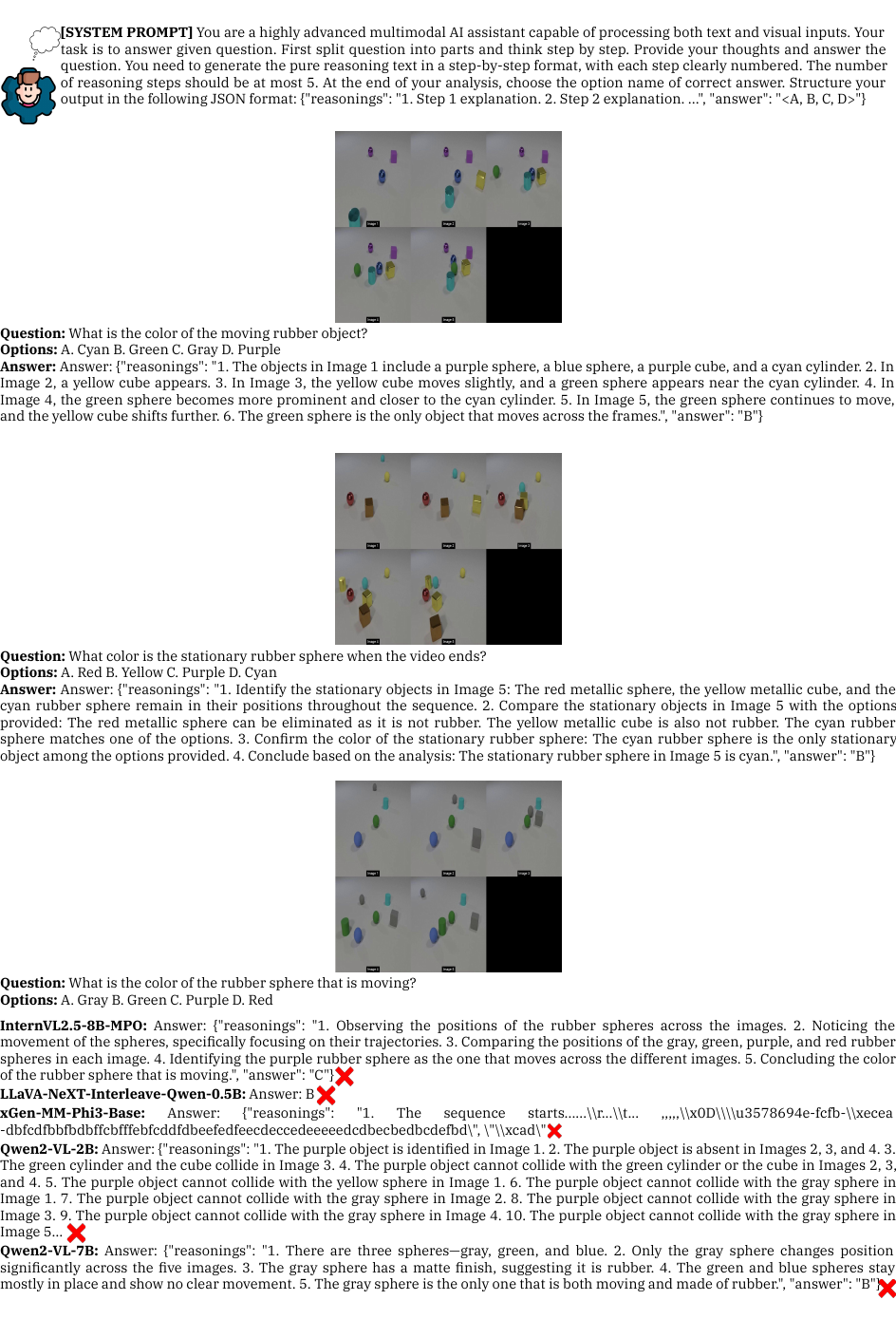}
    \captionsetup{font=footnotesize} %
    \caption{\textbf{Failure cases of CoT usage in MLLMs.} Shown are common CoT failure types: (i) hallucinated reasoning inconsistent with the image, (ii) missing CoT despite prompting, (iii) irrelevant or malformed outputs, (iv) infinite reasoning loops, and (v) correct reasoning but incorrect final answers, possibly due to interference from prior examples in few-shot settings. These issues reveal the fragility of current CoT methods in complex multimodal settings.}
    \label{fig:how_cot_fails}
\end{figure*}

Figure~\ref{fig:how_cot_fails} illustrates critical failure modes in multimodal models when applying CoT prompting to tasks from \benchmark. Despite explicit instructions to reason step by step and produce structured outputs, several models fail to comply. The examples capture common breakdowns, including hallucinated visual reasoning, missing CoT despite prompting, malformed or irrelevant generations, infinite reasoning loops that exceed token limits, and instances where models generate correct reasoning steps but produce incorrect final answers—often influenced by prior few-shot examples.

A closer inspection reveals diverse error types. Some models fabricate visual details absent in the input, such as hallucinating motion or misidentifying object colors. Others omit reasoning altogether, jump to an unsupported answer, or fail to produce an answer. Severe cases exhibit malformed tokens or excessively verbose, incoherent reasoning that terminates prematurely. Notably, even when reasoning is logically sound and grounded, the final prediction may still be incorrect—suggesting interference from contextual bias or in-context priming.

These errors highlight multiple root causes. Many MLLMs lack sufficient instruction tuning for structured, multimodal reasoning. Their pretraining objectives rarely prioritize alignment between visual inputs and textual logic. Moreover, models often lack temporal grounding capabilities essential for tasks involving movement or frame-wise comparison. In absence of such capabilities, they tend to fall back on linguistic priors or pattern matching rather than genuine visual reasoning.

To address these limitations, future multimodal systems should incorporate fine-grained CoT supervision grounded in visual evidence, with intermediate feedback during the reasoning process. Techniques such as constrained decoding, explicit visual-text alignment, and iterative reasoning modules can reduce hallucinations and enforce coherence. Additionally, mitigating few-shot interference may require better context selection or retrieval-aware prompting strategies.

\benchmark is crucial in surfacing these limitations. It challenges models with compositional, temporally dependent, and instruction-sensitive tasks under few-shot settings, offering a diagnostic lens not only into accuracy but reasoning fidelity. As Figure~\ref{fig:how_cot_fails} demonstrates, strong QA performance alone does not ensure coherent, evidence-based reasoning. This underscores the brittleness of current CoT approaches and the importance of benchmarks like \benchmark in guiding the development of reliable, explainable, and instruction-following multimodal systems.

\section{Qualitative Examples}
\label{appd_qualitative_examples}
\begin{figure*}[!t]
\centering
    \includegraphics[page=11, width=\linewidth]{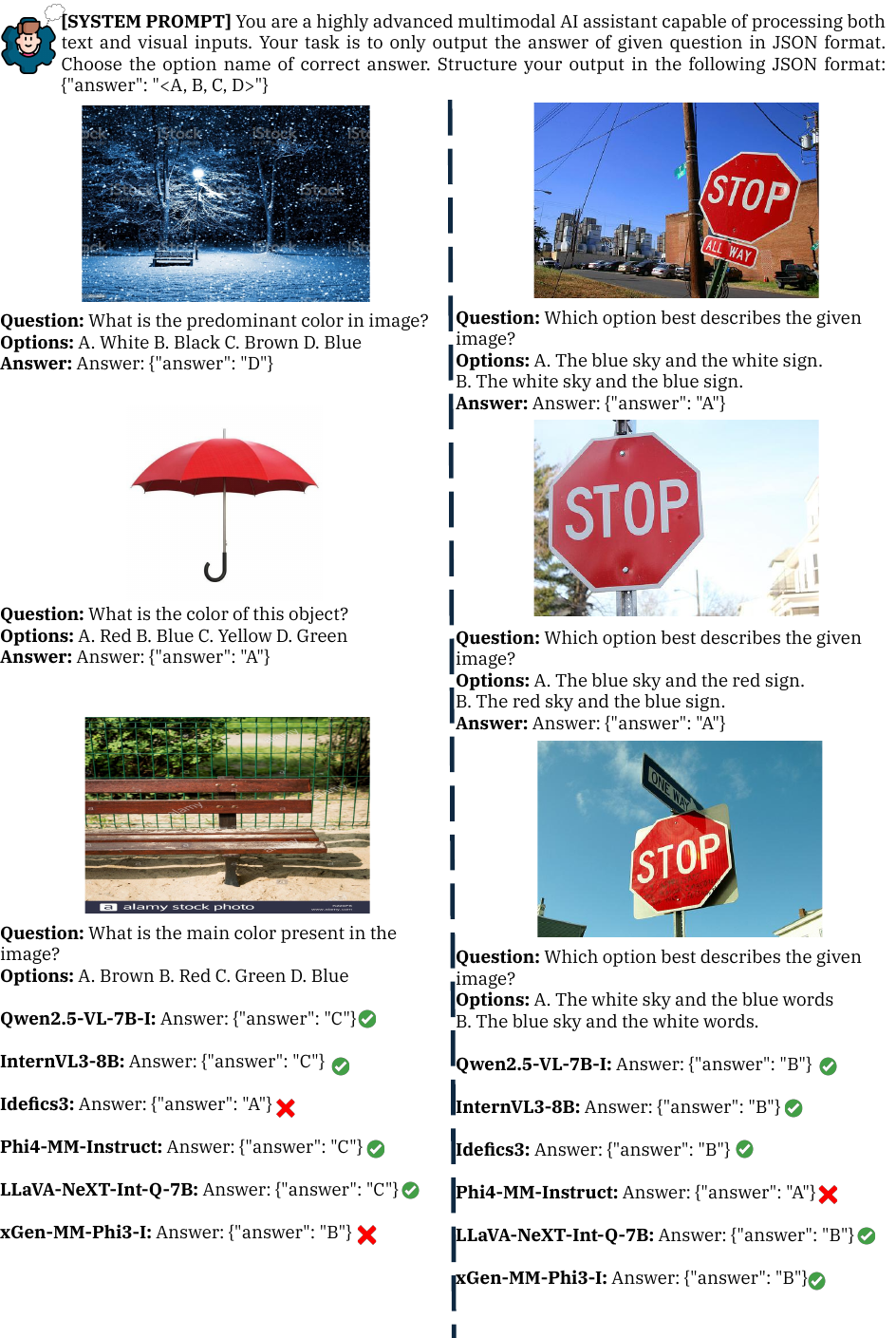}
    \caption{Example model predictions on instances from the \emph{Attribute Recognition} task, where demonstrations are selected based on visual and textual similarity.}
    \label{fig:icl-AttributeRecognition-detailed-examples}
\end{figure*}

\begin{figure*}[!t]
\centering
    \includegraphics[page=12, width=\linewidth]{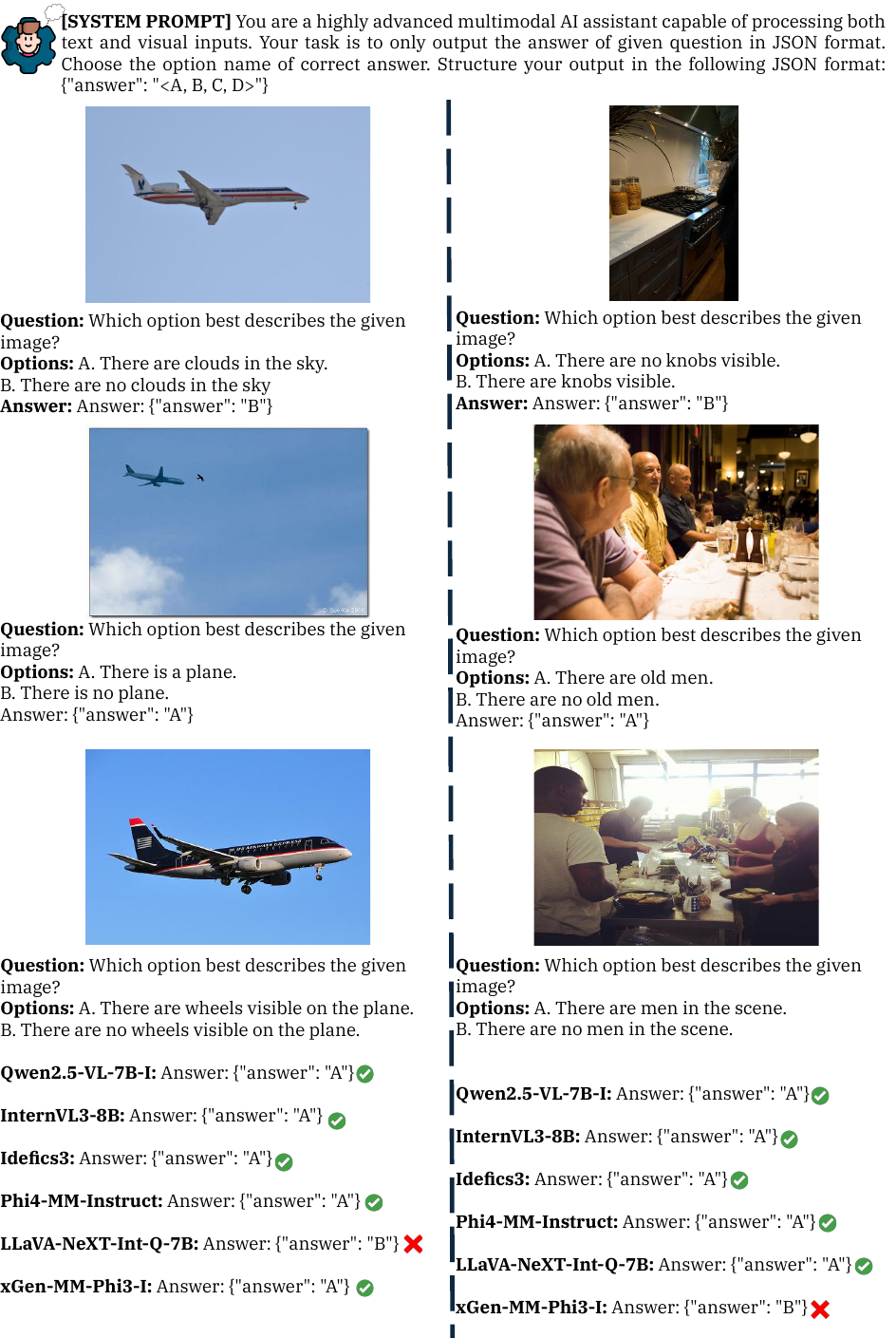}
    \caption{Example model predictions on instances from the \emph{Visual Object Recognition} task, where demonstrations are selected based on visual and textual similarity.}
    \label{fig:icl-VisualObjectRecognition-detailed-examples}
\end{figure*}

\begin{figure*}[!t]
\centering
    \includegraphics[page=13, width=\linewidth]{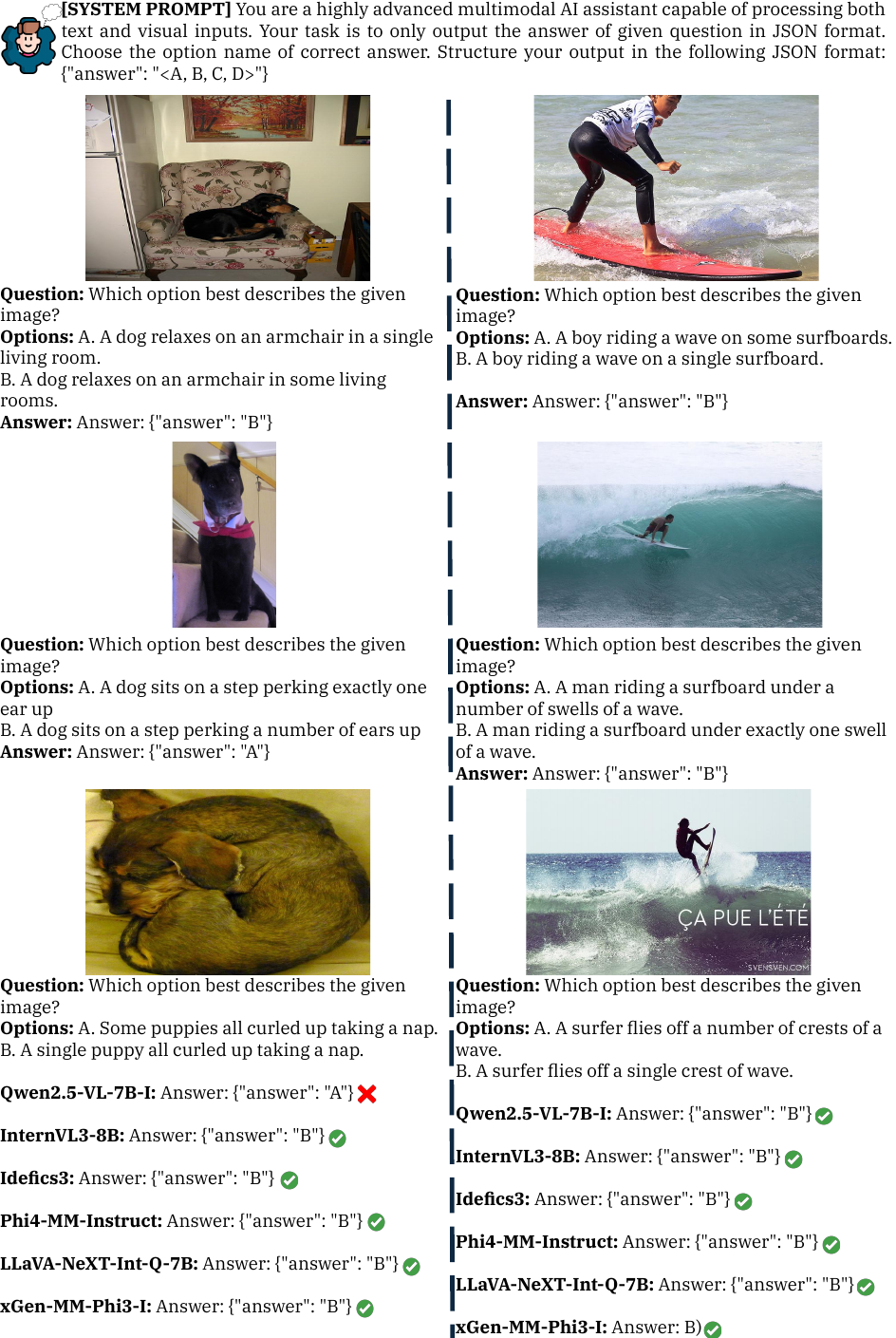}
    \caption{Example model predictions on instances from the \emph{Plurality Recognition} task, where demonstrations are selected based on visual and textual similarity.}
    \label{fig:icl-PluralityRecognition-detailed-examples}
\end{figure*}

\begin{figure*}[!t]
\centering
    \includegraphics[page=14, width=\linewidth]{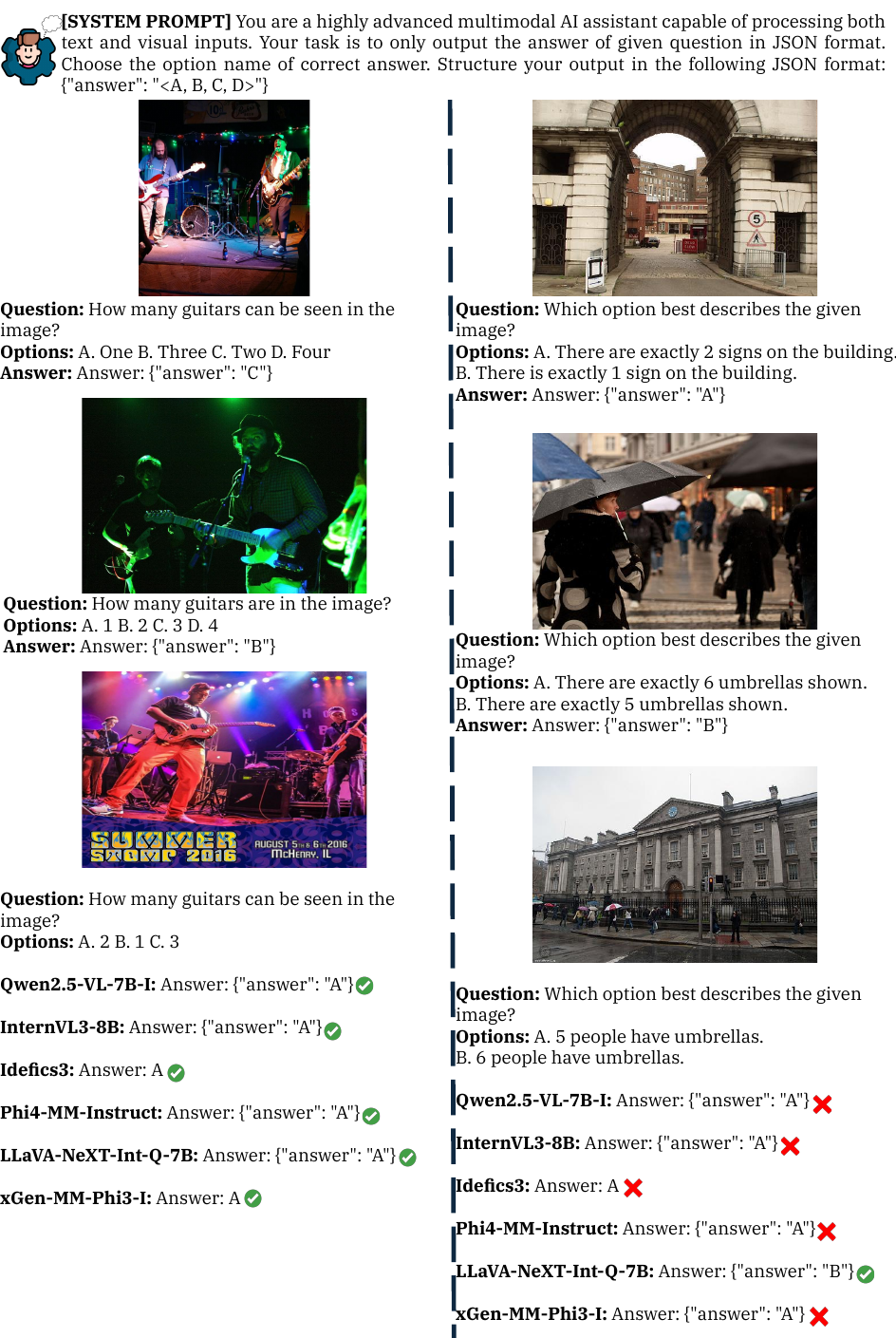}
    \caption{Example model predictions on instances from the \emph{Object Counting} task, where demonstrations are selected based on visual and textual similarity.}
    \label{fig:icl-ObjectCounting-detailed-examples}
\end{figure*}

\begin{figure*}[!t]
\centering
    \includegraphics[page=15, width=\linewidth]{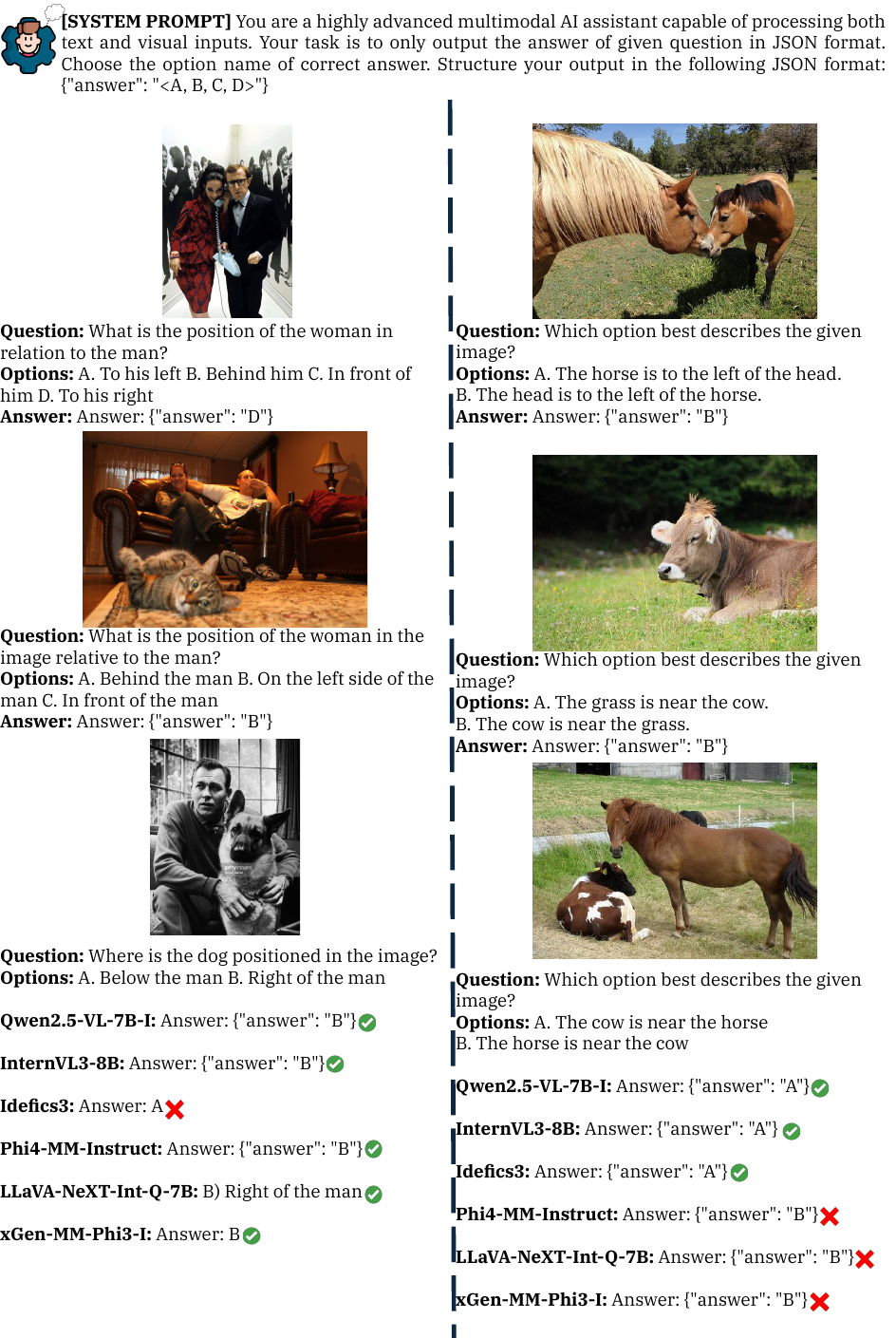}
    \caption{Example model predictions on instances from the \emph{Spatial Relations Understanding} task, where demonstrations are selected based on visual and textual similarity.}
    \label{fig:icl-SpatialRelationsUnderstanding-detailed-examples}
\end{figure*}

\begin{figure*}[!t]
\centering
    \includegraphics[page=16, width=\linewidth]{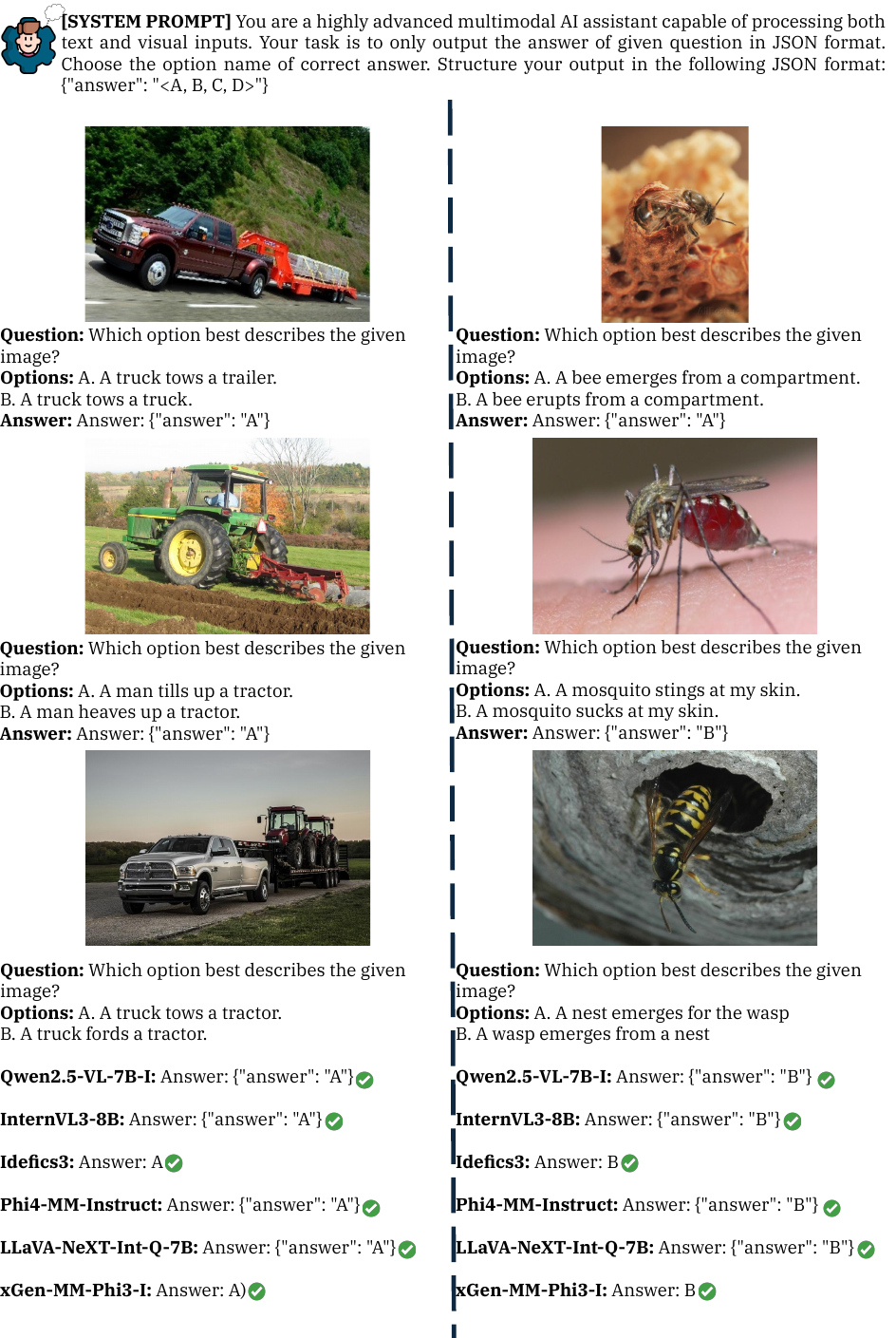}
    \caption{Example model predictions on instances from the \emph{Action Recognition} task, where demonstrations are selected based on visual and textual similarity.}
    \label{fig:icl-ActionRecognition-detailed-examples}
\end{figure*}

\begin{figure*}[!t]
\centering
    \includegraphics[page=17, width=\linewidth]{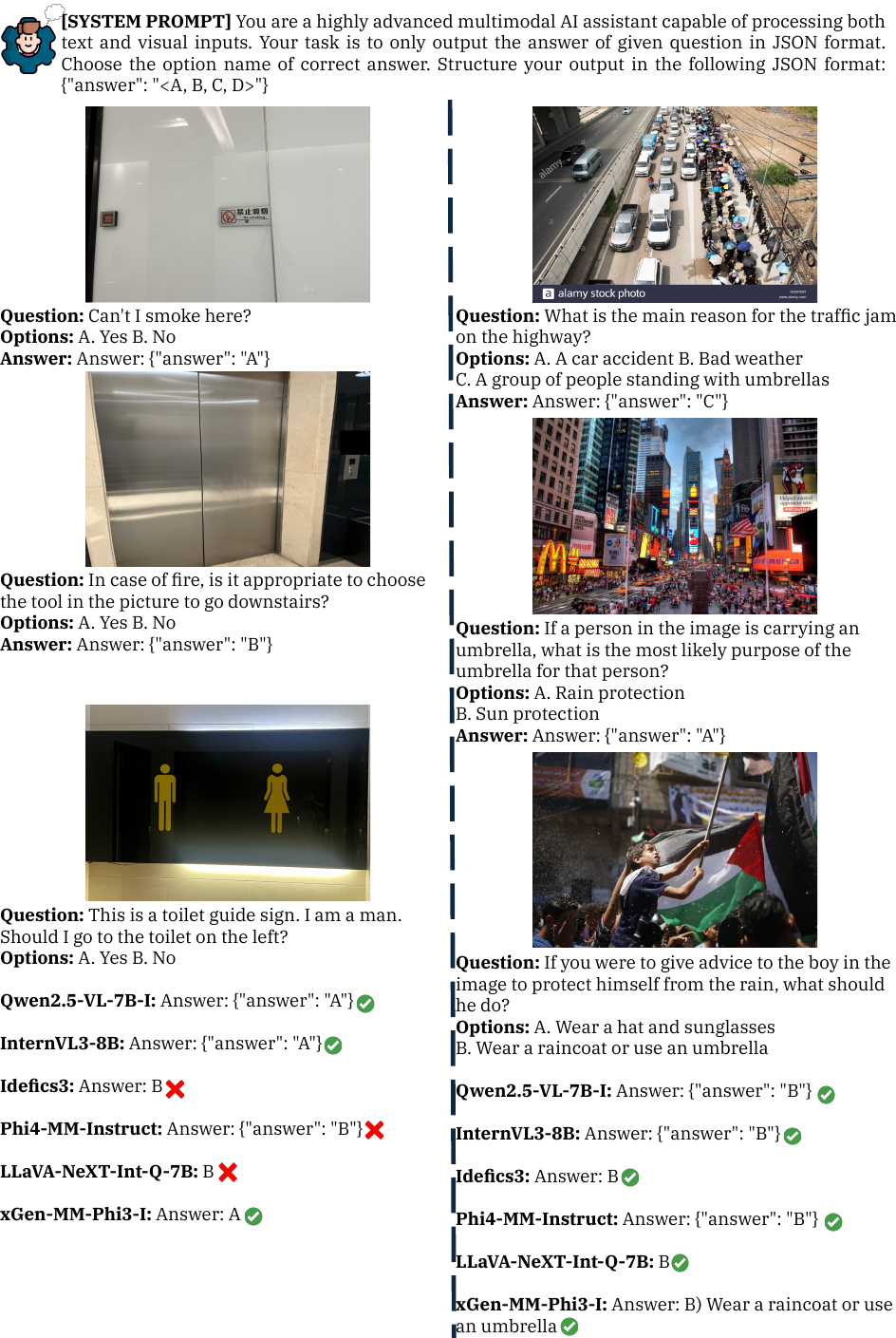}
    \caption{Example model predictions on instances from the \emph{Commonsense Reasoning} task, where demonstrations are selected based on visual and textual similarity.}
    \label{fig:icl-CommonsenseReasoning-detailed-examples}
\end{figure*}

\begin{figure*}[!t]
\centering
    \includegraphics[page=18, width=\linewidth]{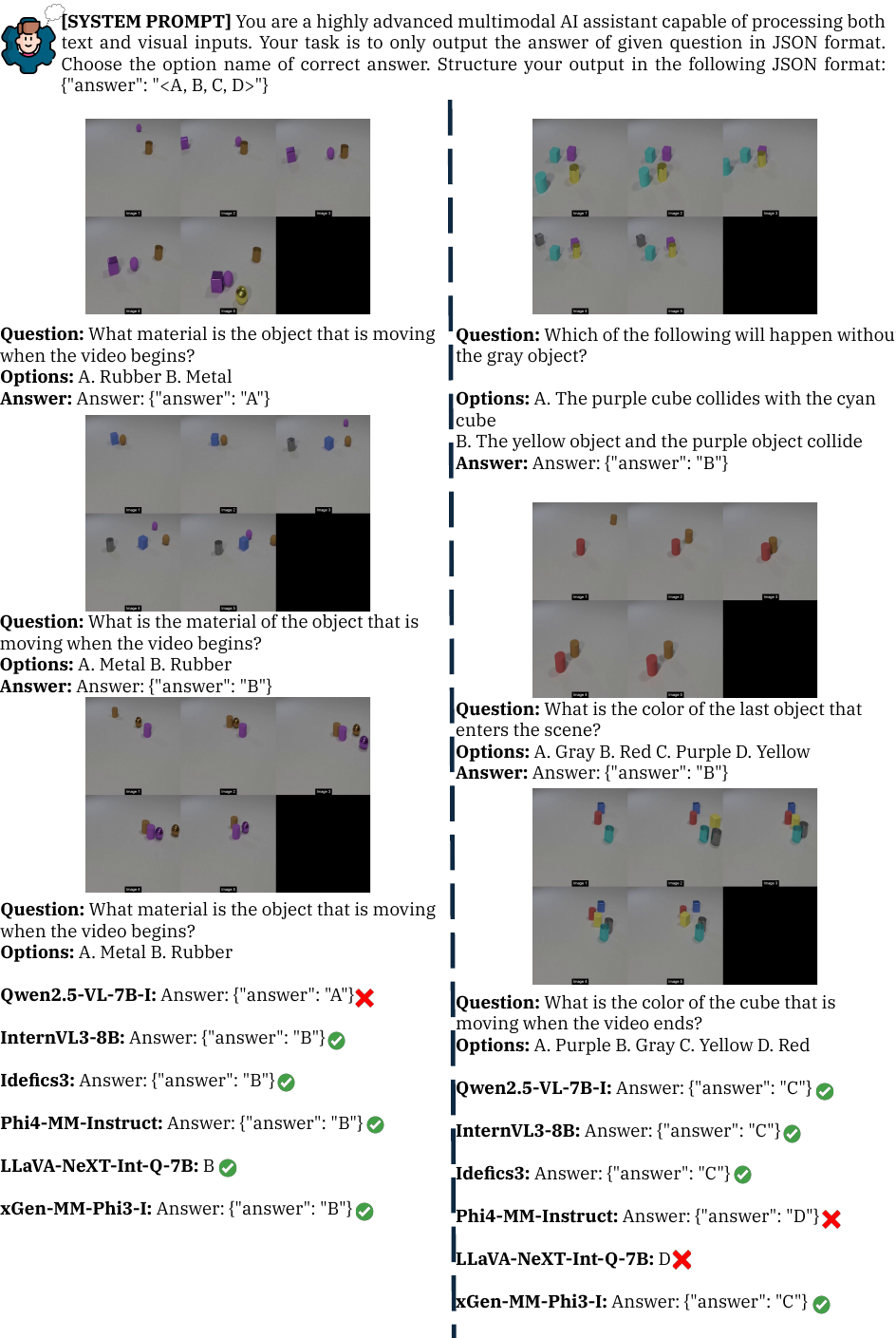}
    \caption{Example model predictions on instances from the \emph{Temporal Reasoning} task, where demonstrations are selected based on visual and textual similarity.}
    \label{fig:icl-TemporalReasoning-detailed-examples}
\end{figure*}

\begin{figure*}[!t]
\centering
    \includegraphics[page=19, width=\linewidth]{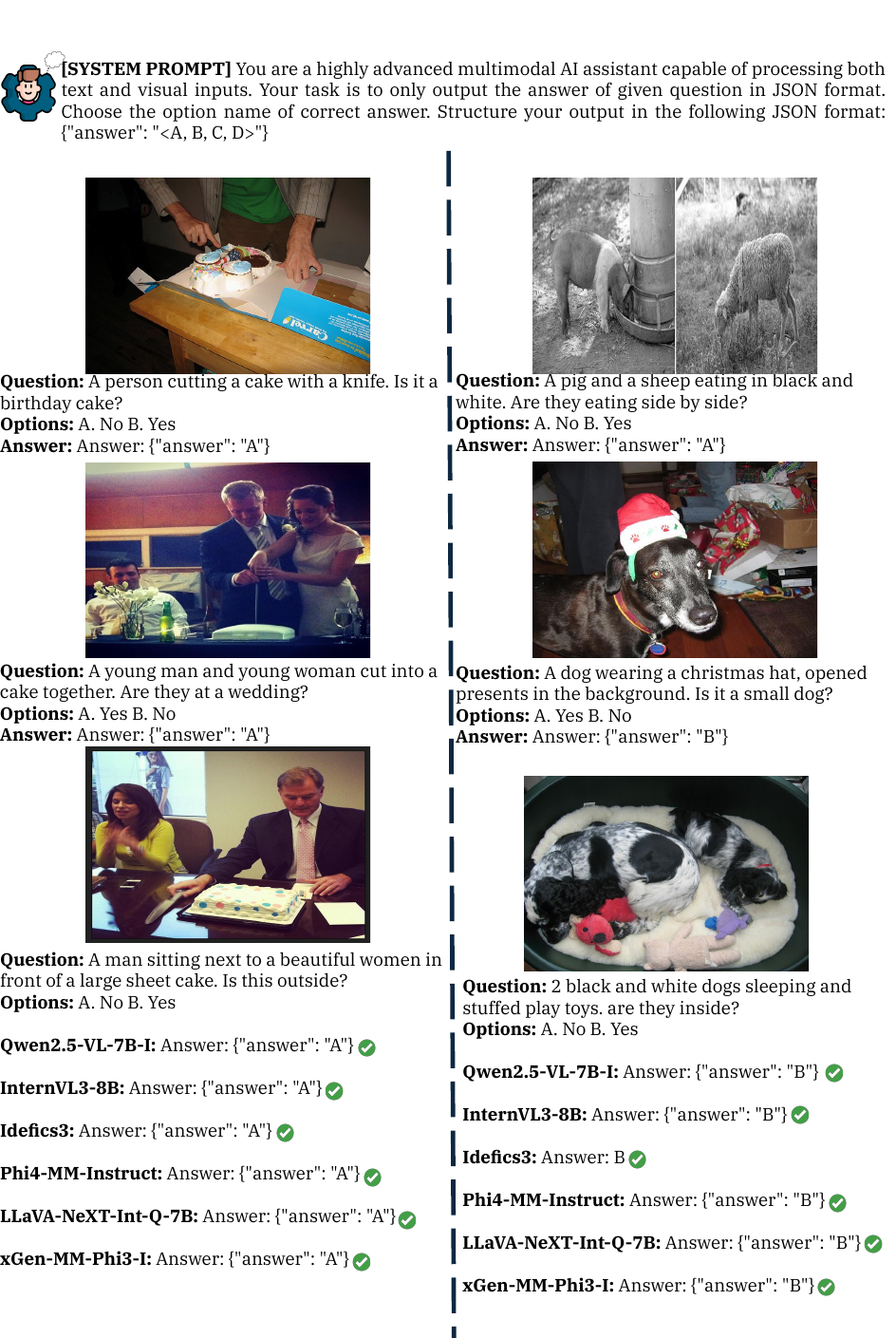}
    \caption{Example model predictions on instances from the \emph{Coreference Resolution} task, where demonstrations are selected based on visual and textual similarity.}
    \label{fig:icl-CoreferenceResolution-detailed-examples}
\end{figure*}

\begin{figure*}[!t]
\centering
    \includegraphics[page=2, width=\linewidth]{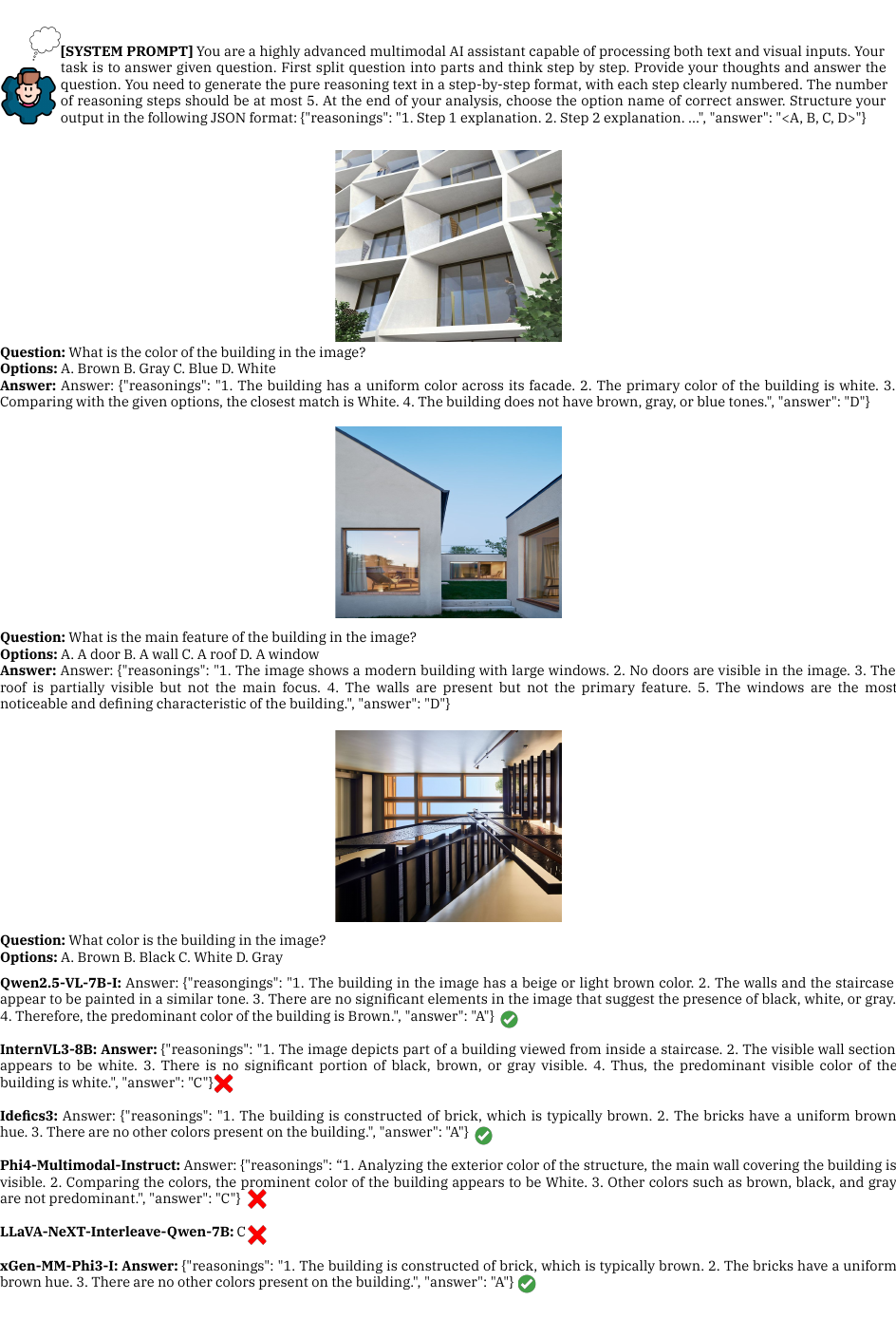}
    \caption{Example model predictions on instances from the \emph{Attribute Recognition} task, where demonstrations are selected based on visual and textual similarity, and Chain-of-Thought reasoning is employed.}
    \label{fig:cot-AttributeRecognition-detailed-examples}
\end{figure*}

\begin{figure*}[!t]
\centering
    \includegraphics[page=3, width=\linewidth]{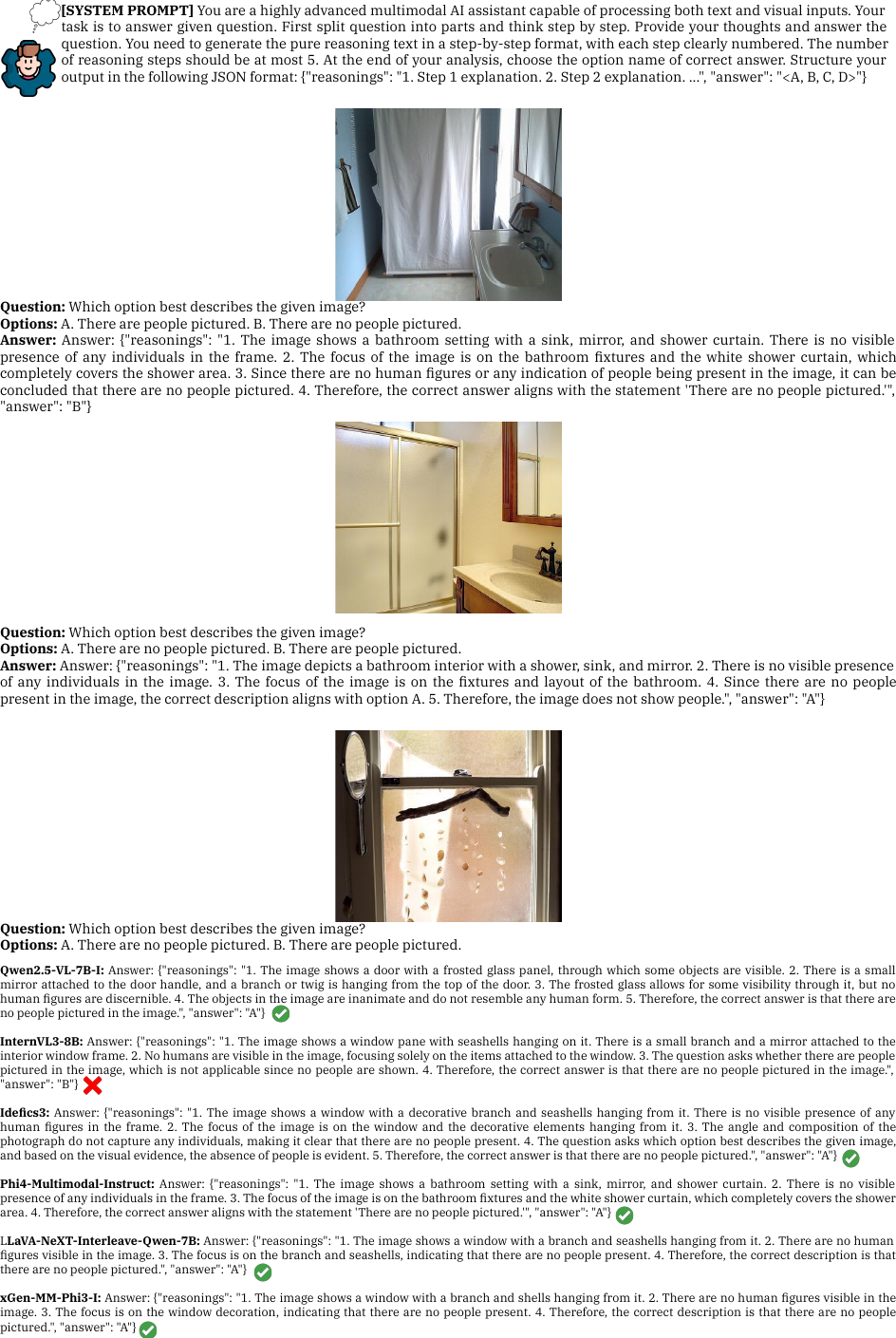}
    \caption{Example model predictions on instances from the \emph{Visual Object Recognition} task, where demonstrations are selected based on visual and textual similarity, and Chain-of-Thought reasoning is employed.}
    \label{fig:cot-VisualObjectRecognition-detailed-examples}
\end{figure*}

\begin{figure*}[!t]
\centering
    \includegraphics[page=4, width=\linewidth]{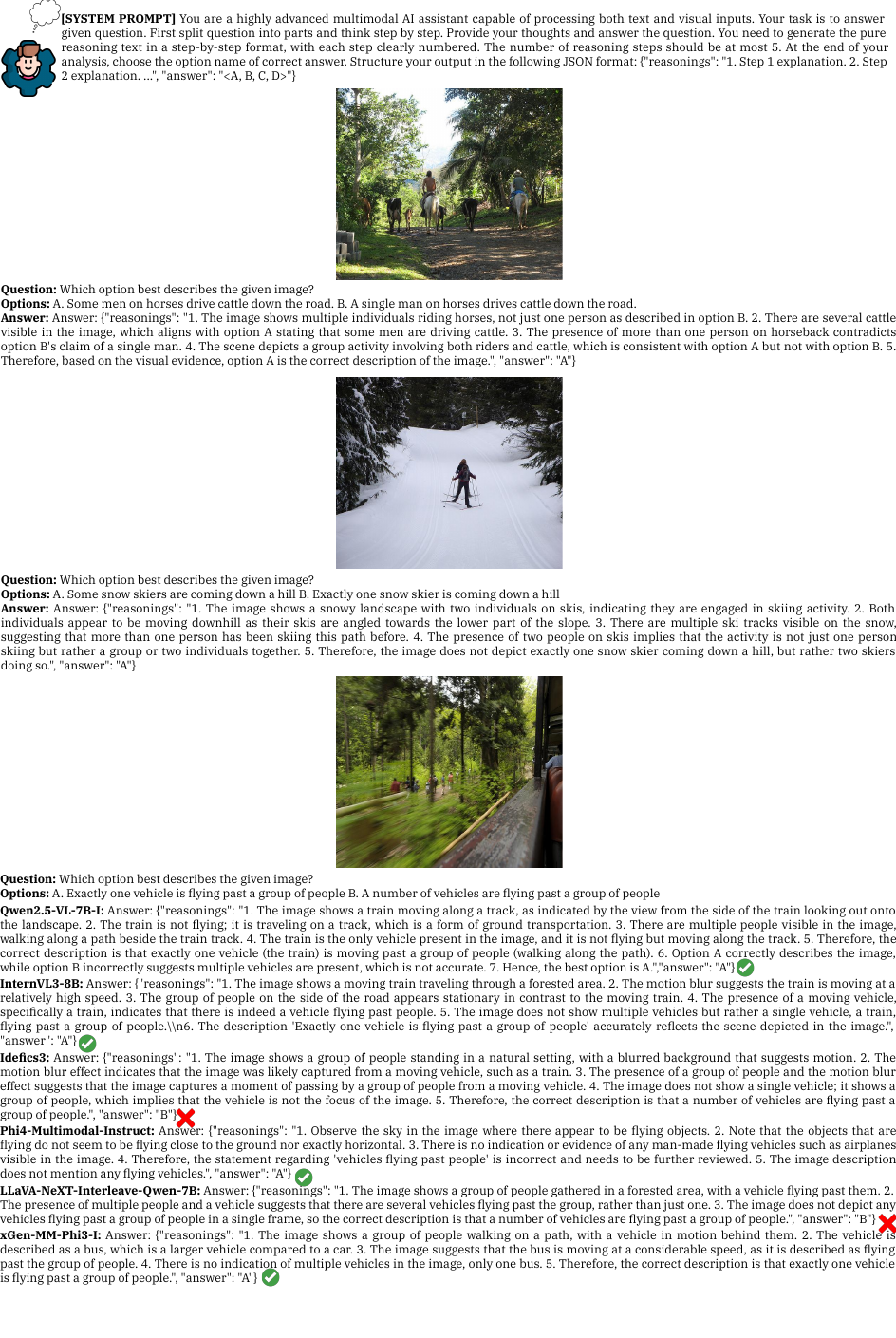}
    \caption{Example model predictions on instances from the \emph{Plurality Recognition} task, where demonstrations are selected based on visual and textual similarity, and Chain-of-Thought reasoning is employed.}
    \label{fig:cot-PluralityRecognition-detailed-examples}
\end{figure*}

\begin{figure*}[!t]
\centering
    \includegraphics[page=5, width=\linewidth]{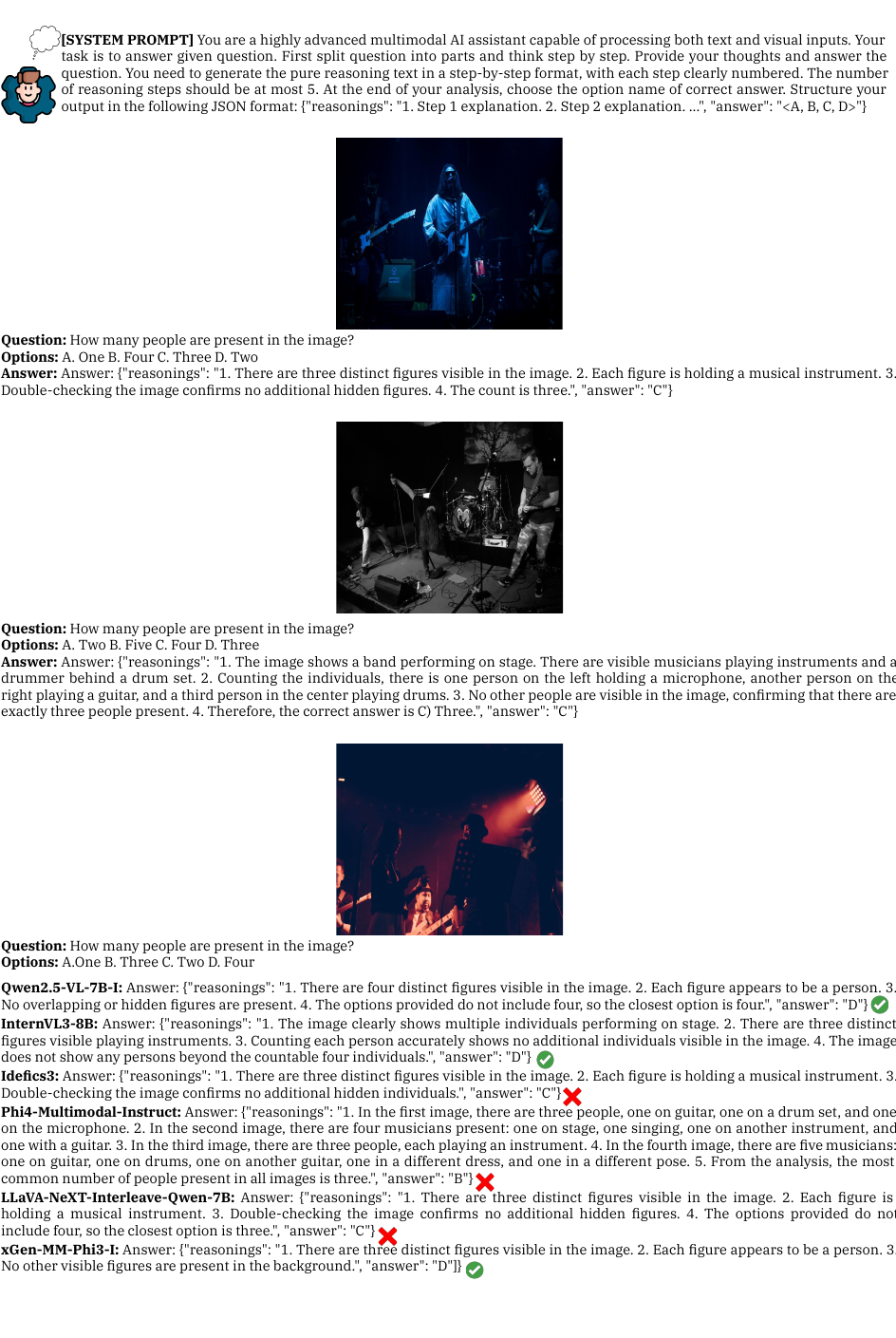}
    \caption{Example model predictions on instances from the \emph{Object Counting} task, where demonstrations are selected based on visual and textual similarity, and Chain-of-Thought reasoning is employed.}
    \label{fig:cot-ObjectCounting-detailed-examples}
\end{figure*}

\begin{figure*}[!t]
\centering
    \includegraphics[page=6, width=\linewidth]{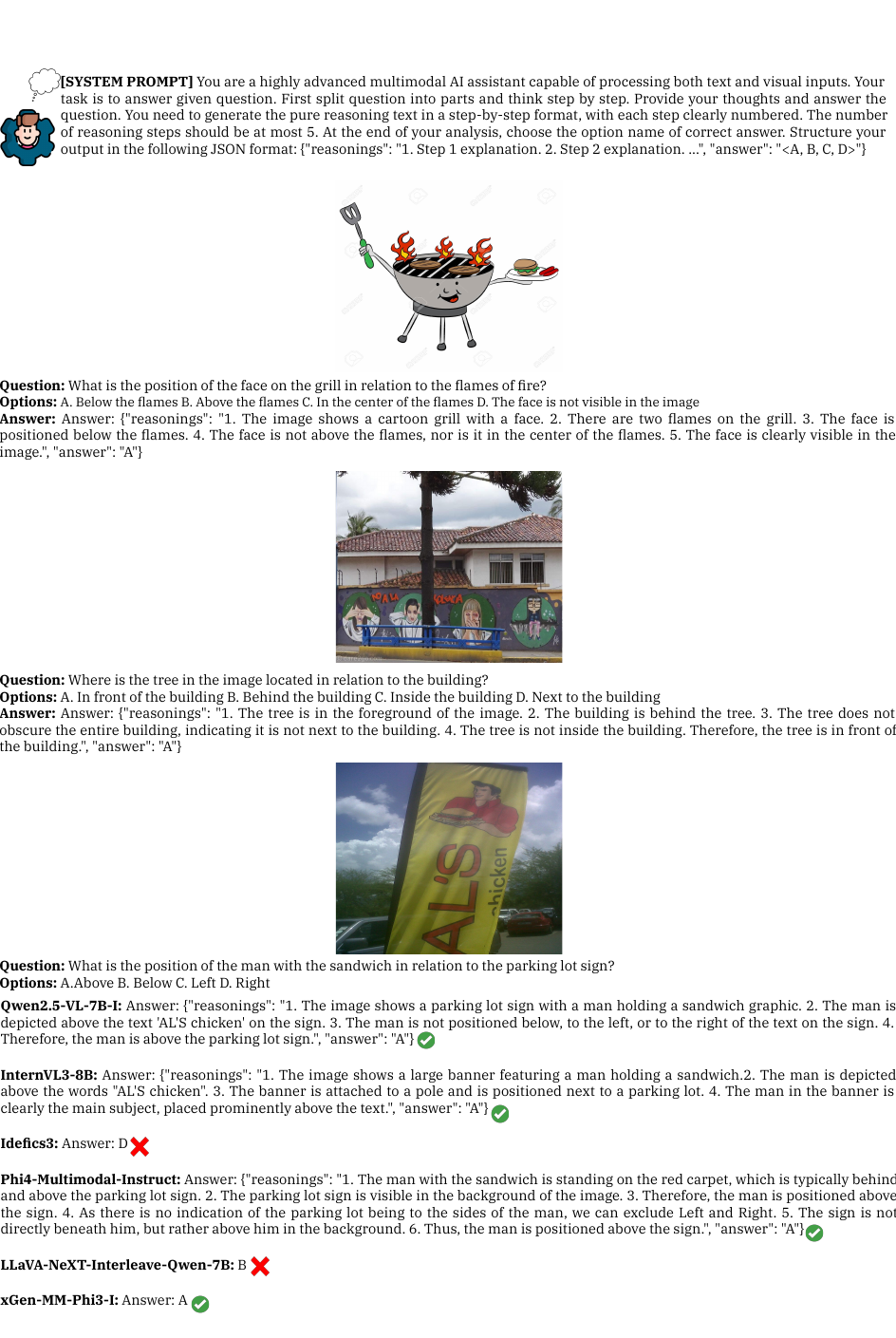}
    \caption{Example model predictions on instances from the \emph{Spatial Relations Understanding} task, where demonstrations are selected based on visual and textual similarity, and Chain-of-Thought reasoning is employed.}
    \label{fig:cot-SpatialRelationsUnderstanding-detailed-examples}
\end{figure*}

\begin{figure*}[!t]
\centering
    \includegraphics[page=7, width=\linewidth]{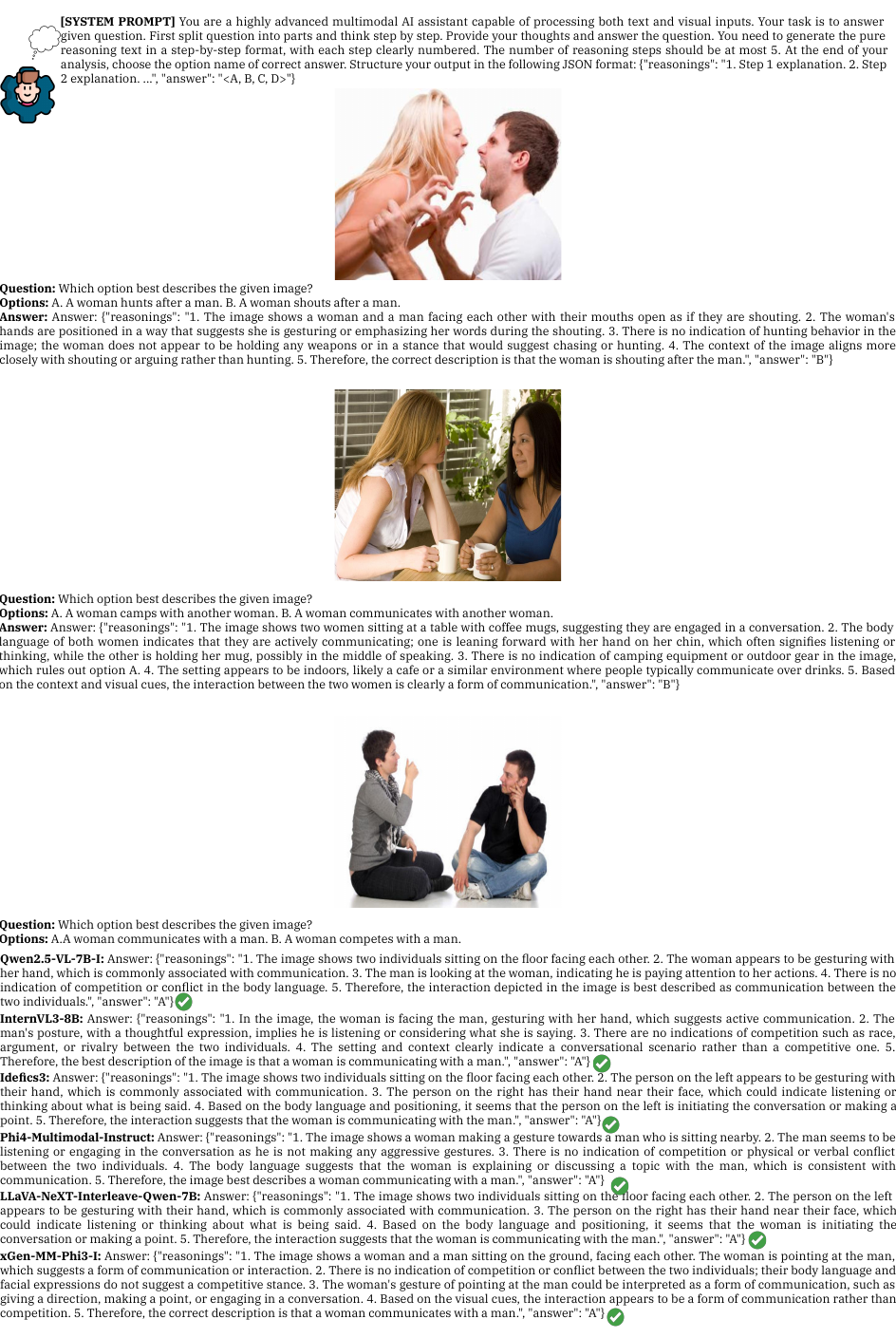}
    \caption{Example model predictions on instances from the \emph{Action Recognition} task, where demonstrations are selected based on visual and textual similarity, and Chain-of-Thought reasoning is employed.}
    \label{fig:cot-ActionRecognition-detailed-examples}
\end{figure*}

\begin{figure*}[!t]
\centering
    \includegraphics[page=8, width=\linewidth]{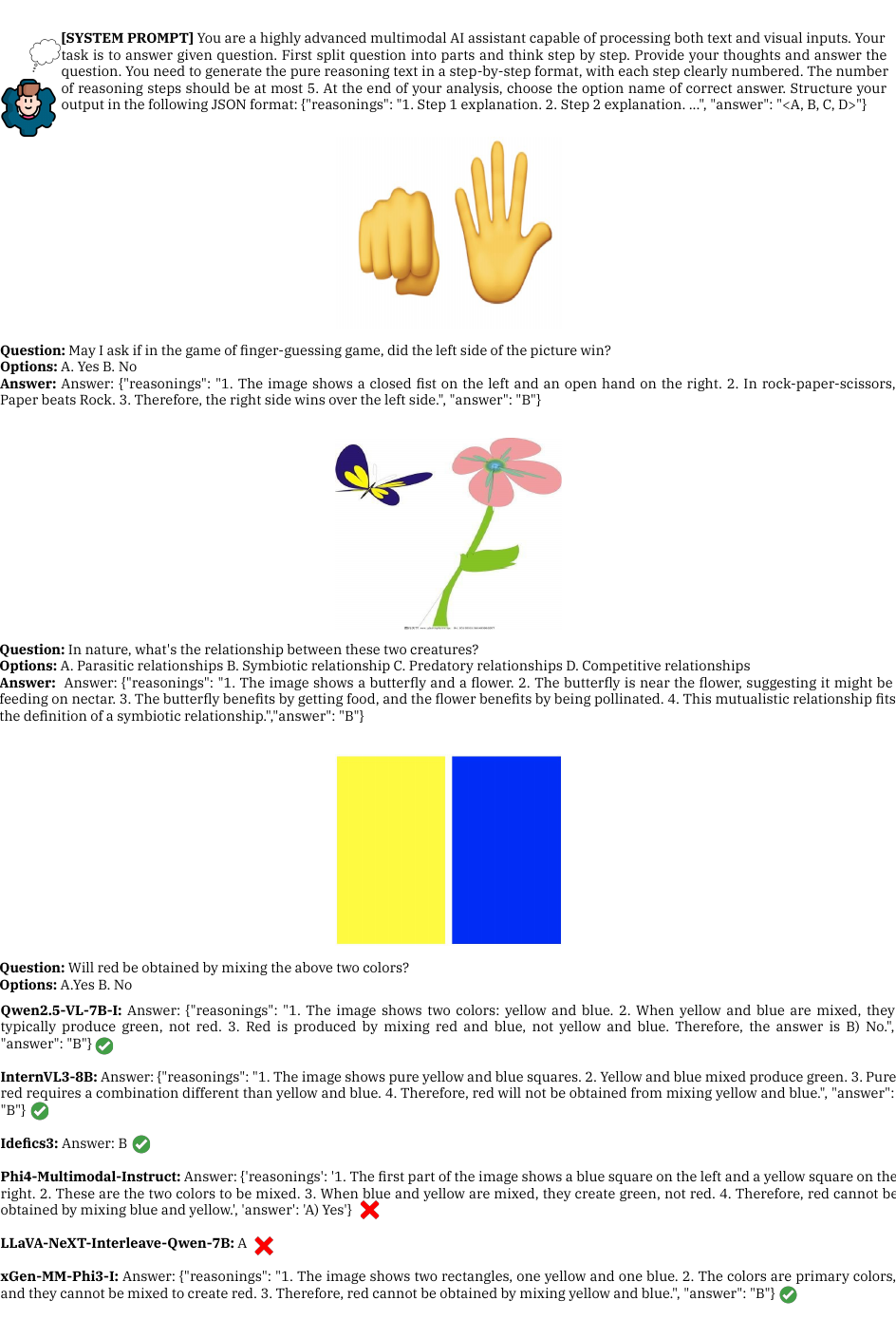}
    \caption{Example model predictions on instances from the \emph{Commonsense Reasoning} task, where demonstrations are selected based on visual and textual similarity, and Chain-of-Thought reasoning is employed.}
    \label{fig:cot-CommonsenseReasoning-detailed-examples}
\end{figure*}

\begin{figure*}[!t]
\centering
    \includegraphics[page=9, width=\linewidth]{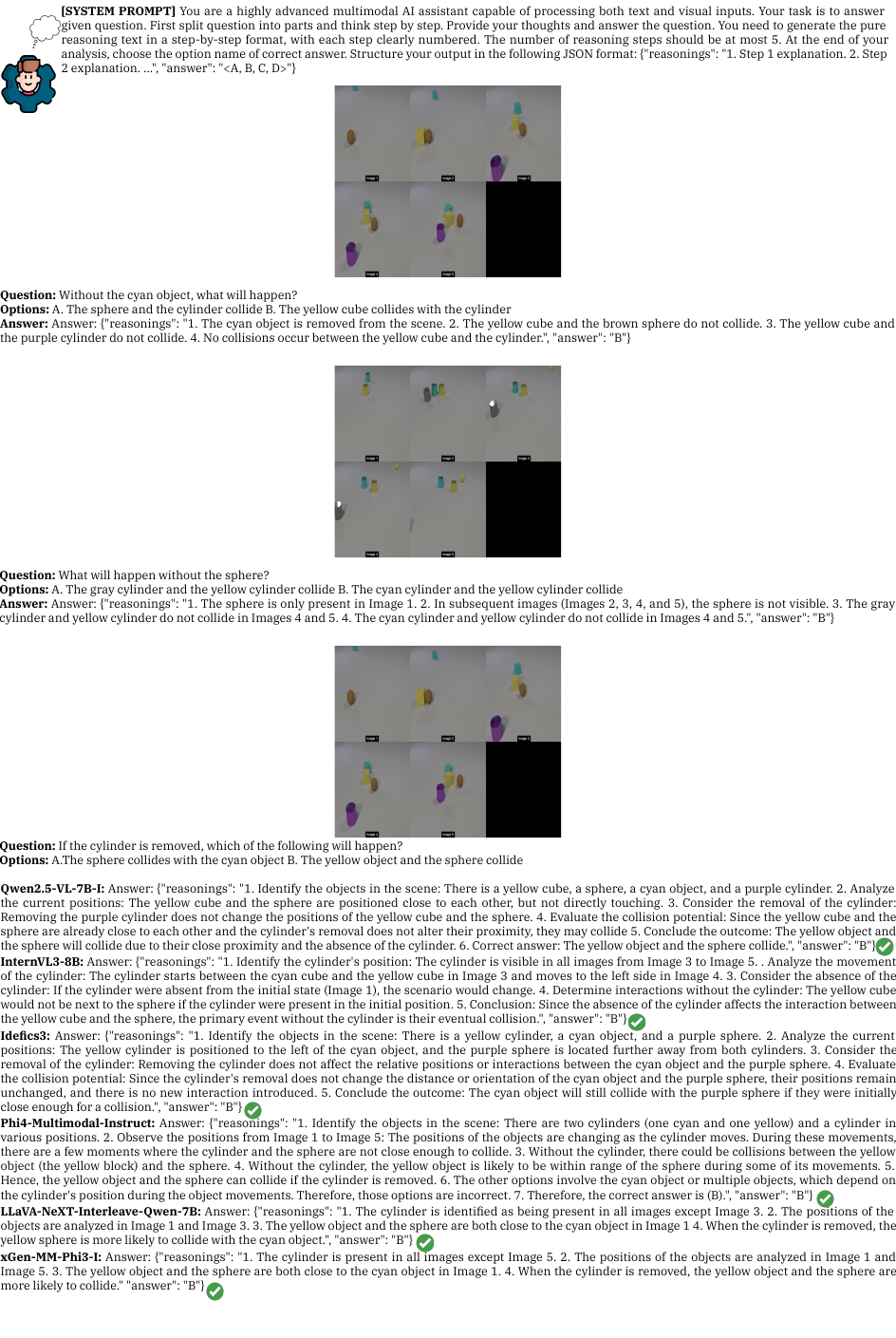}
    \caption{Example model predictions on instances from the \emph{Temporal Reasoning} task, where demonstrations are selected based on visual and textual similarity, and Chain-of-Thought reasoning is employed.}
    \label{fig:cot-TemporalReasoning-detailed-examples}
\end{figure*}

\begin{figure*}[!t]
\centering
    \includegraphics[page=10, width=\linewidth]{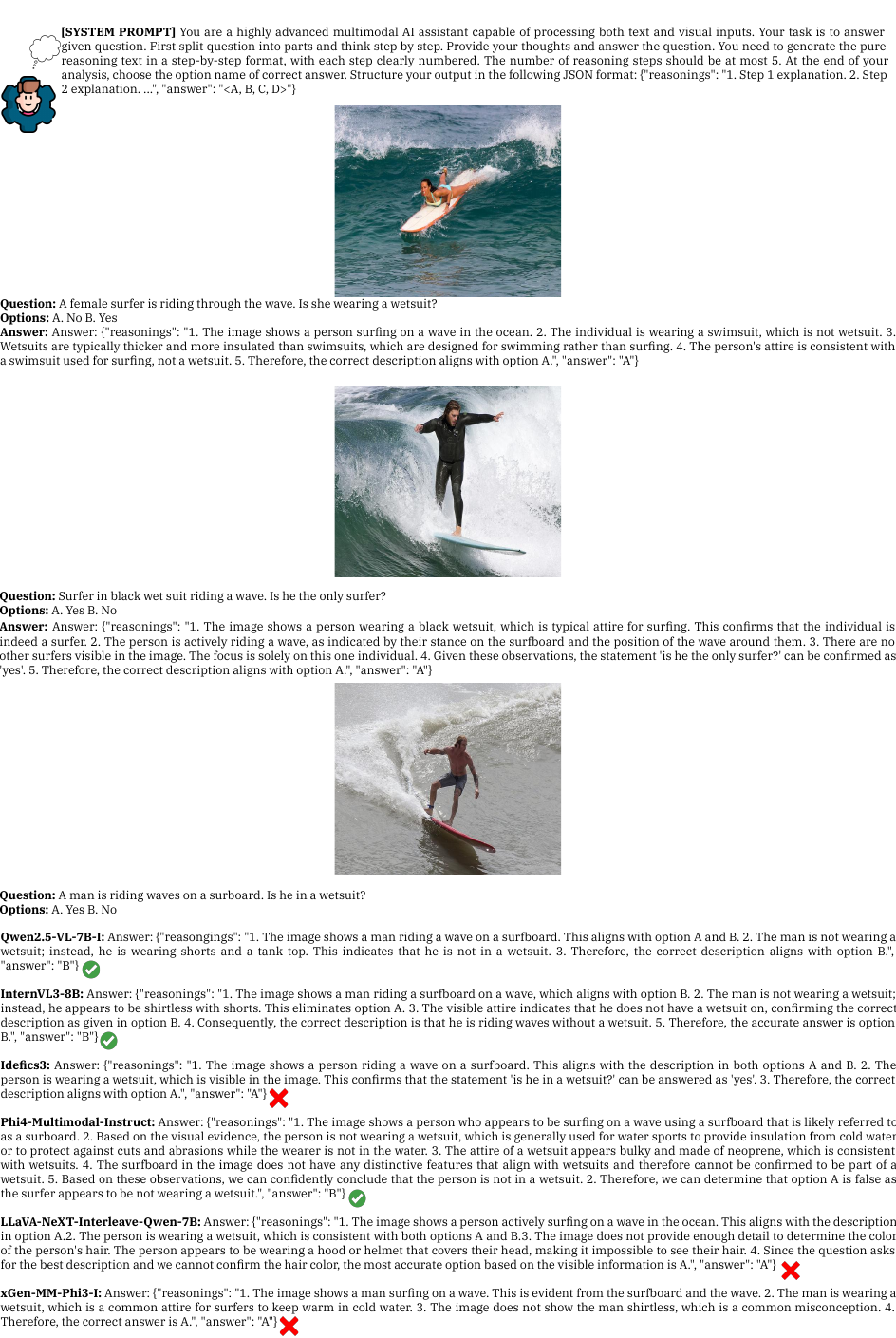}
    \caption{Example model predictions on instances from the \emph{Coreference Resolution} task, where demonstrations are selected based on visual and textual similarity, and Chain-of-Thought reasoning is employed.}
    \label{fig:cot-CoreferenceResolution-detailed-examples}
\end{figure*}

This section presents example model outputs from our evaluation. Figures \ref{fig:icl-AttributeRecognition-detailed-examples}–\ref{fig:icl-CoreferenceResolution-detailed-examples} illustrate the In-Context Learning (ICL) setup, where demonstrations are selected based on similarity to the query and include ground-truth answers. Figures \ref{fig:cot-AttributeRecognition-detailed-examples}–\ref{fig:cot-CoreferenceResolution-detailed-examples} show the ICL+CoT setup, where demonstrations additionally include Chain-of-Thought reasoning to guide the model’s inference on the query.

\section{Limitations}
\label{sec:limitations}
Our study, while comprehensive, is subject to several limitations. First, our investigations were constrained by computational resources, restricting our experiments to MLLMs with a maximum of 9 billion parameters. This limitation naturally precluded the inclusion of larger, and often higher-performing, proprietary models in our comparative analysis, which might exhibit different emergent behaviors.

Second, the evolving landscape of MLLMs meant our evaluations were limited to the models available at the time of our study that were capable of effectively handling interleaved image-text data. This selection criterion, while necessary, might not encompass the full spectrum of architectural innovations emerging in this rapidly developing field.

Furthermore, within our benchmark, we primarily focused on tasks involving language grounding in everyday scenes and conducted all evaluations exclusively in English. This means our findings may not generalize directly to other critical capabilities of MLLMs, different visual domains (e.g., medical imaging, satellite imagery), or other languages. Future work could explore these diverse areas to provide a more holistic understanding of MLLM performance.

Finally, we acknowledge that the performance of MLLMs can be highly sensitive to prompting strategies. In our experiments, we did not extensively explore alterations to prompts or permutations of the order of few-shot examples to further optimize model performance. While we aimed for a standardized evaluation, it's possible that more tailored prompting could yield improved results for specific models or tasks.

Crucially, we recognize the inherent challenge in fully accounting for the quality and diversity of the massive pretraining datasets used to develop MLLMs. These datasets, often composed of a mix of captioning data and interleaved image-text data, can vary significantly in their characteristics. Disentangling the precise contribution of dataset quality, scale, and specific data modalities (e.g., the balance between image-caption pairs versus more complex interleaved sequences) to the overall model performance is exceptionally difficult. Differences in pretraining data characteristics, which are often proprietary and not fully disclosed, can significantly influence and potentially skew reported results. This makes direct comparisons of models trained on vastly different pretraining regimes challenging. We emphasize that our findings reflect the performance of these models given their inherent pretraining characteristics, and further research is needed to isolate the impact of pretraining data itself on emergent MLLM capabilities.

\end{document}